\documentclass[journal]{IEEEtran}
\usepackage[T1]{fontenc}
\usepackage{color}
\usepackage{flushend}
\usepackage{amsmath}
\usepackage{cite}
\usepackage{algpseudocode} 

\usepackage{booktabs}
\usepackage{tabu}
\usepackage[cmintegrals]{newtxmath}
\usepackage{bm}
\usepackage{indentfirst}
         
\usepackage{amsmath,amssymb}
\usepackage{array}
\usepackage{mathrsfs}
\usepackage{cite}
\usepackage{tikz}
\usepackage{color}
\usepackage{amssymb}
\usepackage{url}
\usepackage{hyperref}
\usepackage{todonotes}
\usepackage{color}
\usepackage{bbding}
\usepackage{multirow}
\usepackage{colortbl}
\usepackage{amsmath}
\usepackage[ruled]{algorithm2e}

\begin{document}

\title{Open-Set Deepfake Detection: A Parameter-Efficient Adaptation Method with Forgery Style Mixture}

\author{Chenqi~Kong,~\IEEEmembership{Member,~IEEE}, Anwei~Luo, Peijun Bao, Haoliang~Li,~\IEEEmembership{Member,~IEEE}, 
Renjie~Wan,\\~\IEEEmembership{Member,~IEEE}, Zengwei ~Zheng,
Anderson~Rocha,~\IEEEmembership{Fellow,~IEEE},
and~Alex~C.~Kot,~\IEEEmembership{Life Fellow,~IEEE}%
\thanks{C. Kong and P. Bao are with the Rapid-Rich Object Search (ROSE) Lab, School of Electrical and Electronic Engineering, Nanyang Technology University, Singapore, 639798. (email: chenqi.kong@ntu.edu.sg, peijun001@e.ntu.edu.sg.)}
\thanks{A. Luo is with the School of Computing and Artificial Intelligence, Jiangxi University of Finance and Economics, Nanchang 330013, China, and also with Jiangxi Provincial Key Laboratory of Multimedia Intelligent Processing, Nanchang 330032, China.}
\thanks{R. Wan is with the Department of Computer Science, Hong Kong Baptist University, Hong Kong SAR. (email: renjiewan@hkbu.edu.hk).}
\thanks{H. Li is with the Department of Electrical Engineering, City University of Hong Kong, Hong Kong SAR. (email: haoliang.li@cityu.edu.hk).}
\thanks{Z. Zheng is with the Department of Computer Science and Computing, Zhejiang University City College, Zhejiang, China. (email: zhengzw@zucc.edu.cn).}
\thanks{A. Rocha is with the Artificial Intelligence Lab. (\texttt{Recod.ai}) at the University of Campinas, Campinas 13084-851, Brazil (e-mail: arrocha@unicamp.br), URL: \url{http://recod.ai}}
\thanks{A. C. Kot is with the Rapid-Rich Object Search (ROSE) Lab, School of Electrical and Electronic Engineering, Nanyang Technology University, Singapore, Shenzhen MSU-BIT University, Shenzhen, China, and also with VinUniversity, Hanoi, Vietnam. (email: eackot@ntu.edu.sg.)}
\thanks{Corresponding Author: Anwei Luo}
}

\markboth{Submitted to IEEE Transactions on Circuits and Systems for Video Technology}%
{Shell \MakeLowercase{\textit{et al.}}: Bare Demo of IEEEtran.cls for IEEE Communications Society Journals}

\maketitle

\begin{abstract}
Open-set face forgery detection poses significant security threats and presents substantial challenges for existing detection models. These detectors primarily have two limitations: they cannot generalize across unknown forgery domains or inefficiently adapt to new data. To address these issues, we introduce an approach that is both general and parameter-efficient for face forgery detection. \textcolor{black}{Our method builds on the assumption that different forgery source domains exhibit distinct style statistics. Specifically, we design a forgery-style-mixture formulation that augments the diversity of forgery source domains, enhancing the model’s generalizability across unseen domains. In addition, previous methods typically require fully fine-tuning pretrained networks, consuming substantial time and computational resources.} Drawing on recent advancements in vision transformers (ViT) for face forgery detection, we develop a parameter-efficient ViT-based detection model that includes lightweight forgery feature extraction modules and enables the model to extract global and local forgery clues simultaneously. We only optimize the inserted lightweight modules during training, maintaining the original ViT structure with its pre-trained weights. This training strategy effectively preserves the informative pre-trained knowledge while flexibly adapting the model to the task of Deepfake detection. Extensive experimental results demonstrate that the designed model achieves state-of-the-art generalizability with significantly reduced trainable parameters, representing an important step toward open-set Deepfake detection in the wild. 


\end{abstract}


\IEEEpeerreviewmaketitle
\vspace{-0.5cm}
\section{Introduction}

\begin{figure}[ht]
\centering
\includegraphics[scale=0.3]{  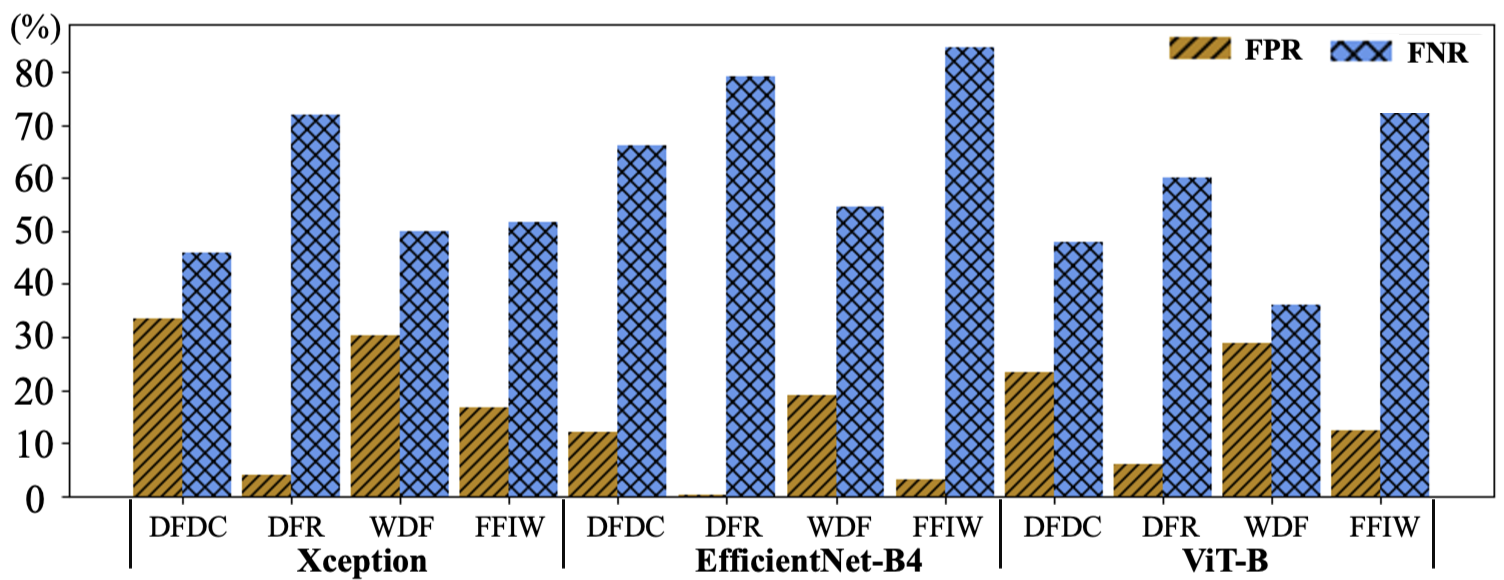}
\vspace{-0.4cm}
\caption{False Positive Rates (FPR) and False Negative Rates (FNR) of Xception, EfficientNet-B4, and ViT-B on four unforeseen Deepfake datasets: DFDC, DFR, WDF, and FFIW.}
\label{bar_fpr}
\end{figure}
\vspace{-0.1cm}
\IEEEPARstart{T}{he} rapid advancement of Artificial Intelligence Generated Content (AIGC) has led to significant progress in face forgery techniques, producing increasingly sophisticated fake faces that are indistinguishable from the human eye. Such advancements have fueled the proliferation of multimedia content on social media, posing high risks of disinformation, misinformation, and identity theft \cite{kong2022digital, Wang2024DeepfakeSurvey}. The recent emergence of diffusion and other advanced large vision-language models has exacerbated these challenges \cite{cardenuto2023age}. Consequently, developing effective detection methods to combat digital face forgery attacks has become critically important \cite{intelligence2024synthetic}. 

Face forgery detection has long posed a challenging yet crucial problem. Early methodologies proposed extracting hand-crafted features such as LBP maps \cite{wang2020face}, color components \cite{li2020identification}, and DCT maps \cite{qian2020thinking} as informative indicators for Deepfake detection. Subsequent works focused on capturing anomalies like eye-blinking \cite{li2018ictu}, head pose artifacts \cite{yang2019exposing}, and face-warping features \cite{li2018exposing}. Many methods aimed to improve detection accuracy by developing learning-based models using generic architectures such as Xception Net \cite{chollet2017xception}, Efficient Net \cite{tan2019efficientnet}, and Capsule Net \cite{nguyen2019capsule}. Later approaches, including Face X-ray \cite{li2020face}, SBI \cite{shiohara2022detecting}, DCL \cite{sun2022dual}, and RECCE \cite{sun2022dual} sought to capture common forgery features to enhance model generalizability. With the advent of ViT \cite{dosovitskiy2020image}, numerous ViT-based detectors \cite{zhuang2022uia, 10234594, cheng2024stacking, dong2022protecting, xu2024learning} have been developed. These approaches exploit global receptive fields and exhibit superior detection performance. However, previous face forgery detection methods still face two major challenges in real-world applications: 
(1) Existing methods struggle to generalize in uncontrolled environments. These models may encounter unforeseen forgery domains in the wild, which can lead to significant performance degradation; (2) \textcolor{black}{In real-world applications, detection models typically need fine-tuning with newly acquired data to adapt to different domains. However, this adaptation process for pre-trained detection models can be computationally expensive, making it challenging to deploy them effectively, particularly on mobile devices with limited computational resources.} 

\begin{figure*}[ht]
\centering
\includegraphics[scale=0.34]{  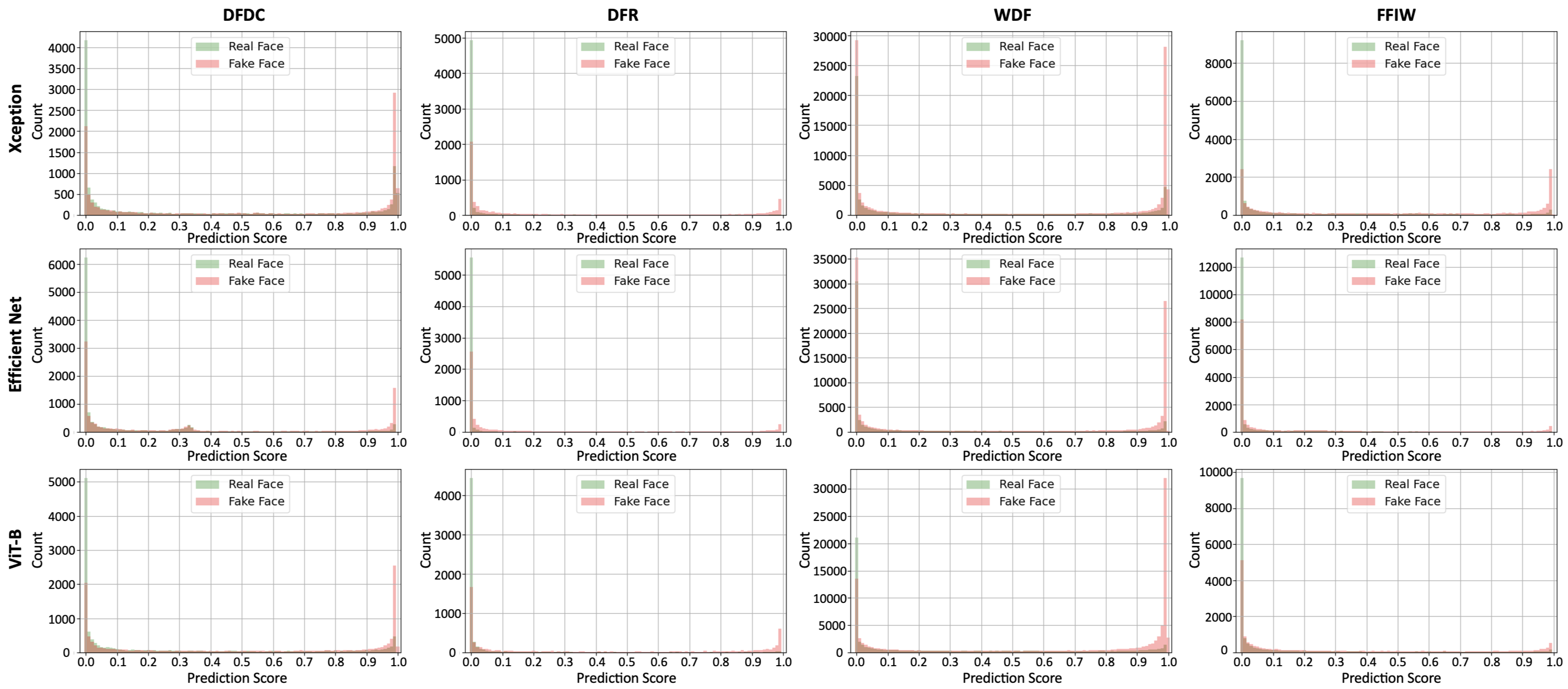}
\vspace{-6.5mm}
\caption{Prediction score distributions of real and fake faces of four unseen Deepfake datasets: DFDC, DFR, WDF, and FFIW. The three rows represent three Deepfake detectors: Xception, Efficient Net, and ViT-B. \textcolor{black}{It can be observed that domain gaps between different Deepfake datasets primarily impact the detection of forgery faces rather than real faces.}}
\label{score_distribution}
\end{figure*}

To address these two challenges, this paper introduces a generalized and parameter-efficient approach for \textbf{O}pen-\textbf{S}et \textbf{D}eep\textbf{F}ake \textbf{D}etection (\textbf{OSDFD}). Different forgery-style statistics may exhibit unique properties critical for face forgery detection. Fig.~\ref{bar_fpr} shows that existing forgery detectors have a significantly higher false negative rate (FNR) than the false positive rate (FPR) in unforeseen forgery domains, indicating a tendency to misclassify fake faces as real when encountering new domains in uncontrolled environments. 

\textcolor{black}{In Fig.~\ref{score_distribution}, we further demonstrate that \textbf{domain gaps between different Deepfake datasets primarily impact the detection of forgery faces rather than real faces.} Based on this observation, we propose a forgery style mixture module to \textbf{increase the diversity of source forgery domains during training}, thereby effectively mitigating forgery domain gaps in cross-dataset evaluations. Specifically, Fig.~\ref{score_distribution} presents cross-dataset experimental results for three representative face forgery detectors: Xception \cite{chollet2017xception}, EfficientNet \cite{tan2019efficientnet}, and ViT-B \cite{dosovitskiy2020image}. These models are trained on FF++ \cite{rossler2019faceforensics++} and evaluated on DFDC \cite{dolhansky2019deepfake}, DFR \cite{jiang2020deeperforensics}, WDF \cite{zi2020wilddeepfake}, and FFIW \cite{Zhou_2021_CVPR}. In Fig.~\ref{score_distribution}, green and red indicate real and fake faces, corresponding to ground-truth labels of `0' and `1', respectively. The $x$-axis and $y$-axis represent model prediction scores and the number of corresponding faces. As shown in Fig.~\ref{score_distribution}, while these detectors perform well in identifying real faces, they struggle with fake faces in cross-dataset evaluations. This finding suggests that discrepancies between source and target datasets primarily arises from \textbf{forgery domain gaps}. Motivated by this observation, we design the forgery style mixture module to enhance the diversity of source forgery domains, thereby improving the model’s generalizability.}

\begin{figure}[ht]
\centering
\includegraphics[scale=0.29]{  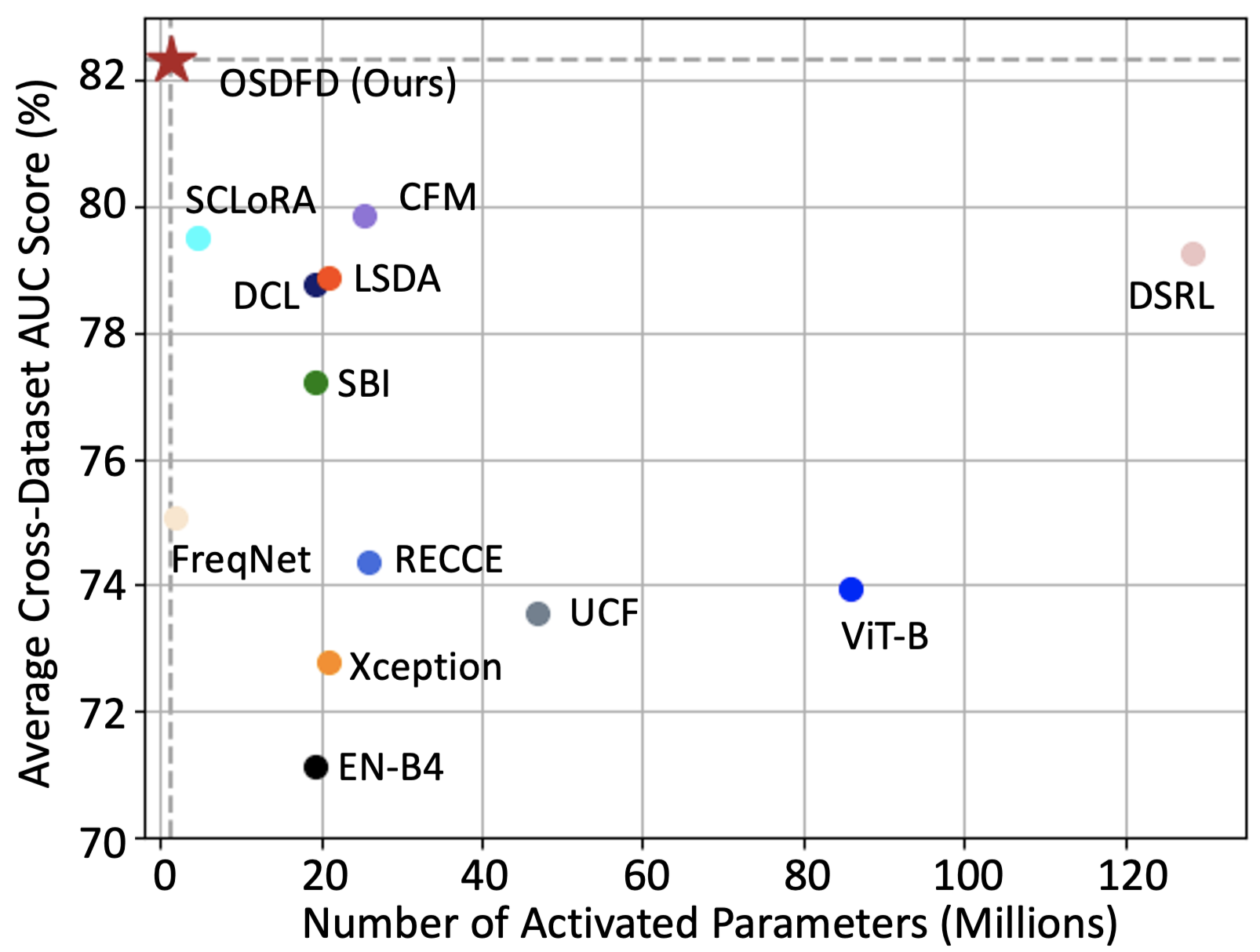}
\vspace{-3.5mm}
\caption{Average cross-dataset AUC score across six unseen datasets vs. the number of activated parameters. Our OSDFD method achieves the best generalizability with the fewest trainable parameters.}
\label{teaser}
\end{figure}

Furthermore, most prior art utilizes generic CNNs and ViTs as backbones and fully fine-tune the networks from weights. However, fully fine-tuning a network is costly, hindering its real-world applications. To mitigate this issue, we integrate lightweight Adapter and LoRA layers into the ViT backbone. During training, we freeze plain ViT backbones with weights and solely update the lightweight Adapter and LoRA layers. These designed modules capture local and global forgery artifacts using slim components.
Additionally, these modules preserve substantial ImageNet knowledge and enable flexible model adaptation for face forgery detection. \textcolor{black}{While parameter-efficient fine-tuning (PEFT) has been widely adopted in various domains, our work makes a contribution by integrating PEFT with a forgery-aware perspective, specifically tailored for open-set face forgery detection—a task that presents unique challenges in generalizability and adaptation.}
Fig.~\ref{teaser} illustrates the average cross-dataset AUC performance across six unforeseen Deepfake datasets versus the number of activated parameters. This demonstrates that our method achieves superior generalizability with the fewest activated parameters. 

This work extends upon our previous conference work \cite{kong2023enhancing}. The main distinctions between this work and \cite{kong2023enhancing} are summarized as follows: (1) We introduce the Central Difference Convolution (CDC) adapter to incorporate local forgery priors into the expressive ViT model, thereby enhancing the model's detection performance, (2) We propose a forgery style mixture module that combines source forgery domains to improve the model's generalizability across unforeseen forgery domains; (3) We further reduce the dimension of LoRA layers, enabling more parameter-efficient face forgery detection; (4) We conduct extensive additional experiments compared with more SOTA methods, including cross-dataset evaluations, robustness experiments, visualization results, and ablation studies.    

In summary, the main contributions of this work are as follows:
\begin{itemize}
\item We design a forgery-style-mixture module to enhance the diversity of source domains. The model's generalizability is effectively improved by randomly mixing forgery styles.
\item \textcolor{black}{Our approach explicitly captures forgery-related artifacts by integrating a CDC adapter and a LoRA module into a ViT-based detection framework. This targeted design enhances the model’s ability to extract both global and local forgery cues, which is crucial for Deepfake detection.}
\item \textcolor{black}{Our designed components jointly contribute to an effective and scalable solution for open-set Deepfake detection.} Extensive experiments demonstrate that our model achieves state-of-the-art generalizability and robustness with the fewest activated parameters. 
\end{itemize}

We arranged the remaining of this paper into four more sections, Sec. II comprehensively reviews previous literature on face forgery detection and parameter-efficient fine-tuning. Sec. III details the proposed detection framework. Sec. IV presents extensive experimental results under various settings—finally, Sec. V concludes the paper and discusses possible future research directions.  

\section{Related Work}
\subsection{Face forgery detection}
Early face forgery detection methods initially focused on extracting hand-crafted features, such as eye-blinking frequency \cite{li2018ictu}, head pose inconsistency \cite{yang2019exposing}, face warping artifacts \cite{li2018exposing}, and rPPG features \cite{ciftci2020fakecatcher}. With the rise of deep learning and artificial intelligence, numerous learning-based methods emerged to improve detection accuracy. Some generic CNNs, such as the Xception network \cite{chollet2017xception, kong2022detect, luo2021generalizing, qian2020thinking, chen2021robust}, EfficientNet-b4 \cite{tan2019efficientnet, luo2023beyond, zhang2023bi, zhang2024face, nguyen2024laa}, and Capsule network \cite{nguyen2019capsule}, have been utilized for face forgery detection, showing promising results. However, these methods often need to be more balanced in training data, leading to poor generalization capabilities. In response, some approaches have focused on extracting common forgery features. For example, Face X-ray \cite{li2020face} and SBI \cite{shiohara2022detecting} aimed to extract common face blending features, enhancing model generalizability. F$^{3}$Net and Luo et al. \cite{luo2021generalizing} mined frequency artifacts, while DCL \cite{sun2022dual} and CFM \cite{luo2023beyond} employed contrastive learning schemes for more generalized Deepfake detection. NoiseDF \cite{wang2023noise} raised an innovative idea to distinguish Deepface-specific noise traces from common noises within each single frame, providing additional interpretability. With the rapid advancements in vision transformers (ViT), several ViT-based approaches \cite{dong2022protecting, zhuang2022uia, kong2023enhancing, guan2022delving, cheng2023voice, wang2023deep, yu2023narrowing, cheng2024diffusion, luo2024forgery, yu2024distilling, wang2023spatial, guo2023ldfnet, liao2023famm, cheng2023voice} have been developed, showing superior detection performance. ICT \cite{dong2022protecting} introduced the Identity Consistency Transformer to capture identity inconsistency within fake faces, while F$^{2}$Trans \cite{miao2023f} designed the High-Frequency Fine-Grained Transformer, achieving more generalized and robust forgery detection. 
\textcolor{black}{Forensics Adapter \cite{cui2025forensics} is a lightweight network that equips CLIP with forgery-specific knowledge by learning blending boundary traces and enhancing visual tokens through a dedicated interaction strategy. TFCU \cite{guo2025face} improves face forgery video detection by sequentially capturing momentary anomalies, gradual inconsistencies, and cumulative distortions through its consecutive correlation, future-guided, and historical review modules. OWG-DS \cite{guo2025towards} advances real-world Deepfake detection by transferring knowledge from limited labeled data to large-scale unlabeled data. GM-DF \cite{lai2024gm} enhances Deepfake detection generalization across multiple datasets, enabling effective handling of diverse real-world scenarios and unseen attacks. DAID \cite{cheng2025fair} mitigates demographic biases in Deepfake detection through demographic-aware data rebalancing and demographic-agnostic feature aggregation.
Recent LVLM-based approaches \cite{huang2024ffaa, jia2024can, yu2025unlocking, zhang2025mfclip, guo2025rethinking, hu2025seeing, chen2024x2} take a further step forward by achieving explainable face forgery detection.} 
However, the scalability of ViT-based methods is largely constrained by the quadratic complexity of attention computation, limiting their deployment, particularly on mobile devices with limited computing resources. In response, this paper introduces a parameter-efficient fine-tuning (PEFT) strategy for face forgery detection to address this issue. Furthermore, we combine different forgery styles to augment the forgery source domains,  enhancing the model's generalization capability.

\subsection{Parameter-efficient fine-tuning (PEFT)}
The attention mechanism has demonstrated significant success across various computer vision tasks. However, the self-attention in ViT scales quadratically concerning the input patch number, leading to substantial computational demands during training. Several methods have been proposed to address this issue, including parameter-efficient fine-tuning (PEFT). PEFT selectively updates a small portion of the model parameters, facilitating model adaptation to downstream tasks. Adapters \cite{houlsby2019parameter, sung2022lst} are typically inserted between feedforward and normalization layers of transformers, comprising a down-sample and up-sample layer. For example, Low-Rank Adaptation (LoRA) \cite{hu2021lora} decomposes a matrix into two trainable low-rank matrices, significantly reducing the number of parameters compared to the original matrix. Visual Prompt Tuning (VPT) \cite{jia2022visual} introduces a small number of trainable parameters in the input space, treating additional tokens as learnable pixels. Convpass \cite{jie2022convolutional} aims to inject local priors into expressive ViT models by integrating convolutional operations, enhancing performance across various computer vision tasks. Neural Prompt Search (NOAH) \cite{zhang2022neural} further advances by incorporating Adapter, LoRA, and VPT into vision transformers and optimizing their design through a neural architecture search algorithm. Despite these advancements, the development of PEFT methods for face forgery detection has been largely understudied, yet it is crucial in practical scenarios. In this paper, we integrate lightweight Adapter and LoRA modules into the plain ViT backbone, optimizing only these modules while keeping the ViT backbone fixed with pretrained weights during training. The designed modules effectively capture global and local face forgery features more efficiently, without the danger of catastrophic forgetting typical of fully finetuning models \cite{ramasesh2021effect}.

\section{Proposed Method}
\subsection{Overall framework}
Fig.~\ref{overview} illustrates the overview of the proposed face forgery detection method. The designed framework relies upon ViT, which is pre-trained on ImageNet as the backbone. As Fig.~\ref{overview} (a) depicts, the features extracted by the transformer blocks are fed forward to the Forgery Style Mixture module. This module exclusively mixes the statistics of forgery styles, augmenting the forgery source domains during training. The processed features are delivered to the MLP head for the final decision-making. Fig.~\ref{overview} (b) illustrates the details of transformer blocks. We insert the trainable Adapter and LoRA layers into the ViT block. The ViT backbone is fixed with pre-trained ImageNet weights. Fig.~\ref{overview} (c) and Fig.~\ref{overview} (d)  detail the Adapter and LoRA layers. In Fig.~\ref{overview} (c), the input hidden state $h^{in}$ is followed by three trainable convolutional layers. Moreover, $h^{in}$ is simultaneously multiplied with the ImageNet pretrained weight. The output hidden state $h^{out}$ is the element summation of the two features. In turn, the LoRA layer dedicated in Fig.~\ref{overview} (d) introduces two  trainable low-rank matrices $W^{down} \in \mathbb{R}^{d \times r}$ and $W^{up}\in \mathbb{R}^{r \times d}$, where $r<<d$. Moreover, $h^{out}$ is obtained by conducting an element-wise summation of the two outputs.

\subsection{Forgery-aware PEFT}
Our proposed method aims to capture critical forgery clues using lightweight modules to facilitate face forgery detection in practical scenarios. We integrate forgery-aware PEFT modules into plain ViTs. These modules include the LoRA layer, injected into the self-attention blocks, and the adapter layer, integrated into the feed-forward networks (FFN).

\begin{figure}[ht]
\centering
\includegraphics[scale=0.36]{  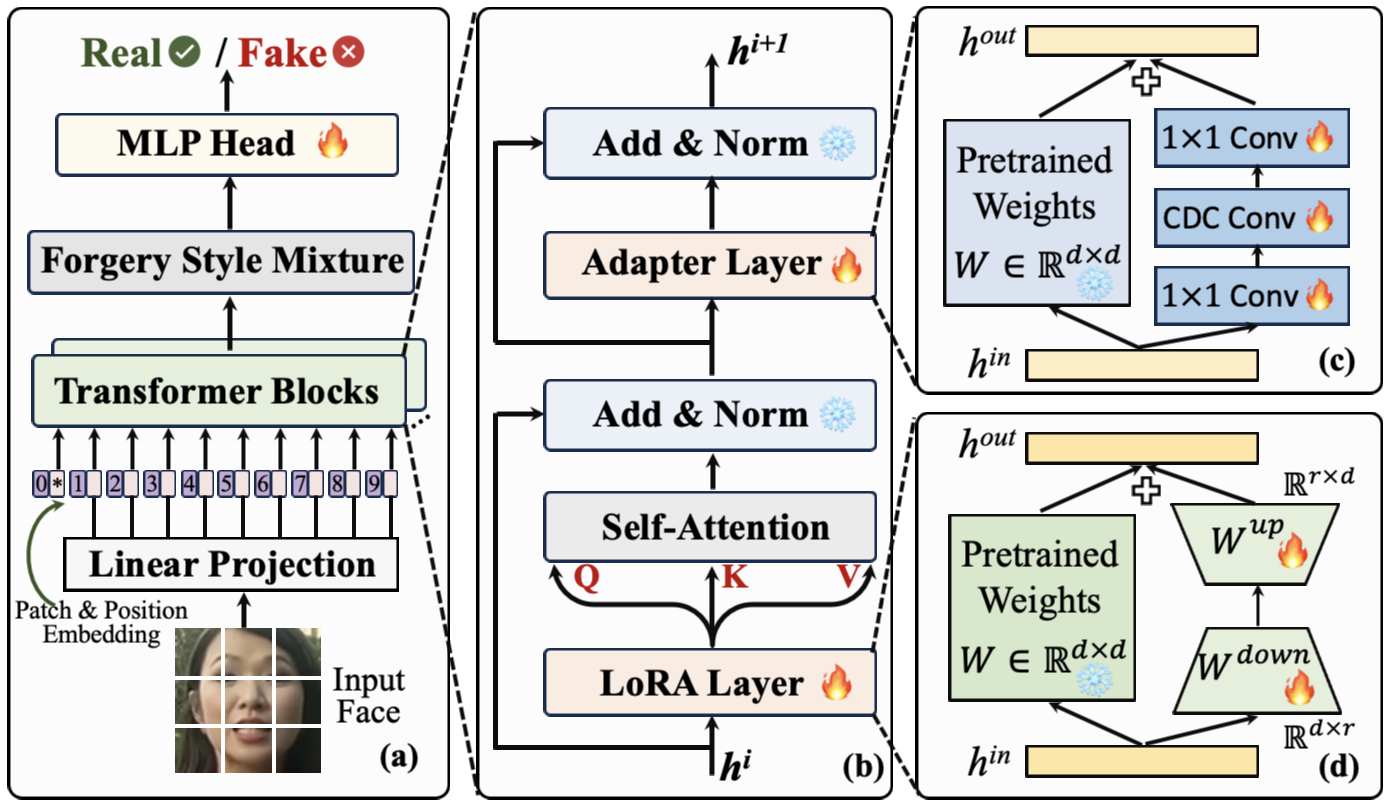}
\vspace{-6.5mm}
\caption{Overview of the designed face forgery detection framework. (a) Overall model structure; (b) Structure of the designed transformer block; (c) Details of the designed adapter layer; (d) Details of the designed LoRA layer.}
\label{overview}
\end{figure}

As Fig.~\ref{overview} (c) depicts, the input hidden state $h^{in}$ is simultaneously forwarded to the fixed ImageNet MLP weights and the designed trainable convolutional layers. Subsequently, the outputs of these processes are element-wise summed before being passed to the subsequent vision transformer blocks:
\begin{equation}
     h^{out} = {\rm MLP}(h^{in}) + {\rm Adapter}(h^{in}).
\end{equation}
The Adapter layer is formulated as:
\begin{equation}
     {\rm Adapter}(h^{in}) = {\rm Conv_{1\times 1}^{up}}({\rm CDC}({\rm Conv_{1\times 1}^{down}}(h^{in}))).
\end{equation}

\begin{figure}[ht]
\centering
\includegraphics[scale=0.32]{  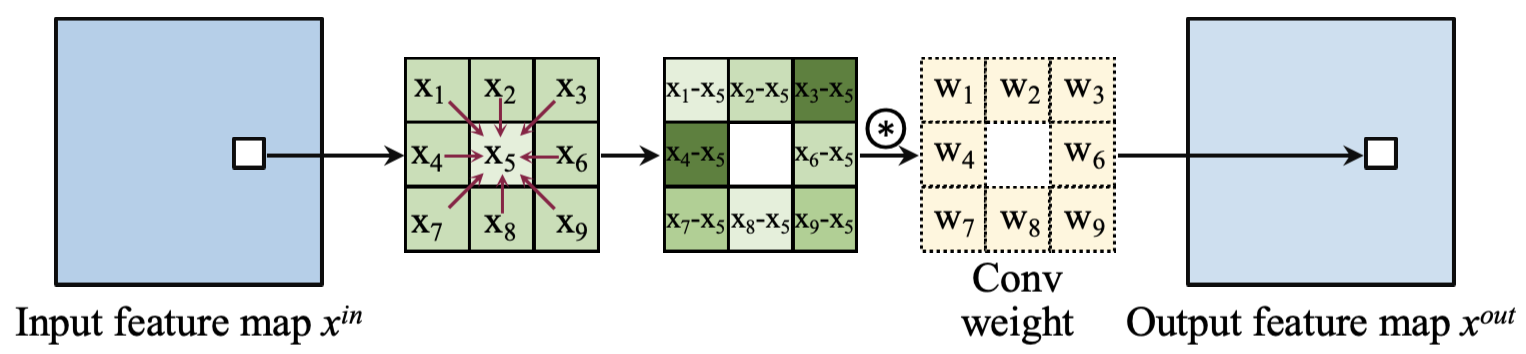}
\vspace{-3.5mm}
\caption{Illustration of the Central-Difference Convolution (CDC) pipeline.}
\label{CDC}
\end{figure}

\noindent We first expand the input hidden state $h^{in}$ into a 2-D dimension. Next, we reduce the dimension of the expanded feature using a $1\times1$ convolutional layer, effectively decreasing the number of parameters activated in the Adapter layer. We then design the Central Difference Convolution (CDC) operator to expose local forgery artifacts for each input query. The pipeline of the CDC operator is illustrated in Fig.~\ref{CDC}. The output feature map $x^{out}$ can be calculated by:
\begin{equation}
     x^{out} = \sum_{i \in \Omega} w_{i} (x^{in}_{i}-x_{c}),
\end{equation}
\noindent where $x_{c}$ indicates the center pixel of the window $\Omega$, and $x_{5}$ represents $x_{c}$ in Fig.~\ref{CDC}. $x_{i}$ indicates the peripheral pixels within $\Omega$. $w_{i}$ denotes the corresponding convolutional weights. Finally, the second $1\times1$ Conv layer restores the channels of the feature, and the produced feature is flattened to the same shape as $h^{in}$. \textcolor{black}{CDC operates within a local sliding window, calculating differences between peripheral and central pixels to extract high-frequency features \cite{su2021pixel}. The CDC adapter functions as a local anomaly extractor for the ViT backbone by capturing subtle forgery cues, such as boundary inconsistencies and local irregularities. These local cues are then integrated with the backbone’s global features, while the PEFT strategy ensures this integration is achieved effectively with minimal trainable parameters.}
Our designed adapter layer effectively introduces inductive bias into standard ViTs, guiding the model in learning local artifacts for face forgery detection. Additionally, considering the effectiveness of high-frequency features in the task of face forgery detection \cite{qian2020thinking, miao2022hierarchical, fei2022learning, yang2021mtd}, our CDC operator enables the model to extract abundant high-frequency information from input faces, thereby enhancing detection performance.    

In addition to the Adapter layer, which introduces local dependencies into plain ViT backbones, we further design the Low-Rank Adaptation (LoRA) layer to obtain global receptive fields with minimal parameters. As depicted in Fig.~\ref{overview} (d), the output hidden state $h^{out}$ is calculated by:
 
\begin{figure}[ht]
\centering
\includegraphics[scale=0.29]{  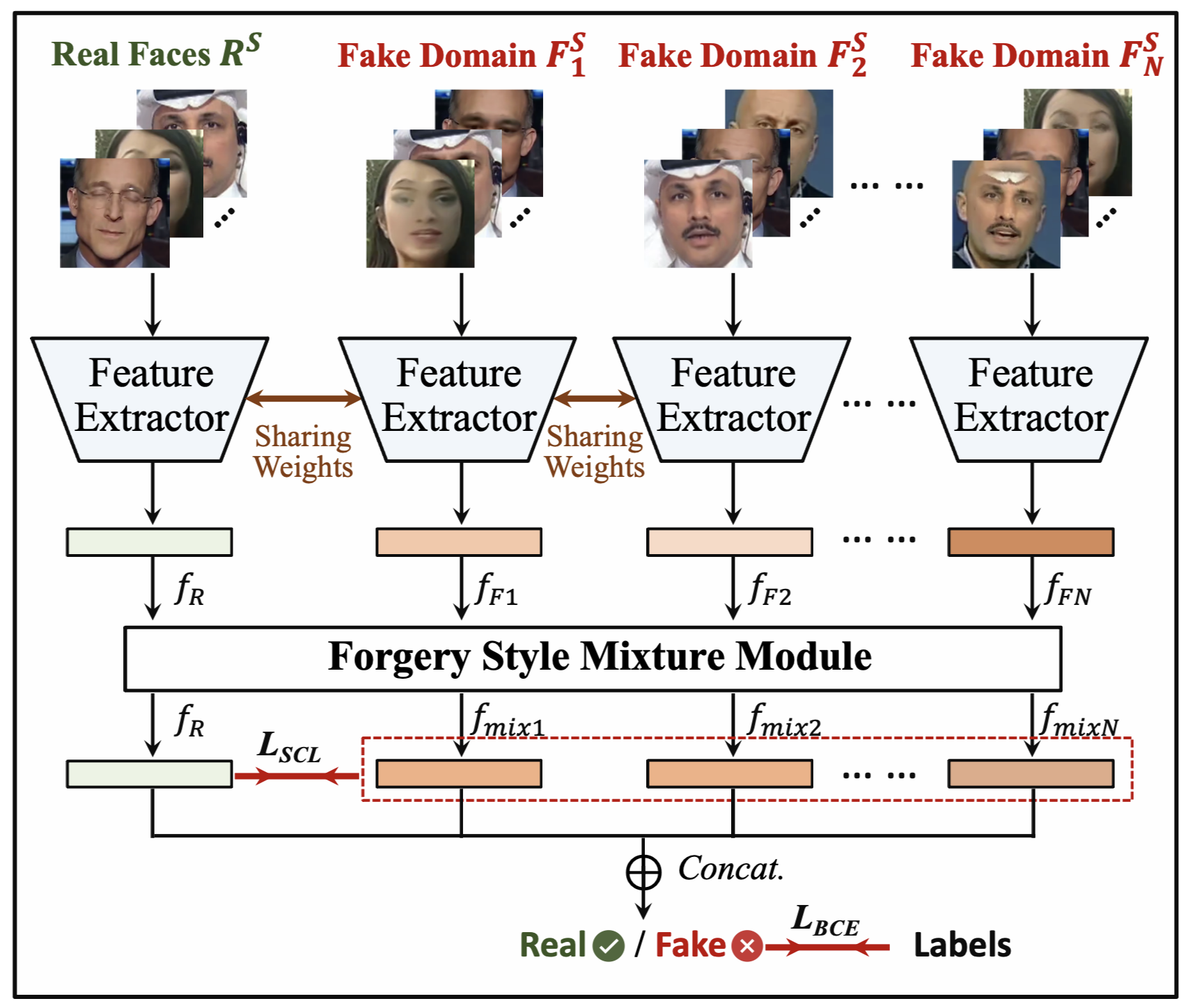}
\vspace{-3.5mm}
\caption{\textcolor{black}{Illustration of forgery style mixture pipeline. The designed feature extractor processes real faces and forgery source domain data, and the extracted features are subsequently delivered to the forgery-style-mixture module.}} 
\label{forgery_mixer}
\end{figure}

\begin{equation}
     h^{out} = W_{q/k/v}h^{in} + {\rm LoRA}(h^{in}).
\end{equation}

The $W_{q/k/v}$ parameter denotes the pre-trained weights fixed during the training process. Moreover, the LoRA module consists of two low-rank matrices:

\begin{equation}
     {\rm LoRA}(h^{in}) = h^{in}W_{q/k/v}^{down}W_{q/k/v}^{up},
\end{equation}
where $h^{in} \in \mathbb{R}^{d \times d}$, $W_{q/k/v}^{down} \in \mathbb{R}^{d \times r}$, $W_{q/k/v}^{up} \in \mathbb{R}^{r \times d}$, and $r<<d$. We optimize the slim, decomposed low-rank matrices during training instead of the redundant pre-trained weights. As such, the number of trainable parameters is significantly reduced. In this work, we set the hidden state dimension $d$ of ViT-B to 768, and the rank $r$ is defaulted to 8. The designed LoRA and CDC adapter respectively capture expressive global features and extract informative local forgery clues, thereby enhancing the final detection performance.   

\textcolor{black}{The designed forgery-aware PEFT modules bear the following advantages for face forgery detection: (1) By optimizing a small subset of parameters, our method reduces computational and memory requirements compared to full fine-tuning while also accelerating training, making frequent updates more practical in open-set environments. Moreover, storing the backbone weights in the cloud while keeping only a minimal set of parameters on edge devices facilitates deployment on resource-constrained mobile platforms. \textcolor{black}{(2) As model sizes continue to grow, our forgery-aware PEFT strategy remains scalable for large models, and the designed modules can be seamlessly integrated with various transformer-based backbones in a plug-and-play fashion.} (3) Our approach efficiently incorporates forgery-specific knowledge while preserving the pretrained model's general knowledge, effectively mitigating overfitting to forgery data and enhancing generalization performance.} 


\begin{figure}[ht]
\centering
\includegraphics[scale=0.26]{  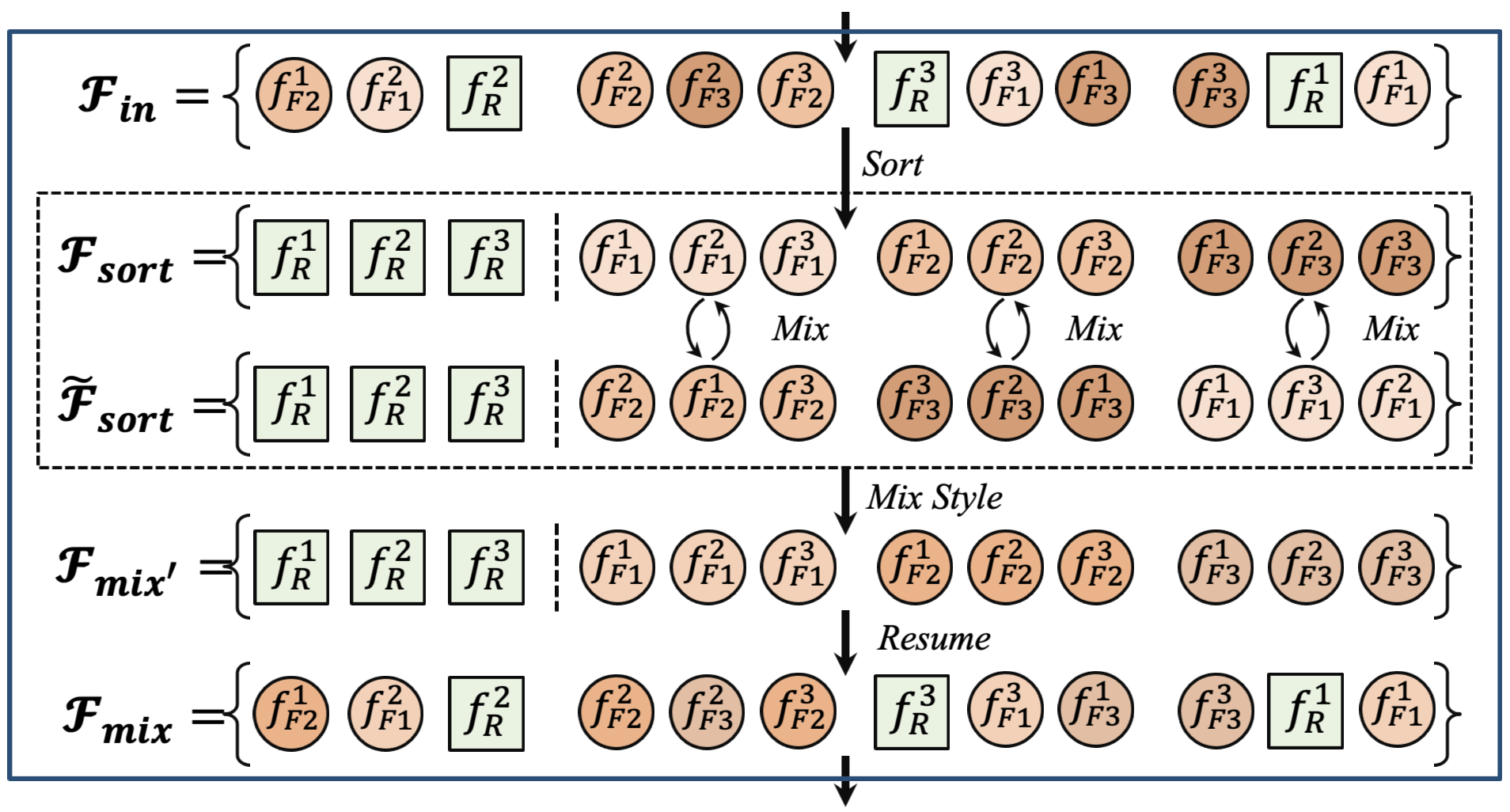}
\vspace{-7.5mm}
\caption{\textcolor{black}{Pipeline of the proposed forgery feature mixture process.}} 
\label{style_mixture}
\end{figure}

\subsection{Forgery style mixture}
In face forgery detection, the training dataset often encompasses multiple forgery techniques. This training setting accommodates practical scenarios where available source domains are utilized. In this vein, we denote the training dataset as $D^{S}=\{R^{S}, F_{1}^{S}, F_{2}^{S}, ... , F_{N}^{S}\}$. \textcolor{black}{$R^{S}$ represents the real source faces, while $F_{i}^{S}$ denotes the $i_{th}$ source forgery domain, where $i\in \{1,2,...,N\}$. Following previous methods in face forgery detection, we use FF++ as the training set, which includes multiple training forgery domains (i.e., manipulation types). } 
\textcolor{black}{As such, the target testing domain can be denoted as $D^{T}=\{R^{T}, F^{T}\}$. Similarly, $R^{T}$ and $F^{T}$ are target real and forgery domains.} 
Our preliminary experiments, illustrated in Fig.~\ref{bar_fpr} and Fig.~\ref{score_distribution}, reveals a high false negative rate (FNR) and a low false positive rate (FPR) in cross-domain evaluations, suggesting a negligible domain gap between $R^{S}$ and $R^{T}$ and significant domain gaps between $F^{S}$ and $F^{T}$. As such, we propose to augment source forgery domains to mitigate this issue. Fig.~\ref{forgery_mixer} depicts the proposed forgery-style-mixture pipeline. \textcolor{black}{Each training batch, composed of one real domain and multiple fake domains, is processed by our proposed feature extractor, which shares weights across all source domains.}
The extracted features from the source forgery domains are passed to the designed forgery-style-mixture module, and the mixed features are subsequently forwarded to the classifier to perform forgery detection.  

Next, we elaborate on the details of the devised forgery style mixture module, as depicted in Fig.~\ref{style_mixture}. Green squares denote real faces in this representation, while red circles indicate the forgery faces. 
\textcolor{black}{This module processes each training batch consisting of real and fake face images sampled from the real source domain and $N$ forgery source domains. In Fig.~\ref{style_mixture}, the subscripts of $f$ represent the domain labels, while the superscripts denote the sample index within each domain. We illustrate the case of $N$=3 forgery domains in Fig.~\ref{style_mixture}, with 3 samples per domain.} 
The features \boldsymbol{$\mathcal{F}_{in}$} of each input batch is proceeded to our designed Forgery Style Mixture module. 

We first sort \boldsymbol{$\mathcal{F}_{in}$} to \boldsymbol{$\mathcal{F}_{sort}$} according to the domain labels, where $f_{R}$ represents real features, and $f_{Fi}$ with $i\in \{1,2,3\}$, denotes features from different forgery domains. To augment the forgery style, we first shuffle the fake features of \boldsymbol{$\mathcal{F}_{sort}$}, ensuring that each forgery domain label of the shuffled \boldsymbol{$\widetilde{\mathcal{F}}_{sort}$} differs from the original \boldsymbol{$\mathcal{F}_{sort}$}. In addition, we keep the real feature order unchanged. \textcolor{black}{Subsequently, we mix the original \boldsymbol{$\mathcal{F}_{sort}$} and the shuffled \boldsymbol{$\widetilde{\mathcal{F}}_{sort}$} by fusing their feature statistics. The details of this process are provided next.}

We draw inspiration from Adaptive Instance Normalization (AdaIN) \cite{huang2017arbitrary}, a widely employed image style transfer technique. The source style of $p$ can be transferred to the target style of $q$ by:
\begin{equation}
     {\rm AdaIN}(p) = \sigma (q) \frac{p-\mu (p)}{\sigma (p)} + \mu (q),
\end{equation}
where $\mu(p)$ and $\sigma(p)$ are the spatial mean and standard deviation of the feature tensor $p$.  

Inspired by the AdaIN, which utilizes the feature statistics of a target style input to transfer the image style, we first calculate the mixed feature statistics of \boldsymbol{$\mathcal{F}_{sort}$} and \boldsymbol{$\widetilde{\mathcal{F}}_{sort}$}: 


\begin{equation}
     \gamma_{mix} = \delta \sigma (\boldsymbol{\mathcal{F}_{sort}}) + (1-\delta)\sigma (\boldsymbol{\widetilde{\mathcal{F}}_{sort}}),
\end{equation}
\begin{equation}
     \eta_{mix} = \delta \mu (\boldsymbol{\mathcal{F}_{sort}}) + (1-\delta)\mu (\boldsymbol{\widetilde{\mathcal{F}}_{sort}}),
\end{equation}
where $\delta$ is a weight sampled from Beta distribution, i.e., $\delta \sim Beta(0.1, 0.1)$. $\gamma_{mix}$ and $\eta_{mix}$ are the weighted summation of the forgery style statistics of \boldsymbol{$\mathcal{F}_{sort}$} and \boldsymbol{$\widetilde{\mathcal{F}}_{sort}$}. Then, the style mixture is performed by the mixed feature statistics $\gamma_{mix}$ and $\eta_{mix}$ to the original feature \boldsymbol{$\mathcal{F}_{sort}$}:
\begin{equation}
     \boldsymbol{\mathcal{F}_{mix^{'}}} = {\rm MixStyle}(\boldsymbol{\mathcal{F}_{sort}}) = \gamma_{mix} \frac{\boldsymbol{\mathcal{F}_{sort}}-\mu (\boldsymbol{\mathcal{F}_{sort})}}{\sigma (\boldsymbol{\mathcal{F}_{sort})}} + \eta_{mix}.
\end{equation}
Finally, we restore the original index order of the mixed features \boldsymbol{$\mathcal{F}_{mix'}$} to obtain \boldsymbol{$\mathcal{F}_{mix}$}, which is forwarded to the classifier for final decision-making. Similar to \cite{zhou2021domain}, we set a default probability of 0.5 to activate the forgery-style-mixture module.
In the inference stage, the forgery-style-mixture is deactivated. \textcolor{black}{As such, the designed module augments the source forgery domains in the training phase, thereby mitigating the overfitting problem and enhance the model's generalizability. This design effectively improves the model’s ability to generalize in open-set face forgery detection.} 

\textcolor{black}{In contrast to conventional computer vision tasks, forgery detection suffers from imbalanced domain gaps: the discrepancies between different fake domains are substantially larger than those among real domains, due to the continual evolution of manipulation techniques. Forgery Style Mixture (FSM) is a label-guided mechanism that explicitly mixes feature statistics across different forgery styles. Unlike prior augmentation strategies that enhance general feature diversity, FSM specifically targets forgery-style diversity, thereby expanding the fake feature space without requiring additional data.}



\begin{table*}
  \centering
  \caption{High quality (c23) cross-manipulation detection performance. (\#Params:  trainable parameter number)}
  \vspace{-2.5mm}
  \scalebox{0.92}{\begin{tabular}{l|c|c|cc|cc|cc|cc|cc}
    \hline
    \multirow{2}{*}{\textbf{Method}} & \multirow{2}{*}{\textbf{Venue}} & \multirow{2}{*}{\textbf{\#Params}} & \multicolumn{2}{c|}{\textbf{FF,FS,NT$\rightarrow$DF}} & \multicolumn{2}{c|}{\textbf{DF,FS,NT$\rightarrow$FF}} & \multicolumn{2}{c|}{\textbf{DF,FF,NT$\rightarrow$FS}} & \multicolumn{2}{c|}{\textbf{DF,FF,FS$\rightarrow$NT}} & \multicolumn{2}{c}{\textbf{Average}}\\ \cline{4-13}
     & & & AUC & ACC & AUC & ACC & AUC & ACC & AUC & ACC & AUC & ACC\\	
    \hline							
    ResNet18 \cite{he2016deep} & CVPR16 & 11.18M & 0.813 & 0.656 & 0.746 & 0.596 & 0.464 & 0.476 & 0.688 & 0.528 & 0.678 & 0.564 \\
    \hline	
    Xception \cite{chollet2017xception} & CVPR17 & 20.81M & 0.907 & 0.795 & 0.753 & 0.558 & 0.460 & 0.472 & 0.744 & 0.557 & 0.716 & 0.596 \\ 		
    \hline	 				
    EN-B4 \cite{tan2019efficientnet} & ICML19 & 19.34M & 0.485 & 0.495 & 0.556 & 0.523 & 0.517 & 0.517 & 0.493 & 0.500 & 0.513 & 0.509 \\ 
    \hline	 				
    All-train EN-B4 \cite{tan2019efficientnet} & ICML19 & 19.34M & 0.911 & 0.824 & 0.801 & 0.633 & 0.543 & 0.500 & 0.774 & 0.608  & 0.757 & 0.641 \\	 
    \hline	 				
    Focal-loss EN-B4 \cite{tan2019efficientnet} & ICML19 & 19.34M & 0.903 & 0.813 & 0.798 & 0.608 & 0.503 & 0.484 & 0.759 &  0.604 & 0.741 & 0.627 \\ 
    \hline			
    Forensics Transfer \cite{cozzolino2018forensictransfer} & ArXiv18 & - & - & 0.720 & - & 0.645 & - & 0.460 & - & 0.569 & - & 0.599 \\ 	
    \hline					
    Multi-task \cite{nguyen2019multi} & BTAS19 & - & - & 0.703 & - & 0.587 & - & 0.497 & - & 0.603 & - & 0.598 \\	
    \hline					
    MLDG \cite{li2018learning} & AAAI18 & 62.38M & 0.918 & 0.842 & 0.771 & 0.634 & 0.609 & 0.527 & {0.780} & 0.621 & 0.770 & 0.656 \\ 
    \hline	 								
    LTW \cite{sun2021domain} & AAAI21 & 20.37M & 0.927 & 0.856 & 0.802 & 0.656 & 0.640 & 0.549 & 0.773 & {0.653} & 0.786 & 0.679 \\ 
    \hline			
    CFM \cite{10315169} & TIFS24 & 25.37M & 0.880 & 0.617 & 0.814 & 0.557 & 0.631 & 0.516 & 0.644 & 0.508 & 0.742 & 0.550\\
    \hline		
    \textcolor{black}{DCL} \cite{sun2022dual} & AAAI22 & 19.35M & 0.949 & 0.877 & 0.829 & 0.684 & - & - & - & - & - & - \\  \hline
    \textcolor{black}{Implicit ID} \cite{huang2023implicit} & CVPR23 & 22.36M  & {0.950} & {0.882} & 0.844 & 0.694 & - & - & - & - & - & - \\
    \hline	
    \textcolor{black}{UCF} \cite{yan2023ucf} & CVPR23 & 20.81M  & - & 0.867 & - & 0.679 & - & - & - & - & - & - \\
    \hline
    \textcolor{black}{DiffusionFake} \cite{sun2024diffusionfake} & NIPS24 & 19.34M & - & {0.882} & - & 0.702 & - & - & - & - & - & - \\
    \hline
    SCLoRA \cite{kong2023enhancing} & MIPR23 & 4.72M & {0.935} & 0.862 & {0.875} & {0.753} & {0.651} & {0.554} & 0.707 & 0.626 & {0.792} & {0.699} \\		
    \hline
    ViT-B \cite{dosovitskiy2020image} & ICLR21 & 85.80M & 0.771 & 0.701 & 0.656 & 0.582 & 0.510 & 0.498 & 0.554 & 0.517 & 0.623 &  0.575 \\		
    \hline	
    \textcolor{black}{CLIP} \cite{radford2021learning} & \textcolor{black}{ICML21} & \textcolor{black}{303.18M} & \textcolor{black}{0.890} & \textcolor{black}{0.766} & \textcolor{black}{0.847} & \textcolor{black}{0.737} & \textcolor{black}{0.541} & \textcolor{black}{0.504} & \textcolor{black}{0.652} & \textcolor{black}{0.538} & \textcolor{black}{0.733} & \textcolor{black}{0.636} \\		
    \hline	
    \hline
    OSDFD (ViT-B) & Ours & 1.34M & \underline{0.962} & \underline{0.885} & \underline{0.932} & \underline{0.846} & \underline{0.698} & \underline{0.595} & \underline{0.781} & \underline{0.682} & \underline{0.843} & \underline{0.752} \\
    \hline
    \textcolor{black}{OSDFD (CLIP)} & \textcolor{black}{Ours} & \textcolor{black}{2.89M} & \textcolor{black}{\textbf{0.990}} & \textcolor{black}{\textbf{0.934}} & \textcolor{black}{\textbf{0.945}} & \textcolor{black}{\textbf{0.858}} & \textcolor{black}{\textbf{0.881}} & \textcolor{black}{\textbf{0.790}} & \textcolor{black}{\textbf{0.793}} & \textcolor{black}{\textbf{0.703}} & \textcolor{black}{\textbf{0.902}}  & \textcolor{black}{\textbf{0.821}}\\
    \hline
  \end{tabular}}
  \label{c23}
\end{table*}
\begin{table*}
  \centering
  \caption{Low quality (c40) cross-manipulation detection performance. (\#Params:  trainable parameter number)}
  \vspace{-2.5mm}
  \scalebox{0.92}{\begin{tabular}{l|c|c|cc|cc|cc|cc|cc}
    \hline
    \multirow{2}{*}{\textbf{Method}} & \multirow{2}{*}{\textbf{Venue}} & \multirow{2}{*}{\textbf{\#Params}} & \multicolumn{2}{c|}{\textbf{FF,FS,NT$\rightarrow$DF}} & \multicolumn{2}{c|}{\textbf{DF,FS,NT$\rightarrow$FF}} & \multicolumn{2}{c|}{\textbf{DF,FF,NT$\rightarrow$FS}} & \multicolumn{2}{c|}{\textbf{DF,FF,FS$\rightarrow$NT}} & \multicolumn{2}{c}{\textbf{Average}}\\ \cline{4-13}
     & & & AUC & ACC & AUC & ACC & AUC & ACC & AUC & ACC & AUC & ACC\\	
    \hline				
    ResNet18 \cite{he2016deep} & CVPR16 & 11.18M & 0.741 & 0.673 & 0.648 & 0.600 & 0.634 & 0.594 & 0.598 & 0.567 & 0.655 & 0.609 \\
    \hline
    Xception \cite{chollet2017xception} & CVPR17 & 20.81M & 0.766 & 0.694 & 0.696 & 0.643 & 0.626 & 0.593 & 0.597 & 0.552 & 0.671 & 0.621\\	
    \hline
    EN-B4 \cite{tan2019efficientnet} & ICML19 & 19.34M & 0.451 & 0.485 & 0.537  & 0.505& 0.512 & 0.503 & 0.499 & 0.497 & 0.500 & 0.498\\ 
    \hline
    All-train EN-B4 \cite{tan2019efficientnet} & ICML19 & 19.34M & 0.753 & 0.676 & 0.674 & 0.614 & 0.614 &  0.580 & 0.600 & 0.564 & 0.660 & 0.609 \\
    \hline
    Focal-loss EN-B4 \cite{tan2019efficientnet} & ICML19 & 19.34M & 0.749 & 0.674 & 0.672 & 0.610 & 0.596 & 0.575 & 0.605 & 0.566 & 0.656 & 0.606 \\ 				
    \hline
    Forensics Transfer \cite{cozzolino2018forensictransfer} & ArXiv18 & - & - & 0.682 & - & 0.550 & - & 0.530 & - & 0.550 & - & 0.578 \\ 		
    \hline
    Multi-task \cite{nguyen2019multi} & BTAS19 & - & - & 0.667 & - & 0.565 & - & 0.517 & - & 0.560 & - & 0.577 \\	
    \hline
    MLDG \cite{li2018learning} & AAAI18 & 62.38M & 0.730 & 0.671 & 0.617 & 0.581 & 0.617 &  0.581 & 0.607 & 0.569 & 0.643 & 0.601 \\
    \hline 				
    LTW \cite{sun2021domain} & AAAI21 & 20.37M & 0.756 & 0.691 & {0.724} & {0.657} & 0.681 & 0.625 & {0.608} & {0.585} & 0.692 & 0.640\\
    \hline
    CFM \cite{10315169} & TIFS24 & 25.37M & 0.789 & 0.686 & 0.622 & 0.541 & 0.688 & 0.607 & 0.566 & 0.517 & 0.666 & 0.588\\
    \hline
     \textcolor{black}{DCL} \cite{sun2022dual} & AAAI22 & 19.35M & 0.838 & 0.759 & \underline{0.751} & {0.679} & - & - & - & - & - & - \\  \hline
    \textcolor{black}{Implicit ID} \cite{huang2023implicit}& CVPR23 & 22.40M  & {0.846} & \underline{0.769} & {0.748} & {0.680} & - & - & - & - & - & - \\
    \hline
    \textcolor{black}{UCF} \cite{yan2023ucf} & CVPR23 & 20.81M & - & 0.746 & - & 0.673 & - & - & - & - & - & - \\
    \hline
    \textcolor{black}{DiffusionFake} \cite{sun2024diffusionfake} & NIPS24 & 19.34M & - & 0.751 & - & \textbf{0.713} & - & - & - & - & - & - \\
    \hline
    SCLoRA \cite{kong2023enhancing} & MIPR23 & 4.72M & {0.818} & {0.735} & 0.686 & 0.638 & {0.710} &  {0.653} & 0.582 & 0.542 & {0.699} & {0.642} \\ 	
    \hline
    ViT-B \cite{dosovitskiy2020image} & ICLR21 & 85.80M & 0.739 & 0.643 & 0.650 & 0.595 & 0.592 & 0.560 & 0.552 & 0.538 & 0.633 & 0.584 \\		
    \hline
    \textcolor{black}{CLIP} \cite{radford2021learning} & \textcolor{black}{ICML21} & \textcolor{black}{303.18M} & \textcolor{black}{0.800} & \textcolor{black}{0.707} & \textcolor{black}{0.676} & \textcolor{black}{0.615} & \textcolor{black}{0.651} & \textcolor{black}{0.606} & \textcolor{black}{0.592} & \textcolor{black}{0.544} & \textcolor{black}{0.680} & \textcolor{black}{0.618} \\		
    \hline									
    \hline
    OSDFD (ViT-B) & Ours & 1.34M & \underline{0.853} & {0.762} & {0.717} & {0.663} & \underline{0.713} & \underline{0.661} & \textbf{0.630} & \underline{0.595} & \underline{0.728} & \underline{0.670}\\
    \hline
    \textcolor{black}{OSDFD (CLIP)} & \textcolor{black}{Ours} & \textcolor{black}{2.89M} & \textcolor{black}{\textbf{0.924}} & \textcolor{black}{\textbf{0.819}} & \textcolor{black}{\textbf{0.761}} & \textcolor{black}{\underline{0.684}} & \textcolor{black}{\textbf{0.817}} & \textcolor{black}{\textbf{0.716}} & \textcolor{black}{\underline{0.628}} & \textcolor{black}{\textbf{0.601}} & \textcolor{black}{\textbf{0.783}} & \textcolor{black}{\textbf{0.705}} \\ 				
    \hline 				
  \end{tabular}}
  \label{c40}
\end{table*}

\subsection{Objective function}
During training, as we augment source-forgery features using the designed forgery style mixture module, we must train a classifier that can effectively distinguish between real face features and the augmented forgery features. Therefore, this work adopts the Single-Center Loss (SCL) \cite{sun2021domain} to obtain a clearer boundary between bonafide and fake faces. The overall objective function consists of two losses: 
\begin{equation}
     L = L_{BCE} + \lambda L_{SCL},
\end{equation}
$\lambda$ balances the two loss components. $L_{BCE}$ represents the binary cross-entropy loss. The Single-Center Loss, denoted as $L_{SCL}$, aims to compact the feature distribution of real faces while simultaneously moving fake features away from the center of real features: 
\begin{equation} 
    L_{SCL}=Dist_{R}+max(Dist_{R}-Dist_{F}+margin,0)
\end{equation}
\textcolor{black}{where margin (default: 0.01) is a hyperparameter that specifies how much farther the fake features should be from the real center compared to the real features.} $Dist_R$ and $Dist_F$ denote the average distance between the real feature center $C$ and real feature $f_{R}$ and fake feature $f_{F}$, respectively: 
\begin{equation} 
    Dist_{R}=\frac{1}{N_R}\sum_{{i=1}}^{N_R}||f_{R}^{i}-C||_{2};  \  Dist_{F}=\frac{1}{N_F}\sum_{{j=1}}^{N_F}||f_{F}^{j}-C||_{2}
\end{equation}
where the real center $C$ is the average of all $N_R$ real features $f_{R}$:
\begin{equation} 
    C=\frac{1}{N_R}\sum_{{i=1}}^{N_R}f_{R}^{i}.
\end{equation}

We select the features after the second-to-last fully-connected layer to calculate $L_{SCL}$. This loss enhances the discriminative and separable characteristics of the real and fake face features, thereby improving the overall performance of face forgery detection. \textcolor{black}{FSM is coupled with Single Center Loss ($L_{SCL}$) to jointly form a forgery-oriented regularization strategy. }

\section{Experiments and Results}
\subsection{Experimental setup}
\subsubsection{Implementation details}
\textcolor{black}{In this study, we employ the popular \texttt{dlib} face detector \cite{king2009dlib}. The detected face regions are enlarged by a factor of 1.3 and subsequently resized to 224×224 pixels before being input into the model.}
The proposed framework is implemented on the PyTorch \cite{paszke2019pytorch} platform, and the model is trained using Adam optimizer \cite{kingma2014adam} with $\beta_{1}$ = 0.9 and $\beta_{2}$ = 0.999. We set the learning rate at 3e-5, and no learning rate decay is applied throughout the whole training process. \textcolor{black}{In the proposed OSDFD framework, the ViT-Base model pre-trained on ImageNet-21K \cite{deng2009imagenet} serves as the backbone. It consists of 12 transformer blocks, each comprising a 12-head self-attention layer and a multi-layer perceptron (MLP) layer. We use the ViT-L/14 as the CLIP backbone.} \textcolor{black}{We set the batch size to 48, and the loss weight $\lambda$ in Eq. (10) is set to 1.} During training, we use AUC as the validation metric. The training process consists of 30,000 iterations, as we observe that the validation AUC converges after about 20,000 iterations.

\subsubsection{Datasets}
We adopt the FaceForensics++ (FF++) dataset \cite{rossler2019faceforensics++} for training, which includes original real videos (OR) and four types of forgeries: Deepfakes (DF) \cite{hm16_20}, Face2Face (FF) \cite{thies2016face2face}, FaceSwap (FS) \cite{dfcode}, and NeuralTextures (NT) \cite{thies2019deferred}. Each category comprises 1,000 videos in raw, high-quality (c23), and low-quality (c40) formats. Following the official data split strategy \cite{rossler2019faceforensics++}, we allocate 720 videos for training, 140 for validation, and 140 for testing. Raw videos represent non-compressed videos, whereas c23 and c40 videos denote slightly compressed and highly compressed videos. Our experiments exclusively employ the compressed c23 and c40 videos to align with practical applications.



\subsubsection{Evaluation Metrics} To facilitate head-to-head comparisons with previous methodologies, we employ Accuracy (ACC), Area Under the receiver operating characteristic Curve (AUC), and Equal Error Rate (EER) as the evaluation metrics in this work. ACC is a straightforward metric that quantifies the proportion of correctly classified samples from the total number of samples. AUC measures the Area under the Receiver Operating Characteristic (ROC) curve. EER denotes the point of ROC where False Positive Rate (FPR) is equal to the True Positive Rate (TPR). Unless specified otherwise, we report the image-level detection results throughout this paper.
\begin{table*}[]
  \caption{Cross-dataset evaluation on six unseen datasets. `*' indicates the trained model provided by the authors. `\dag' indicates our re-implementation using the public official code. Top and bottom tables show frame- and video-level results, respectively. The trainable parameter number of each model is also reported. We separate the CLIP-based methods from others.}
  \vspace{-2.5mm}
  \label{cross-dataset}
\centering
\scalebox{0.89}{\begin{tabular}{l|c|c|cc|cc|cc|cc|cc|cc|cc}
\hline
\multirow{2}{*}{\textbf{Method}} & \multirow{2}{*}{\textbf{Venue}}& \multirow{2}{*}{\textbf{\#Params}} & \multicolumn{2}{c|}{\textbf{CDF}} & \multicolumn{2}{c|}{\textbf{WDF}} & \multicolumn{2}{c|}{\textbf{DFDC-P}} & \multicolumn{2}{c|}{\textbf{DFDC}} & \multicolumn{2}{c|}{\textbf{DFR}} & \multicolumn{2}{c|}{\textbf{FFIW}} & \multicolumn{2}{c}{\textbf{Avgerage}}   \\ \cline{4-17} 
& & & AUC & EER & AUC & EER & AUC & EER & AUC & EER & AUC & EER & AUC & EER & AUC & EER\\ \hline
Face X-ray \cite{li2020face} & CVPR20 & 41.97M & 74.20 & - & - & - & 70.00 & - &  - & - & - & - & - & - & - & - \\
GFF \cite{luo2021generalizing} & CVPR21 & 53.25M & 75.31 & 32.48 & 66.51 & 41.52 & 71.58 & 34.77 &  - & -  & - & - & - & - & - & - \\
LTW \cite{sun2021domain} & AAAI21 & 20.37M & 77.14 & 29.34 & 67.12 & 39.22 & 74.58 & 33.81 & - & -  & - & - & - & - & - & - \\
F2Trans-S \cite{miao2023f} & TIFS23 & 117.52M & 80.72 & - & - & - & 71.71 & - & - & - & -  & - & - & - & - & -\\
SBI* \cite{shiohara2022detecting} & CVPR22 & 19.34M & {81.33} & 26.94 & 67.22 & 38.85 & \textbf{79.87} & \textbf{28.26} & 68.45 & 37.23 & 84.90 & 23.13 & 81.47 & 27.33 & 77.21 & 30.29 \\
DCL* \cite{sun2022dual} & AAAI22 & 19.35M & 81.05 & {26.76} & 72.95 & 35.73 & 71.49 & 35.90 & {72.69} & 34.91 & {92.26} & {14.81} & 82.08 & 27.75 & 78.75 & 29.31 \\
Xception\dag \cite{chollet2017xception} & ICCV19 & 20.81M & 64.14 & 39.77 & 68.90 & 38.67 & 69.56 & 36.94 & 64.52 & 39.29 & 91.93 & 15.52 & 77.58 & 31.21 & 72.77 & 33.57\\
RECCE\dag \cite{cao2022end} & CVPR22 & 25.83M & 61.42 & 41.71 & {74.38} & {32.64} & 64.08 & 40.04 & 67.61 & 36.98 & 92.93 & 14.74 & \underline{85.84} & {24.12} & 74.38 & 31.71 \\
EN-B4\dag \cite{tan2019efficientnet} & ICML19 & 19.34M & 65.24 & 39.41 & 67.89 & 37.21 & 67.96 & 37.60 & 63.48 & 40.14 & 92.18 & 15.51 & 70.03 & 35.40 & 71.13 & 35.40 \\
\textcolor{black}{Implicit ID} \cite{huang2023implicit}& CVPR23 & 22.36M  & \textbf{83.80} & 24.85 & - &- & - & - & \textbf{81.23} & \textbf{26.80} & - & - & - & - & - & - \\
\textcolor{black}{SFDG} \cite{wang2023dynamic}& CVPR23 & 22.40M & 75.83 & 30.30 & 69.27 & 37.70 & - & - & 73.64 & 33.70 & 92.10 & 15.10 & - & - & - & - \\
\textcolor{black}{UCF*} \cite{yan2023ucf} & CVPR23 & 46.80M & 77.15& 30.20 & 77.40  & 29.80 & 69.30 & 35.80 & 68.40 & 37.80 & 80.80 & 27.00 & 68.30 & 36.70 & 73.56 & 32.88 \\
\textcolor{black}{LSDA} \cite{yan2024transcending} & CVPR24 & 20.81M & 82.70 & - & 71.40 & - & 76.30 & - & 71.10 & - & 87.50 & - & 84.20 & - & 78.87 & - \\
\textcolor{black}{DiffusionFake} \cite{sun2024diffusionfake} & NIPS24 & 19.34M & {83.17} & \underline{24.59} & 75.17 & 33.25 & \underline{77.35} & \underline{30.17} & - & - & - & - & - & - & - & - \\
\textcolor{black}{DSRL*} \cite{cao2024towards} & IJCV24 & 128.31M & 74.23 & 30.48 & 74.18 & 31.16 & 72.01 & 34.03 &  {73.74} & 31.35 & 94.50 & 13.09 & \textbf{86.77} & \textbf{20.41} & 79.24 & 26.75 \\
\textcolor{black}{FreqNet\dag} \cite{tan2024frequency} & AAAI24 & 1.85M & 72.81 & 33.01 & 73.37 & 32.67 & 72.89 & 33.37 & 68.78 & 36.65 & 86.94 & 26.48 & 75.67 & 30.26 & 75.08 & 32.07 \\ 
CFM \cite{10315169} & TIFS24 & 25.37M & 82.78 & {24.74} & \underline{78.39} & {30.79} & {75.82} & {31.67} & 68.28 & 37.14 & \underline{95.18} & 11.87 & 78.71 & 29.28 & \underline{79.86} & 27.58 \\ 
\textcolor{black}{LAA-Net*} \cite{nguyen2024laa} & \textcolor{black}{CVPR24} & \textcolor{black}{27.12M} & \textcolor{black}{81.99} & \textcolor{black}{25.34} & \textcolor{black}{73.83} & \textcolor{black}{32.59} & \textcolor{black}{75.71} & \textcolor{black}{31.61} & \textcolor{black}{66.97} & \textcolor{black}{37.71} & \textcolor{black}{80.56} & \textcolor{black}{26.77} & \textcolor{black}{72.38} & \textcolor{black}{34.05} & \textcolor{black}{75.24}& \textcolor{black}{31.34} \\
ViT-B \cite{dosovitskiy2020image} & ICLR21 & 85.80M & 72.35 & 34.50 & 75.29 & 33.40 & 75.58 & 32.11 & 71.48 & 35.54 & 80.47 & 26.97 & 68.49 & 37.12 & 73.94 & 33.27 \\ 
SCLoRA \cite{kong2023enhancing}  & MIPR23 & 4.72M & 73.56 & 32.87 & 77.91 & \underline{29.56} & {77.08} & {30.37} & 72.32 & {34.52} & 94.78 & \underline{11.76} & 81.42 & 26.59 & 79.51 & \underline{26.59} \\ 	
\hline
OSDFD (ViT-B) & Ours & 1.34M & \underline{83.35} & \textbf{24.47} & \textbf{78.53} & \textbf{28.46} & 76.28 & 31.17 & \underline{75.52} & \underline{31.99} & \textbf{95.54} & \textbf{9.90} & {84.61} & \underline{24.09} & \textbf{82.31} & \textbf{25.01} \\
\hline  \hline
\textcolor{black}{ForAda*} \cite{cui2025forensics} & \textcolor{black}{CVPR25} & \textcolor{black}{7.54M} & \textcolor{black}{\underline{87.61}} & \textcolor{black}{\underline{21.71}} & \textcolor{black}{\underline{79.90}} & \textcolor{black}{\underline{27.60}} & \textcolor{black}{\textbf{83.87}} & \textcolor{black}{\textbf{24.74}} & \textcolor{black}{\textbf{79.42}} & \textcolor{black}{\textbf{28.08}} & \textcolor{black}{93.51} & \textcolor{black}{13.96} & \textcolor{black}{77.66} & \textcolor{black}{29.62} & \textcolor{black}{\underline{83.67}}& \textcolor{black}{\underline{24.29}} \\
\textcolor{black}{UDD*} \cite{fu2025exploring} & \textcolor{black}{AAAI25} & \textcolor{black}{0.66M} & \textcolor{black}{82.68} & \textcolor{black}{25.20} & \textcolor{black}{75.92} & \textcolor{black}{31.31} & \textcolor{black}{\underline{79.56}} & \textcolor{black}{\underline{27.92}} & \textcolor{black}{73.67} & \textcolor{black}{32.42} & \textcolor{black}{90.93} & \textcolor{black}{16.84} & \textcolor{black}{78.79} & \textcolor{black}{28.10} & \textcolor{black}{80.26}& \textcolor{black}{26.97} \\ 
\textcolor{black}{CLIP} \cite{radford2021learning}  & \textcolor{black}{ICML21} & \textcolor{black}{303.18M} & \textcolor{black}{81.23} & \textcolor{black}{26.46} & \textcolor{black}{76.41} & \textcolor{black}{31.55} & \textcolor{black}{77.97} & \textcolor{black}{29.76} & \textcolor{black}{72.54} & \textcolor{black}{33.97} & \textcolor{black}{\underline{95.16}} & \textcolor{black}{\underline{11.69}} & \textcolor{black}{\underline{83.63}} & \textcolor{black}{\underline{25.19}} & \textcolor{black}{81.15}& \textcolor{black}{24.77} \\  \hline
\textcolor{black}{OSDFD (CLIP)} & \textcolor{black}{Ours} & \textcolor{black}{2.89M} & \textcolor{black}{\textbf{88.32}} & \textcolor{black}{\textbf{20.45}} & \textcolor{black}{\textbf{85.39}} & \textcolor{black}{\textbf{21.97}} & \textcolor{black}{79.42} & \textcolor{black}{30.00} & \textcolor{black}{\underline{77.88}} & \textcolor{black}{\underline{30.15}} & \textcolor{black}{\textbf{99.17}} & \textcolor{black}{\textbf{3.46}} & \textcolor{black}{\textbf{93.71}} & \textcolor{black}{\textbf{14.88}} & \textcolor{black}{\textbf{87.32}} & \textcolor{black}{\textbf{20.15}} \\ \hline
\midrule[1.5pt]
{Lisiam} \cite{wang2022lisiam} & TIFS22 & - & 78.21 & - & - & - & - & - & - & - & - & - & - & - & - & - \\
{F3Net} \cite{qian2020thinking} & ECCV20 & 42.53M & 68.69 & - & - & - & 67.45 & - & - & - & - & - & - & - & - & - \\
{FTCN} \cite{zheng2021exploring} & ICCV21 & 26.60M & 86.90 & - & - & - & 74.00 & - & - & - & - & - & - & - & - & - \\
{SBI}* \cite{shiohara2022detecting} & CVPR22 & 19.34M & {88.61} & 19.41 & 70.27 & 37.63 & \textbf{84.80} & \textbf{25.00} & 71.70 & 35.27 & 90.04 & 17.53 & {86.34} & {23.45} & 81.96 & 26.38 \\
{DCL}* \cite{sun2022dual} & AAAI22 & 19.35M & 88.24 & {19.12} & 76.87 & 31.44 & 77.57 & 29.55 & {75.03} & {30.94} & {97.41} & 9.96 & 86.26 & 24.32 & 83.56 & 24.22 \\
{RECCE}\dag \cite{cao2022end} & CVPR22 & 25.83M & 69.25 & 34.38 & {76.99} & {30.49} & 66.90 & 39.39 & 70.26 & 34.25 & {97.15} & {9.29} & \textbf{92.05} & \underline{17.74} & 78.77 & 27.59 \\
{UCF* \cite{yan2023ucf}} & CVPR23 & 46.80M & 83.70  & 28.20 & 77.20  & 26.10 & 71.00  & 33.50 & 70.90  & 35.80 &  86.40 & 20.40 & 69.70  & 34.20 & 76.48 & 29.70 \\
{LSDA \cite{yan2024transcending}} & CVPR24 & 20.81M & 89.80 & - & 75.60 & - & \underline{81.20} & - & 73.50 & - & 89.20 & - & 84.20 & - & 82.25 & - \\
{DSRL* \cite{cao2024towards}} & IJCV24 & 128.31M & 84.56 & 22.90 & 77.31 & 27.32 & 75.64 & 30.17 & \underline{77.99} & \textbf{27.63} & 97.23 & 10.44 & 88.69 & 19.82 & 83.57 & \underline{23.05} \\
{FreqNet\dag \cite{tan2024frequency}} & AAAI24 & 1.85M & 75.79 & 28.59 & 79.54 & 29.14 & 75.07 & 31.16 &  69.51 &  36.11 &  88.79 & 25.97 & 79.56 & 26.47 & 78.04 & 29.57 \\ CFM \cite{10315169} & TIFS24 & 25.37M & {89.65} & {17.65} & \underline{82.27} & 26.80 & {80.22} & 27.48 & 70.59 & 35.02 & {97.59} & {9.04} & 83.81 & 24.49 & \underline{84.02} & {23.41} \\
\textcolor{black}{LAA-Net*} \cite{nguyen2024laa} & \textcolor{black}{CVPR24} & \textcolor{black}{27.12M} & \textcolor{black}{\textbf{91.99}} & \textcolor{black}{\textbf{15.73}} & \textcolor{black}{78.84} & \textcolor{black}{27.78} & \textcolor{black}{78.78} & \textcolor{black}{28.72} & \textcolor{black}{70.28} & \textcolor{black}{35.26} & \textcolor{black}{87.86} & \textcolor{black}{19.90} & \textcolor{black}{77.97} & \textcolor{black}{29.75} & \textcolor{black}{80.95}& \textcolor{black}{26.19} \\ 
{ViT-B} \cite{dosovitskiy2020image} & ICLR21 & 85.80M & 78.12 & 30.59 & 78.95 & 29.59 & 79.43 & 28.62 & 73.35 & 33.05 & 86.42 & 19.90 & 70.34 & 35.45 & 77.77 & 29.53 \\ 
{SCLoRA} \cite{kong2023enhancing} & MIPR23 & 4.72M & 81.11 & 26.32 & \textbf{82.78} & \textbf{24.85} & 80.20 & \underline{26.68} & 74.96 & 35.01 & \underline{98.53} & \underline{6.63} & 85.59 & 22.70 & 83.86 & 23.70 \\ \hline
{OSDFD (ViT-B)} & Ours & 1.34M & \underline{90.81} & \underline{16.94} & 81.42 & \underline{24.90} & 79.49 & 28.31 & \textbf{78.74} & \underline{29.37} & \textbf{98.89} & \textbf{4.48} & \underline{90.01} & \textbf{17.50} & \textbf{86.56} & \textbf{20.25} \\
\hline  \hline
\textcolor{black}{ForAda*} \cite{cui2025forensics} & \textcolor{black}{CVPR25} & \textcolor{black}{7.54M} & \textcolor{black}{\textbf{93.87}} & \textcolor{black}{\textbf{14.41}} & \textcolor{black}{\underline{85.64}} & \textcolor{black}{\underline{22.47}} & \textcolor{black}{\textbf{91.40}} & \textcolor{black}{\textbf{16.53}} & \textcolor{black}{\textbf{83.77}} & \textcolor{black}{\textbf{23.51}} & \textcolor{black}{97.48} & \textcolor{black}{8.46} & \textcolor{black}{85.12} & \textcolor{black}{23.22} & \textcolor{black}{\underline{89.55}} & \textcolor{black}{\underline{18.10}} \\
\textcolor{black}{UDD*} \cite{fu2025exploring} & \textcolor{black}{AAAI25} & \textcolor{black}{0.66M} & \textcolor{black}{90.31} & \textcolor{black}{17.29} & \textcolor{black}{78.90} & \textcolor{black}{32.20} & \textcolor{black}{\underline{84.97}} & \textcolor{black}{\underline{23.76}} & \textcolor{black}{77.14} & \textcolor{black}{29.85} & \textcolor{black}{96.17} & \textcolor{black}{11.44} & \textcolor{black}{85.54} & \textcolor{black}{23.51} & \textcolor{black}{85.51}& \textcolor{black}{23.01} \\  
\textcolor{black}{CLIP} \cite{radford2021learning}  & \textcolor{black}{ICML21} & \textcolor{black}{303.18M} & \textcolor{black}{88.87} & \textcolor{black}{21.47} & \textcolor{black}{80.33} & \textcolor{black}{28.54} & \textcolor{black}{80.17} & \textcolor{black}{27.76} & \textcolor{black}{76.07} & \textcolor{black}{30.94} & \textcolor{black}{\underline{98.67}} & \textcolor{black}{\underline{6.97}} & \textcolor{black}{\underline{90.91}} & \textcolor{black}{\underline{17.22}} & \textcolor{black}{85.83}& \textcolor{black}{22.15} \\
\hline 
\textcolor{black}{OSDFD (CLIP)} &\textcolor{black}{Ours} & \textcolor{black}{2.89M} & \textcolor{black}{\underline{93.23}} & \textcolor{black}{\underline{16.29}} & \textcolor{black}{\textbf{87.16}} & \textcolor{black}{\textbf{19.51}} & \textcolor{black}{83.13} & \textcolor{black}{25.76} & \textcolor{black}{\underline{80.65}} & \textcolor{black}{\underline{27.53}} & \textcolor{black}{\textbf{99.62}} & \textcolor{black}{\textbf{1.49}} & \textcolor{black}{\textbf{97.20}} & \textcolor{black}{\textbf{9.76}} & \textcolor{black}{\textbf{90.17}} & \textcolor{black}{\textbf{16.72}} \\ \hline 
\end{tabular}} 								
\end{table*}
\vspace{-6mm}
\subsection{Cross-manipulation evaluation}
In real-world applications, face forgery detectors inevitably encounter data generated by new forgery techniques, leading to significant performance drops. To evaluate the generalization capability, we train the models on the training sets comprising three source manipulation types of FF++. Subsequently, we evaluate the trained models on the test sets of the unseen target manipulation type. Table~\ref{c23} presents the detection AUC and ACC of different models on the FF++ high-quality (c23) dataset. We bold the best detection results and underline the second-best results. \textcolor{black}{We apply OSDFD to two backbone networks, ViT-B and CLIP (ViT-L/14). The results show that OSDFD built on ViT-B surpasses previous methods, while OSDFD (CLIP) achieves SOTA AUC and ACC across all evaluation settings.}

Moreover, generalization to poor-quality data is crucial for real-world face forgery detectors. Therefore, we conduct the cross-manipulation experiments on low-quality (c40) data, as shown in Table~\ref{c40}, where OSDFD outperforms previous approaches by a clear margin under three settings. This underscores the proposed method's outstanding generalizability across diverse data qualities. \textcolor{black}{OSDFD exhibits relatively lower performance on the (DF, FS, NT $\rightarrow$ FF) setting under low-quality cross-manipulation evaluation. This may be because the PEFT framework and style mixture module are well suited for capturing appearance-related anomalies but are less effective at generalizing to manipulations dominated by expression and motion artifacts, thereby limiting transferability to FF.} Additionally, our approach demonstrates substantial improvements over the ViT-B baseline, demonstrating the effectiveness of the designed framework. Furthermore, compared to our conference version SCLoRA \cite{kong2023enhancing}, our method achieves superior performance on both c23 and c40 data, which can be attributed to the effectiveness of the designed forgery style mixture module and local CDC feature extractor. \textcolor{black}{Moreover, OSDFD (CLIP) exhibits good generalizability on low-quality data while requiring only 2.89M trainable parameters, making it well suited for deployment in open-set scenarios.}

\begin{table*}
  \caption{\textcolor{black}{Cross-dataset evaluations on DF40 (CDF) and FakeAVCeleb datasets.}}
  \vspace{-2.5mm}
  \label{df40-data2}
  \centering
  \renewcommand\arraystretch{1.15}
  \scalebox{0.85}{\textcolor{black}{\begin{tabular}{ccccccccccc}
\hline
\multirow{2}{*}{Method}  & \multicolumn{2}{c}{FS (CDF)} & \multicolumn{2}{c}{FR (CDF)} & \multicolumn{2}{c}{EFS (CDF)} & \multicolumn{2}{c}{FakeAVCeleb} & \multicolumn{2}{c}{Average}\\ \cline{2-11} 
 & AUC & EER & AUC & EER & AUC & EER & AUC & EER & AUC & EER\\ \hline
SBI\cite{shiohara2022detecting}  & 70.01 & 35.20 & 48.90 & 50.98 & 65.61 & 39.21 & 85.87 & 23.07 & 67.59 & 37.11\\ \hline			
CFM\cite{10315169}   & \underline{78.65} & \underline{28.73} & \textbf{69.11} & \textbf{35.96} & \underline{71.49} & \underline{34.33} & 82.84&  26.27& \underline{75.52} & \underline{31.32}\\ \hline
DCL\cite{sun2022dual} & 72.67 & 35.04 & 62.88 & 41.01 & 63.18 & 41.20 &  \underline{87.64} & \underline{21.00} & 71.59 & 34.56\\ \hline
LAA-Net\cite{nguyen2024laa} & 70.46 & 34.99 & 55.24 & 46.91 & 67.07 & 38.07 &  \textbf{88.20} & \textbf{19.74} & 70.24 & 34.92\\ \hline
DiffusionFake\cite{sun2024diffusionfake} & 70.34 & 35.75 & 55.37 & 46.85 & 63.70 & 41.57 & 85.91 & 22.54 & 68.83 & 36.67 \\ \hline
ViT-B \cite{dosovitskiy2020image} & 67.78 & 37.19 & 55.65 & 46.57 & 60.50 & 42.64 & 82.98&  26.14&  66.72 & 38.13\\ \hline
OSDFD (ViT-B) & \textbf{82.50} & \textbf{24.59} & \underline{64.78} & \underline{39.22} & \textbf{77.68} & \textbf{26.65} & {84.04} & {23.03} & \textbf{77.25} & \textbf{28.37} \\ \hline \hline
CLIP \cite{radford2021learning} & 78.93 & 29.21 & 62.33 & 42.49 & 71.23 & 34.10 & 82.86&  25.27&  73.83 & 32.76\\ \hline
ForAda\cite{cui2025forensics}   & \underline{83.33} & \underline{25.13} & \underline{63.76} & \underline{40.73} & \underline{82.30} & \underline{25.34} & 83.67& 25.41&  \underline{78.26} & \underline{29.15}\\ \hline
UDD\cite{fu2025exploring} & 61.44 & 41.64 & 59.01 & 43.63 & 52.64 & 47.54 & \underline{90.72} & \underline{16.02} &  65.95 & 37.20 \\ \hline
OSDFD (CLIP) & \textbf{90.77} & \textbf{16.97} & \textbf{71.06} & \textbf{35.00} & \textbf{90.09} & \textbf{18.70} & \textbf{92.95} & \textbf{14.54} & \textbf{86.21}  & \textbf{21.30}\\
\hline
\end{tabular}}}
\end{table*}

\begin{table*}
  \caption{\textcolor{black}{Efficiency evaluations w/ and w/o OSDFD.}}
  \label{efficiency2}
  \vspace{-2.5mm}
  \centering
  \renewcommand\arraystretch{1.15}
  \scalebox{0.85}{\textcolor{black}{\begin{tabular}{cc|ccc|c|c|c}
\hline
    \multirow{2}{*}{Backbone} & \multirow{2}{*}{OSDFD} & Activated & Training & Training GPU & Inference & \multirow{2}{*}{FLOPs (G)$\downarrow$} & AVG. \\ 
     &  & \# Para.(M)$\downarrow$ & Speed (Iter/s)$\uparrow$ & Memory (GB)$\downarrow$ & Time (ms)$\downarrow$ &  & AUC$\uparrow$ \\ \hline
 ViT-B & - & 85.80 & 6.53$\pm$0.01 &  4.58 & 5.05$\pm$0.02 & 17.58 & 73.94 \\ \hline	
\cellcolor[HTML]{E0DBDB}ViT-B & \cellcolor[HTML]{E0DBDB}\Checkmark  & \cellcolor[HTML]{E0DBDB}1.34  & \cellcolor[HTML]{E0DBDB}7.14$\pm$0.01  & \cellcolor[HTML]{E0DBDB}3.55 & \cellcolor[HTML]{E0DBDB}6.36$\pm$0.01 & \cellcolor[HTML]{E0DBDB}17.74 & \cellcolor[HTML]{E0DBDB}82.31 \\ 
 \cellcolor[HTML]{ADD8E6} & \cellcolor[HTML]{ADD8E6} & \cellcolor[HTML]{ADD8E6}(\textbf{98.44\%} \bm{$\downarrow$}) & \cellcolor[HTML]{ADD8E6}(\textbf{9.34\%} \bm{$\uparrow$})  & \cellcolor[HTML]{ADD8E6}(\textbf{22.49\%} \bm{$\downarrow$}) & \cellcolor[HTML]{ADD8E6}(\textbf{+1.31}) & \cellcolor[HTML]{ADD8E6}(\textbf{+0.16}) & \cellcolor[HTML]{ADD8E6}(\textbf{+8.37}) \\ \hline
 CLIP & - & 303.18& 0.83$\pm$0.01 & 19.59 & 
28.57$\pm$0.03 & 81.08 &81.15\\ \hline 
\cellcolor[HTML]{E0DBDB}CLIP & \cellcolor[HTML]{E0DBDB}\Checkmark & \cellcolor[HTML]{E0DBDB}2.89  & \cellcolor[HTML]{E0DBDB}1.00$\pm$0.01 & \cellcolor[HTML]{E0DBDB}18.04 & \cellcolor[HTML]{E0DBDB}32.25$\pm$0.02 & \cellcolor[HTML]{E0DBDB}81.79 & \cellcolor[HTML]{E0DBDB}87.32  \\ 
\cellcolor[HTML]{ADD8E6}& \cellcolor[HTML]{ADD8E6} & \cellcolor[HTML]{ADD8E6}(\textbf{99.05\%} \bm{$\downarrow$}) & \cellcolor[HTML]{ADD8E6}(\textbf{20.48\%} \bm{$\uparrow$})  & \cellcolor[HTML]{ADD8E6}(\textbf{7.91\%} \bm{$\downarrow$}) & \cellcolor[HTML]{ADD8E6}(\textbf{+3.68}) & \cellcolor[HTML]{ADD8E6}(\textbf{+0.71}) & \cellcolor[HTML]{ADD8E6}(\textbf{+6.17}) \\ \hline
\end{tabular}}} 
\end{table*}
\subsection{Cross-dataset evaluation}
Due to the significant domain gaps between training and testing datasets \cite{10138555, cai2023glitch} in face forgery detection, cross-dataset evaluation has been a persistent challenge. We further conduct cross-destatet evaluations to assess the generalizability of forgery detectors. \textcolor{black}{In cross-dataset evaluation, our method OSDFD adopts the FF++ (c23) dataset \cite{rossler2019faceforensics++} as the training set for fair comparisons. Our training set includes original real faces and four types of forgeries: Deepfakes, Face2Face, FaceSwap, and NeuralTextures, and the four manipulation types are randomly mixed during the forgery style mixture.} To balance positive and negative samples during training, real faces are augmented fourfold.
We test the trained models on six popular Deepfake datasets, including CelebDF-v2 (CDF) \cite{li2020celeb}, WildDeepfake (WDF) \cite{zi2020wilddeepfake}, DeepFake Detection Challenge (DFDC) \cite{dolhansky2019deepfake}, DeepFake Detection Challenge Preview (DFDC-P) \cite{dolhansky2019deepfake}, DeepForensics-1.0 (DFR) \cite{jiang2020deeperforensics}, and Face Forensics in the Wild (FFIW) \cite{Zhou_2021_CVPR}. These testing datasets simulate real-world scenarios well, encompassing various environmental variables. AUC and EER performance are reported in Table~\ref{cross-dataset}, where `*' indicates the trained model provided by the authors, and `\dag' indicates our re-implementation using the public official code. 
The upper table presents the frame-level detection results, while the bottom table shows video-level results. Video-level detection scores are calculated by averaging all frame scores within each video. \textcolor{black}{Since CLIP serves as a more powerful backbone for face forgery detection, we separate the CLIP-based methods (ForAda \cite{cui2025forensics}, UDD \cite{fu2025exploring}, CLIP \cite{radford2021learning}, and OSDFD (CLIP)) from the other approaches to ensure fair comparisons.}
The best and second-best results are highlighted in bold and underlined, respectively. The trainable parameter number of each detector is provided in the table for reference. \textcolor{black}{OSDFD (ViT-B) achieves promising average detection performance in both frame-level and video-level evaluations. In contrast, OSDFD built on the CLIP backbone attains better AUC and EER, as CLIP’s multimodal pretraining enables it to learn representations that align visual and semantic concepts, thereby enhancing generalization to distribution shifts.}
\textcolor{black}{OSDFD’s relatively lower performance on DFDC and DFDC-P can be attributed to their large domain gaps with FF++ and the suppression of forgery artifacts by compression, which pose significant challenges for PEFT-based methods.} Compared to the ViT-B and CLIP backbones, OSDFD demonstrates significant performance improvements across all six datasets, demonstrating its exceptional generalization capability. 
Furthermore, OSDFD outperforms our conference version SCLoRA by a significant margin, underscoring the effectiveness of the designed CDC Adapter layer and forgery-style-mixture module.

\textcolor{black}{We expand our evaluation to two other recently released and challenging datasets: DF40 \cite{yan2024df40} and FakeAVCeleb \cite{khalid2021fakeavceleb}. DF40 encompasses 40 diverse face forgery methods. Its domain-specific fake data generated from FF++ and CDF real faces. FakeAVCeleb is a multimodal benchmark comprising 20,000 videos, built from real YouTube celebrity footage across four ethnic backgrounds. We train our models on FF++ and report frame-level cross-dataset evaluations in Table~\ref{df40-data2}. To prevent facial identity leakage, we test on three challenging DF40 subsets (FS (CDF), FR (CDF), and EFS (CDF)) derived from CDF real faces, and additionally evaluate on FakeAVCeleb (visual modality only). The upper part of Table~\ref{df40-data2} shows results with conventional backbones, while the lower part reports CLIP-based models. Despite the sophistication of the testing datasets, OSDFD (ViT-B) achieves the best average AUC in the upper table, while OSDFD (CLIP) consistently outperforms all compared methods across all subsets, demonstrating superior generalizability on recent challenging datasets. The superior performance of CLIP-based models compared to ViT-B arises from CLIP’s large-scale multimodal pretraining, which equips it with stronger generalization ability to unseen forgery patterns. }

\subsection{Robustness evaluation}
\begin{figure*}[ht]
\centering
\includegraphics[scale=0.85]{  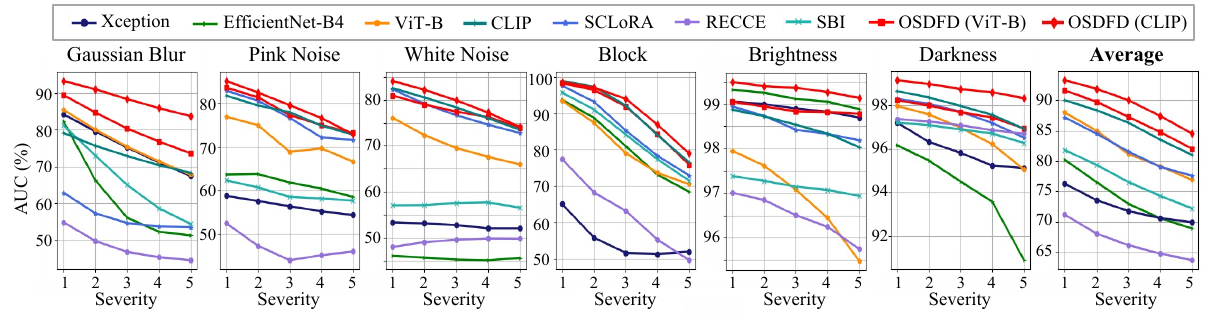}
  \vspace{-3.5mm}
\caption{Robustness to common perturbations at five severity levels: Gaussian Blur, Pink Noise, White Noise, Block, Brightness, and Darkness.}
\label{tmm_robustness}
\end{figure*}

We incorporate five severity levels for each perturbation type to mimic unpredictable environments. The robustness evaluation results are depicted in Fig.~\ref{tmm_robustness}. OSDFD does not employ data augmentations during training, facilitating efficient training, or fine-tuning for real-world applications. All models depicted in Fig.~\ref{tmm_robustness} are trained on the FF++ c23 training set, with distortions applied to the corresponding test set. As the severity of distortion increases, all face forgery detection methods experience performance degradation. \textcolor{black}{The average robustness results across the six distortion types indicate that both OSDFD (ViT-B) and OSDFD (CLIP) exhibit superior resilience to common image perturbations.}
\textcolor{black}{OSDFD (ViT) demonstrates a lower robustness than EfficientNet-B4 under brightness perturbations. A potential reason lies in our use of the PEFT strategy. Pretrained knowledge captures real visual statistics (e.g., natural lighting), whereas forgeries frequently violate these subtle priors (e.g., illumination inconsistencies), leaving detectable forgery cues. However, brightness perturbations during robustness evaluation can obscure these artifacts, thereby diminishing our method’s performance. The designed FSM module enhances robustness by generating broader feature distributions to mitigate overfitting, introducing variability that improves resistance to perturbations, and guiding the model to focus on intrinsic forgery cues rather than superficial styles.}

\begin{table}
  \centering
  \caption{Ablation Experiments on different model components. Average Cross-Manipulation detection performance on FF++ c23 dataset. }
  \vspace{-0.3cm}
  \scalebox{1.0}{\begin{tabular}{ccccccc}
    \hline
    ViT & LoRA & Adapter & MixStyle & SCL & AUC & ACC \\	
    \hline				
    \Checkmark & - & - & - & - & 0.623 & 0.575 \\
    \hline
    \Checkmark & \Checkmark & - & - & - & 0.816 & 0.719 \\
    \hline
    \Checkmark & - & \Checkmark & - & - & 0.825 & 0.723 \\
    \hline
    \Checkmark & - & - & \Checkmark & - & 0.795 & 0.712 \\
    \hline
    \Checkmark & \Checkmark & \Checkmark & - & - & 0.832 & 0.740 \\
    \hline
    \Checkmark & \Checkmark & \Checkmark & \Checkmark & - & 0.836 & 0.743 \\
    \hline
    \Checkmark & \Checkmark & \Checkmark & \Checkmark & \Checkmark & \textbf{0.843} & \textbf{0.752}\\
    \hline
  \end{tabular}}
  \label{Ablation}
\end{table}

\subsection{\textcolor{black}{Efficiency evaluations}}
\textcolor{black}{OSDFD demonstrates high training efficiency and strong generalization capability with only negligible additional FLOPs. By introducing a small number of trainable parameters, OSDFD achieves lower storage requirements, faster training, and reduced GPU memory consumption compared with the vanilla backbones.}

\textcolor{black}{We conducted comprehensive efficiency evaluations of OSDFD using two backbones, ViT-B and CLIP, as presented in Table~\ref{efficiency2}. All experiments were implemented on an NVIDIA A40 GPU and an Intel(R) Xeon(R) Gold 6530 CPU. The training speed and inference time were measured over five trials, while the training speed and GPU memory were recorded with a batch size of 32. For reference, we also report the average AUC (AVG. AUC), representing the mean performance across six unseen datasets.}

\textcolor{black}{Compared with the vanilla ViT-B and CLIP backbones, OSDFD demonstrates strong parameter efficiency by introducing only a small number of trainable parameters. This design results in {lower storage requirements}, as the backbone weights can be stored in the cloud while only minimal parameters are retained locally. Moreover, OSDFD enables {faster training} because fewer parameters are updated during optimization. As shown in Table~\ref{efficiency2}, OSDFD improves training speed by 9.34\% and 20.48\% over the ViT-B and CLIP baselines, respectively. This efficiency makes OSDFD well-suited for frequent model updates, such as continual learning with newly emerging forgery data in open-set Deepfake detection. Furthermore, OSDFD effectively {reduces GPU memory consumption} during training, making it practical for deployment on resource-constrained devices.}

\textcolor{black}{The designed PEFT and forgery-style-mixture components introduce only a marginal increase in model size, leading to negligible additional inference time and computational overhead. Specifically, OSDFD incurs only 1.31 ms and 3.68 ms inference time, and introduces merely 0.16 G and 0.71 G additional FLOPs for the ViT-B and CLIP backbones, respectively, which are acceptable for real-world applications.}

\textcolor{black}{Beyond training efficiency, OSDFD provides additional benefits. The forgery-aware PEFT and forgery-style-mixture modules effectively integrate forgery-specific knowledge while preserving the pretrained model’s general representations. This design mitigates overfitting to forgery data and enhances generalization performance. }

\subsection{Ablation study}
In this subsection, we carry out extensive ablation experiments to showcase the effectiveness of the designed components, the flexibility of the proposed learning scheme, the rationale behind the adopted training strategy, and the impacts of the mixed features.  

\subsubsection{Effectiveness of the designed modules}
We conduct ablation experiments to study the impact of each designed module. Table~\ref{Ablation} reports the average cross-manipulation (c23) detection AUC and ACC scores across four experimental trials. The designed LoRA and Adapter modules significantly enhance the model's generalizability. The combination of LoRA and Adapter enables the model to simultaneously explore global and local anomalies within input forgery face images, enhancing the final detection performance. The forgery-style-mixture module further boosts the AUC and ACC scores by augmenting the forgery source domains during training. The Single-Center Loss (SCL) encourages the real features to cluster more tightly while pushing features from fake images away from the real center. This design leads to a clearer decision boundary between real and fake faces, further enhancing the generalizability of the detection. Overall, the experimental results presented in Table~\ref{Ablation} underscore the effectiveness of the designed components.

\textcolor{black}{The proposed method showcases its adaptability by being flexibly adapted to other transformer-based backbones.} OSDFD comprises forgery-aware, parameter-efficient fine-tuning modules and a forgery-style-mixture module, both designed as plug-and-play components that improve the generalization capability of transformer backbones. As shown in Table~\ref{Backbone}, we apply our method OSDFD to various models, including ViT \cite{dosovitskiy2020image} and Swin \cite{liu2021swin}. The results demonstrate that OSDFD enhances performance across these backbones, leading to significant improvements in AUC scores in the cross-manipulation setting. Notably, the highest performance gains are observed in ViT-L (+0.167 AVG AUC) and Swin-L (+0.112 AVG AUC), suggesting that OSDFD is particularly effective when integrated with larger backbones. While the average cross-manipulation performance improves with increasing model capacity, larger models tend to require significantly more computing resources and inference time, thereby impeding their real-world deployment. These findings indicate that OSDFD is a flexible and effective module for improving the generalizability of transformer-based face forgery detection models.



\begin{table}
  \centering
  \caption{\textcolor{black}{Cross-manipulation AUC performance on FF++ (c23) dataset with different backbones.}}
  \vspace{-2.5mm}
  \scalebox{0.9}{\begin{tabular}{ccccccc}
    \hline
    Backbone & OSDFD & DF & FF & FS & NT & AVG \\	
    \hline				
    ViT-T & - & 0.853 &  0.658 & 0.488 & 0.584 & 0.646\\
    \hline
    \cellcolor[HTML]{E0DBDB}ViT-T & \cellcolor[HTML]{E0DBDB}\Checkmark & \cellcolor[HTML]{E0DBDB}0.919 & \cellcolor[HTML]{E0DBDB}0.860 & \cellcolor[HTML]{E0DBDB}0.563 & \cellcolor[HTML]{E0DBDB}0.709 & \cellcolor[HTML]{E0DBDB}0.763 \\
    \hline
    ViT-S & - & 0.870 & 0.688 & 0.465 & 0.566 & 0.647 \\
    \hline
    \cellcolor[HTML]{E0DBDB}ViT-S & \cellcolor[HTML]{E0DBDB}\Checkmark & \cellcolor[HTML]{E0DBDB}0.929 & \cellcolor[HTML]{E0DBDB}0.897 & \cellcolor[HTML]{E0DBDB}0.600 & \cellcolor[HTML]{E0DBDB}0.791 & \cellcolor[HTML]{E0DBDB}0.804 \\
    \hline
    ViT-L & - & 0.850 & 0.708 & 0.572 & 0.593 & 0.681 \\
    \hline
    \cellcolor[HTML]{E0DBDB}ViT-L & \cellcolor[HTML]{E0DBDB}\Checkmark & \cellcolor[HTML]{E0DBDB}0.955 & \cellcolor[HTML]{E0DBDB}0.941 & \cellcolor[HTML]{E0DBDB}0.710 & \cellcolor[HTML]{E0DBDB}0.785 & \cellcolor[HTML]{E0DBDB}0.848\\
    \hline
    \hline
    Swin-S & - & 0.897 & 0.768 & 0.400 & 0.666 & 0.683 \\
    \hline
    \cellcolor[HTML]{E0DBDB}Swin-S & \cellcolor[HTML]{E0DBDB}\Checkmark & \cellcolor[HTML]{E0DBDB}0.922 & \cellcolor[HTML]{E0DBDB}0.867 & \cellcolor[HTML]{E0DBDB}0.538 & \cellcolor[HTML]{E0DBDB}0.759 & \cellcolor[HTML]{E0DBDB}0.772\\
    \hline
    Swin-B & - & 0.877 & 0.715 & 0.463 & 0.668 & 0.681 \\
    \hline 				
    \cellcolor[HTML]{E0DBDB}Swin-B & \cellcolor[HTML]{E0DBDB}\Checkmark & \cellcolor[HTML]{E0DBDB}0.943 & \cellcolor[HTML]{E0DBDB}0.888 & \cellcolor[HTML]{E0DBDB}0.583 & \cellcolor[HTML]{E0DBDB}0.734 & \cellcolor[HTML]{E0DBDB}0.787\\
    \hline 								
    Swin-L & - & 0.923 & 0.743 & 0.446 & 0.646 & 0.690 \\
    \hline
    \cellcolor[HTML]{E0DBDB}Swin-L & \cellcolor[HTML]{E0DBDB}\Checkmark & \cellcolor[HTML]{E0DBDB}0.956 & \cellcolor[HTML]{E0DBDB}0.906 & \cellcolor[HTML]{E0DBDB}0.601 & \cellcolor[HTML]{E0DBDB}0.745 & \cellcolor[HTML]{E0DBDB}0.802\\
    \hline
  \end{tabular}}
  \label{Backbone}
\end{table}

In this work, we adopt a Parameter-Efficient Fine-Tuning (PEFT) strategy, a reliable approach, to train the face forgery detection model. Specifically, we only optimize the Adapter and LoRA layers while keeping the other ViT parameters fixed with ImageNet weights. We compare the PEFT strategy with the full fine-tuning (FT) strategy to validate its effectiveness. Table~\ref{training} presents the cross-manipulation (c23) AUC results, where DF, FF, FS, and NT denote the testing manipulation types. The models are trained on three manipulations and tested on the remaining one. The two experimental settings employ two training strategies while maintaining the same network architectures. The results in Table~\ref{training} demonstrate that the PEFT approach outperforms the fully FT strategy across all experimental trials. Notably, the average AUC score of PEFT exhibits a substantial enhancement, increasing from 68.2\% to 84.3\%. Hence, it can be concluded that the improvements in face forgery detection performance benefit from the PEFT strategy adopted. Furthermore, the PEFT strategy effectively preserves the abundant ImageNet prior knowledge while learning the forgery-aware knowledge through the designed Adapter and LoRA modules in an adaptive manner, reinforcing the reliability of our research.

\textcolor{black}{Although ImageNet is not face-specific, its learned representations of natural visual statistics (e.g., smooth textures, natural lighting) remain highly relevant, as forgeries often violate these subtle priors (e.g., abnormal texture smoothness, illumination mismatches). A backbone pretrained on ImageNet therefore provides strong priors regarding what ``real'' images should look like, facilitating more effective forgery detection. Retaining these pretrained weights preserves natural image priors that help distinguish real from fake distributions, while PEFT introduces forgery-aware adaptability without overwriting the pretrained knowledge, thereby improving generalization and robustness to unseen domains. Moreover, since forgery artifacts are diverse and continuously evolving techniques, a backbone grounded in natural image distributions offers a stable reference frame, enabling consistent detection across varied manipulation types.}


\begin{table}
  \centering
  \caption{Cross-Manipulation AUC performance on FF++ (c23) dataset with different training strategies.}
  \vspace{-2.5mm}
  \scalebox{0.95}{\begin{tabular}{cccccc}
    \hline
    Strategy & DF & FF & FS & NT & AVG \\	
    \hline				
    Fully FT & 0.896 & 0.731 & 0.496 & 0.603 & 0.682\\
    \hline
    PEFT (Ours) & \textbf{0.962} & \textbf{0.932} & \textbf{0.698} & \textbf{0.781} & \textbf{0.843} \\
    \hline
  \end{tabular}}
  \label{training}
\end{table}

\begin{table}
  \centering
  \caption{Impacts of Forgery Style Mixture (FSM).}
  \vspace{-3mm}
  \scalebox{0.75}{\begin{tabular}{l|ccc|ccc|ccc}
    \hline
    \multirow{2}{*}{\textbf{Setting}} & \multicolumn{3}{c|}{\textbf{CDF}} & \multicolumn{3}{c|}{\textbf{DFR}} & \multicolumn{3}{c}{\textbf{WDF}} \\ \cline{2-10}
     & mAP$\uparrow$ & ACC$\uparrow$ & FNR$\downarrow$ & mAP$\uparrow$ & ACC$\uparrow$ & FNR$\downarrow$ & mAP$\uparrow$ & ACC$\uparrow$ & FNR$\downarrow$ \\	
    \hline				
    w/o FSM & 88.91 & 72.87 & 30.69 & 93.42 & 84.28 & 15.40 & 84.74 & 70.94 & 23.61 \\
    \hline
    w/ FSM & \textbf{90.33} & \textbf{75.05} & \textbf{22.82} & \textbf{95.25} & \textbf{88.22} & \textbf{11.01} & \textbf{86.70} & \textbf{72.77} & \textbf{16.13} \\	
    \hline
  \end{tabular}}
  \label{FSM}
\end{table}


\subsubsection{\textcolor{black}{Effectiveness of FSM}} \textcolor{black}{To further evaluate the effectiveness of the FSM module, we report additional metrics, including mAP, ACC, and FNR, in Table~\ref{FSM}. The cross-dataset evaluation results with and without FSM are presented in Table~\ref{FSM}, showing that the inclusion of FSM effectively improves face forgery detection performance. Notably, FSM significantly enhances the FNR, indicating that it effectively mitigates forgery domain gaps between training and testing sets. This, in turn, enhances the model's ability to detect forgery faces from another point of view.}

\subsubsection{\textcolor{black}{Impacts of training forgery type number}} \textcolor{black}{To investigate the effect of the number of training forgery types, we conducted ablation experiments under the cross-dataset setting. As shown in Table~\ref{forgery_num}, without the Forgery Style Mixture (FSM) module, incorporating more forgery types improves cross-dataset performance, which aligns with the expectation that larger and more diverse training sets enhance generalizability. In contrast, when the FSM module is applied, the model achieves further performance gains on unseen datasets, further confirming its effectiveness in diversifying training forgery style statistics. Moreover, the effectiveness of FSM increases with the number of forgery types. These findings suggest a promising direction for incorporating additional forgery domains to further enrich forgery style diversity and enhance generalizability.}

\begin{figure}[ht]
\centering
\includegraphics[scale=0.235]{  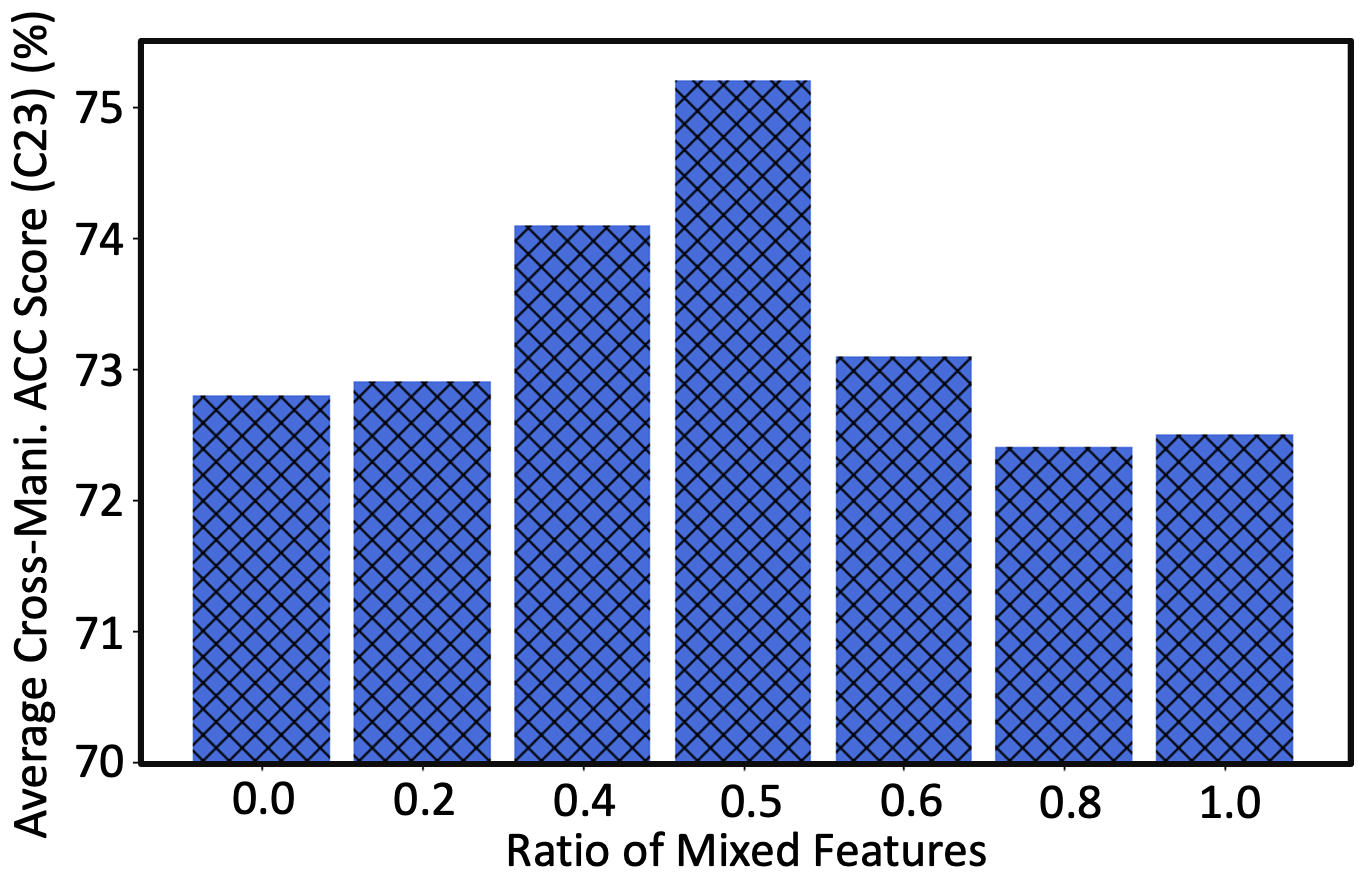}
\vspace{-2.5mm}
\caption{Average cross-manipulation ACC scores across four experimental trials with respect to different mixed feature ratios.}
\label{abl_acc}
\end{figure}

\begin{figure}[ht]
\centering
\includegraphics[scale=0.26]{  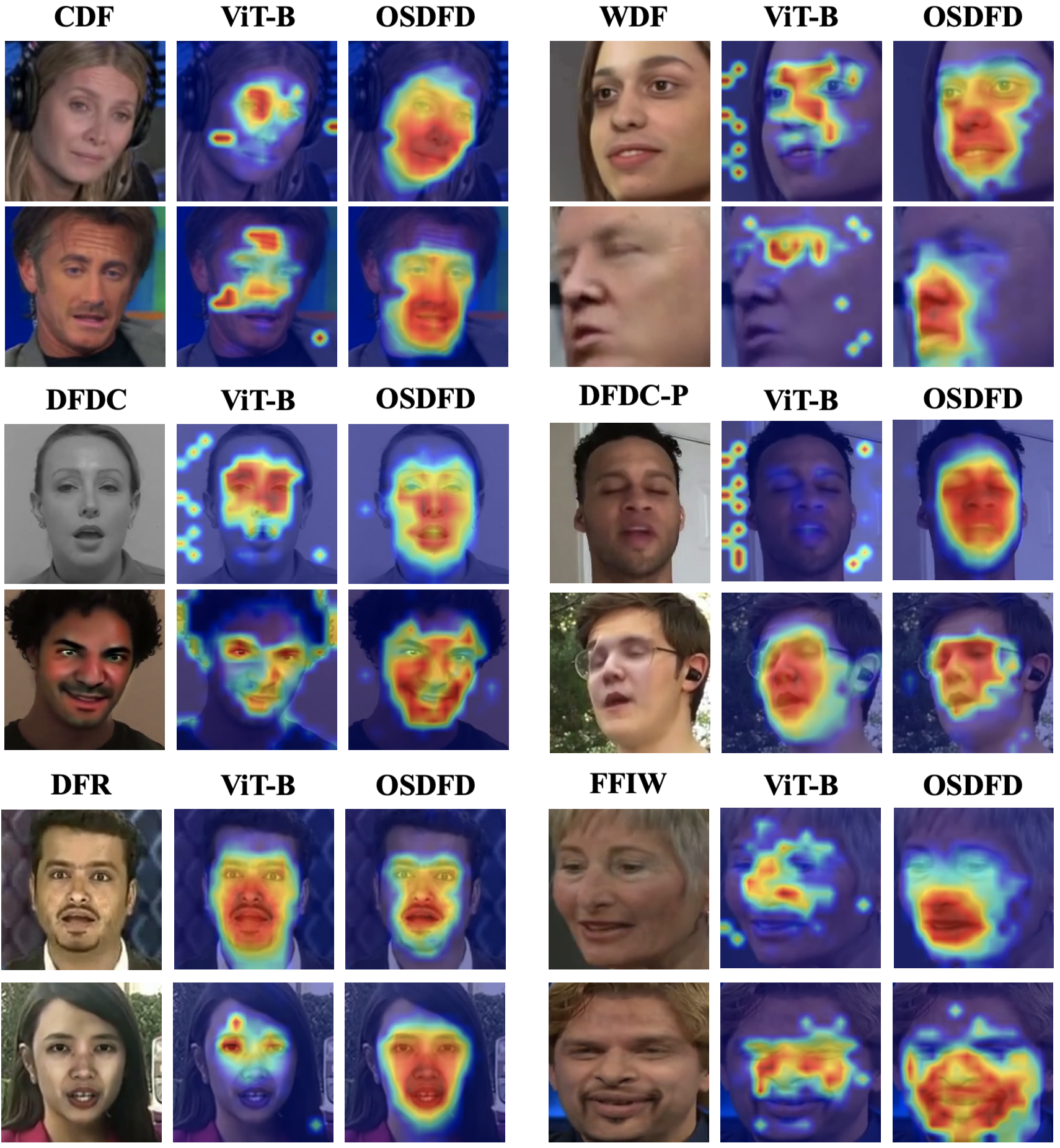}
\vspace{-3.5mm}
\caption{Grad-CAM \cite{selvaraju2017grad}     visualization results of the ViT-B baseline and OSDFD on six unforeseen Deepfake datasets.}
\label{gradcam}
\end{figure}
\subsubsection{\textcolor{black}{Effectiveness of CDC}} 
\textcolor{black}{To validate the effectiveness of the CDC operator, we conduct an ablation study by integrating various conventional convolutions with different kernel sizes. Table~\ref{local} presents the model's cross-dataset experimental results using different local feature extractors, where Conv$_{3\times3}$, Conv$_{5\times5}$, and Conv$_{7\times7}$ represent standard convolutions with kernel sizes of $3\times3$, $5\times5$, and $7\times7$, respectively. The Second-order Difference Convolution (SDC) module, as proposed by Fei et al. \cite{fei2022learning}, employs a second-order local anomaly learning mechanism to detect anomalies in local regions. These baseline results are obtained by replacing the CDC convolution in Fig.~\ref{overview} with the corresponding operators. The results demonstrate that the CDC operator outperforms other local feature extractors across various unforeseen datasets, which can be attributed to its outstanding capability for local high-frequency feature extraction.}

\textcolor{black}{\subsubsection{Impacts of different $\lambda$ values}To investigate the effect of the loss weight of $L_{SCL}$, we conducted ablation experiments under the cross-dataset setting, as reported in Table~\ref{lambda1}. Compared with the baseline ($\lambda$=0), incorporating the Single-Center Loss $L_{SCL}$ effectively enhances the model's generalizability. We varied the $\lambda$ value from 0.6 to 1.4 and found that $\lambda$=1.0 achieves better performance across unseen datasets.}

\begin{figure}[ht]
\centering
\includegraphics[scale=0.24]{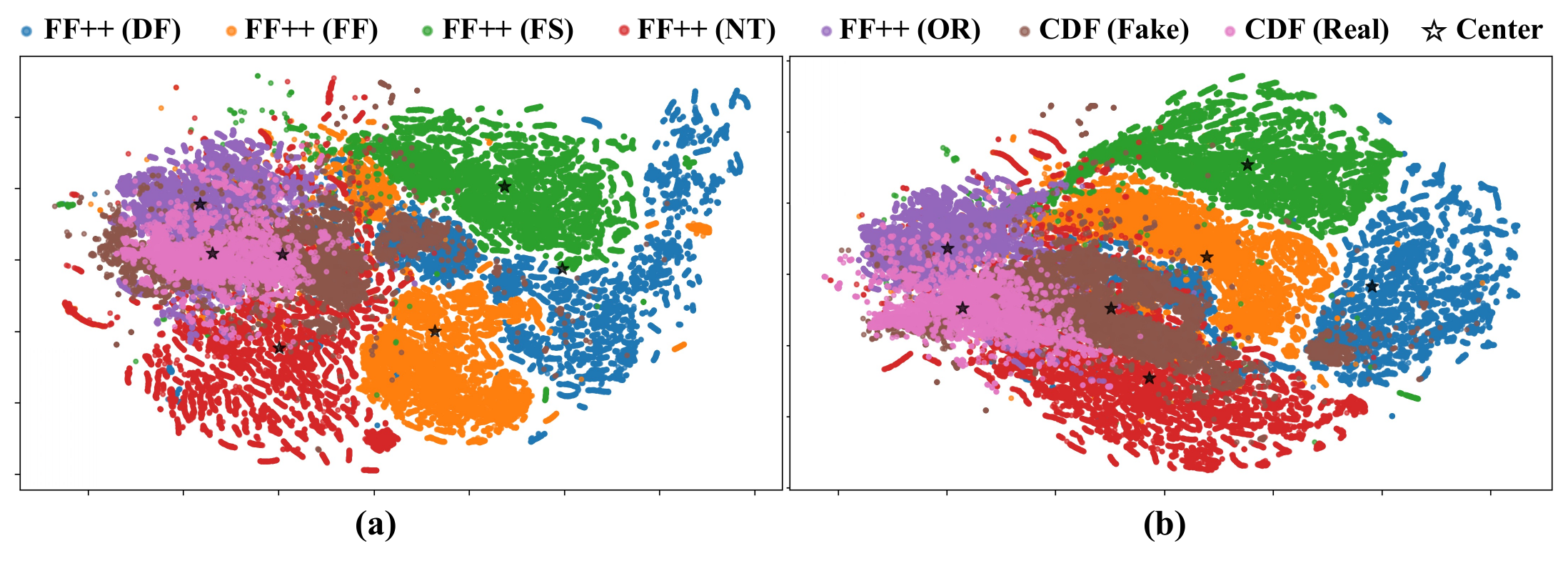}
\vspace{-9mm}
\caption{\textcolor{black}{T-SNE visualizations on FF++ and CDF datasets. (a) ViT-B baseline; (b) ViT-B with Forgery Style Mixture (FSM) module. (DF: Deepfakes, FF: Face2Face, FS: FaceSwap, NT: NeuralTextures, OR: Original.)}} 
\label{tsne_new}
\end{figure}

\subsubsection{Impacts of different ratios of mixed features}
Our proposed method utilizes a feature mixture module to augment forgery domains in the feature space. As illustrated in Fig.~\ref{abl_acc}, the $x$-axis represents the ratio of the mixed forgery features among all forgery features. We analyze the impact of this ratio by plotting the cross-manipulation (c23) ACC scores against ratio values in Fig.~\ref{abl_acc}. Similarly, we present the average score across four cross-manipulation evaluation trials. A ratio of `0.0' indicates the pure use of mixed forgery features during training, while `1.0' means no forgery-style-mixture is applied. Notably, relying solely on original or mixed forgery features results in limited face forgery detection accuracy. In contrast, combining original and mixed forgery features enhances the model's generalization capability by augmenting the source forgery domains. The model achieves the highest average ACC score when the ratio of mixed forgery features is 50\%. This observation underscores the effectiveness of the designed forgery feature mixture module from another perspective.

\begin{table}
  \caption{\textcolor{black}{Ablation experiment on training manipulation type number (cross-dataset evaluation).}}
  \label{forgery_num}
  \vspace{-3mm}
  \centering
  \renewcommand\arraystretch{1.15}
  \scalebox{0.80}{\textcolor{black}{\begin{tabular}{ccccccccccc}
\hline
    \multirow{2}{*}{FSM} & \multirow{2}{*}{DF} & \multirow{2}{*}{FF} & \multirow{2}{*}{FS} & \multirow{2}{*}{NT} & \multicolumn{2}{c}{CDF} & \multicolumn{2}{c}{DFR} & \multicolumn{2}{c}{WDF}  \\ \cline{6-11} 
 & & & & & AUC & EER & AUC & EER & AUC & EER \\ \hline
- & \Checkmark  & - & - & - & 75.85 & 32.08 & 91.22 & 15.86 & 72.13 & 34.43 \\ \hline 					
- & \Checkmark  & \Checkmark & - & - & 77.51 & 29.80 & 92.13 & 15.61 & 73.68 & 33.81 \\ \hline 					
\cellcolor[HTML]{E0DBDB}\Checkmark & \cellcolor[HTML]{E0DBDB}\Checkmark  & \cellcolor[HTML]{E0DBDB}\Checkmark & \cellcolor[HTML]{E0DBDB}- & \cellcolor[HTML]{E0DBDB}- & \cellcolor[HTML]{E0DBDB}78.03 & \cellcolor[HTML]{E0DBDB}29.71 & \cellcolor[HTML]{E0DBDB}92.44 & \cellcolor[HTML]{E0DBDB}14.78 & \cellcolor[HTML]{E0DBDB}74.26 & \cellcolor[HTML]{E0DBDB}31.91 \\ \hline 	
- & \Checkmark  & \Checkmark & \Checkmark & - & 78.12 & 30.01 & 93.01 & 14.57 & 75.18 & 31.82  \\ \hline			
\cellcolor[HTML]{E0DBDB}\Checkmark & \cellcolor[HTML]{E0DBDB}\Checkmark  & \cellcolor[HTML]{E0DBDB}\Checkmark & \cellcolor[HTML]{E0DBDB}\Checkmark & \cellcolor[HTML]{E0DBDB}- & \cellcolor[HTML]{E0DBDB}79.88 & \cellcolor[HTML]{E0DBDB}28.26 & \cellcolor[HTML]{E0DBDB}93.77 & \cellcolor[HTML]{E0DBDB}12.88 & \cellcolor[HTML]{E0DBDB}76.75 & \cellcolor[HTML]{E0DBDB}30.89 \\ \hline 	
- & \Checkmark & \Checkmark & \Checkmark & \Checkmark & 80.76 & 26.58 & 94.18  & 11.69 & 76.68 & 30.76 \\ \hline
\cellcolor[HTML]{E0DBDB}\Checkmark & \cellcolor[HTML]{E0DBDB}\Checkmark & \cellcolor[HTML]{E0DBDB}\Checkmark & \cellcolor[HTML]{E0DBDB}\Checkmark & \cellcolor[HTML]{E0DBDB}\Checkmark & \cellcolor[HTML]{E0DBDB}83.35 & \cellcolor[HTML]{E0DBDB}24.47
&  \cellcolor[HTML]{E0DBDB}95.54 & \cellcolor[HTML]{E0DBDB}9.90 & \cellcolor[HTML]{E0DBDB}78.53 & \cellcolor[HTML]{E0DBDB}28.46 \\ \hline	
\end{tabular}}}
\end{table}



\vspace{-0.2cm}
\subsection{Visualization results}
In this subsection, we provide visualization results to illustrate further the effectiveness of the proposed face forgery detection method. 

\subsubsection{T-SNE feature map visualization} \textcolor{black}{We visualize the embedding distributions of ViT-B baseline (a) w/o and (b) w/ the Forgery Style Mixture (FSM) module in Fig.~\ref{tsne_new} to demonstrate the effectiveness of FSM, which is a key component of our method. The models are trained on the FF++ training sets and tested on both the FF++ test sets and the unseen CDF dataset. Feature centers of each domain are marked with “{$\star$}”. In Fig.~\ref{tsne_new}(a), the FF++ (OR) and CDF (Real) domains are closely aligned, whereas the unseen CDF (Fake) domain is significantly distant from the FF++ fake domains. This explains why the model effectively detects real faces but struggles with fake ones. Motivated by this observation, we design a FSM module to diversify the fake training data and expand the fake feature space without introducing additional data. As shown in Fig.~\ref{tsne_new}(b), the FSM module effectively enlarges the forgery feature space, leading to greater overlap between the FF++ fake domains and the unseen CDF (Fake) domain. Consequently, the decision boundary becomes more robust, enabling improved detection of unseen face forgeries.}

\begin{figure*}[ht]
\centering
\includegraphics[scale=0.36]{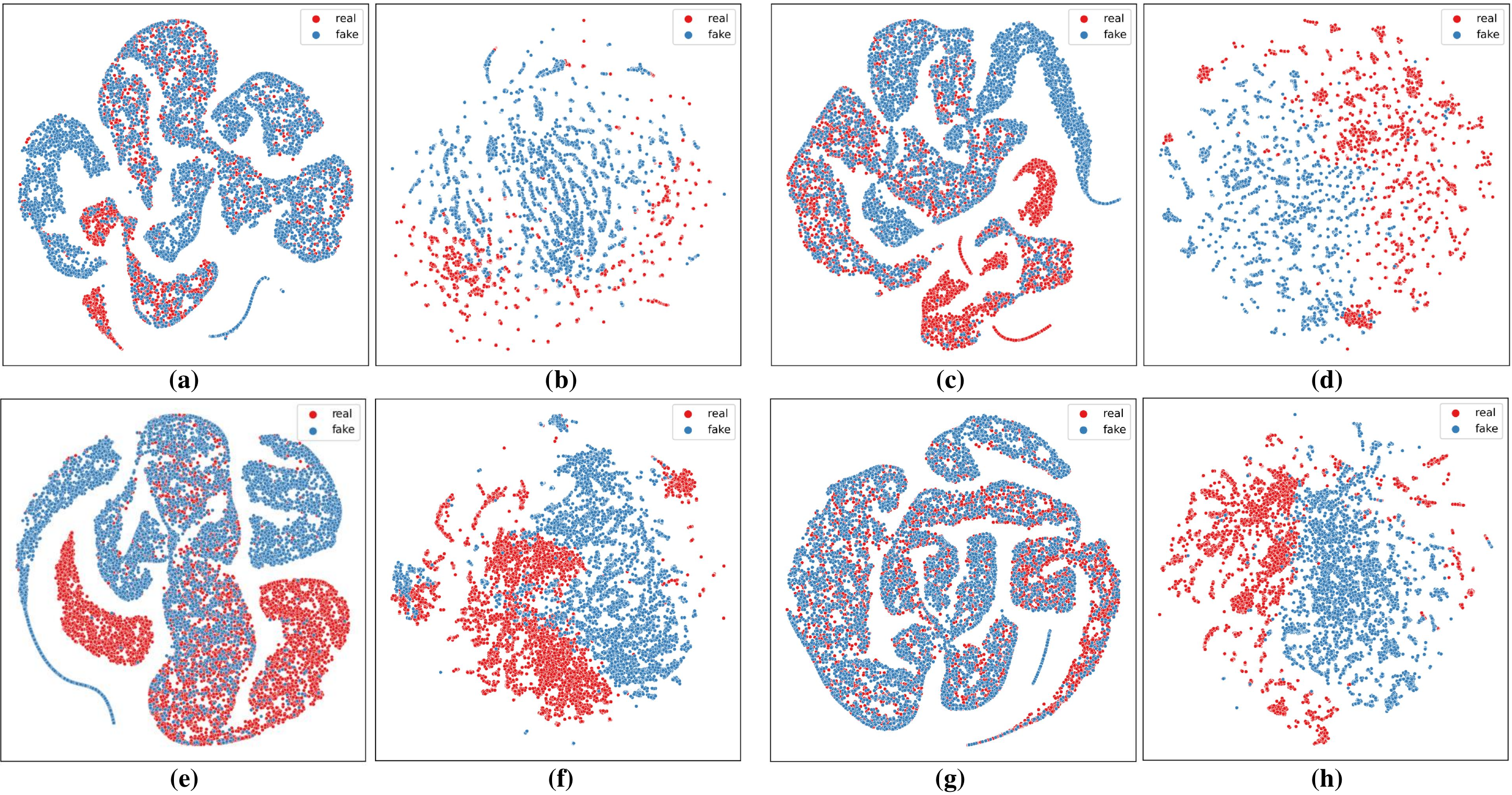}
\vspace{-0.3cm}
\caption{T-SNE feature embedding visualization of (a) the ViT-B baseline on the CDF dataset, (b) OSDFD on the CDF dataset, (c) the ViT-B baseline on the WDF dataset, (d) OSDFD on the WDF dataset, (e) the ViT-B baseline on the DFDC dataset, (f) OSDFD on the DFDC dataset, (g) the ViT-B baseline on the FFIW dataset, and (h) OSDFD on the FFIW dataset. The red and blue dots represent real and fake data samples, respectively. }
\label{tsne}
\end{figure*}

In Fig.~\ref{tsne}, we present the t-SNE \cite{van2008visualizing} feature embedding visualization results of both the ViT-B baseline and our method OSDFD. Fig.~\ref{tsne} (a) and (b) showcase the results on the unseen CDF dataset, while Fig.~\ref{tsne} (c)-(h) depict the results on the WDF, DFDC, and FFIW datasets, respectively. Red and blue dots represent real and fake data samples in these visualizations. The manifolds of real and fake feature distributions entangle for the baseline model. This observation indicates that the baseline model struggles to discriminate forgery faces from the real ones. In contrast, OSDFD effectively separates real and fake images in the feature space. \textcolor{black}{Specifically, for CDF and FFIW datasets, it is obvious that OSDFD can better discriminate real faces from fake ones compared to the ViT baseline. The AUC score goes from 72.35 to 83.35 on the CDF dataset, and 68.49 to 84.61 on the FFIW dataset. In turn, ViT-B can achieve good detection performance on WDF and DFDC datasets, while our proposed method OSDFD, benefiting from the designed forgery-aware parameter efficient fine-tuning and the forgery style mixture modules, exhibits a better face forgery detection performance. }

\vspace{-1mm}
\subsubsection{Grad-CAM map visualization}
We visualize the Grad-CAM \cite{selvaraju2017grad} maps to identify attention regions of the developed face forgery detector for input queries, as shown in Fig.~\ref{gradcam}.
These maps illustrate the results of the ViT-B baseline model and OSDFD across six unforeseen Deepfake datasets: CDF, WDF, DFDC, DFDC-P, DFR, and FFIW. These datasets encompass practical environmental scenarios, including lighting conditions, head poses, image quality, and facial occlusion. Compared to the baseline model, our proposed method accurately highlights informative manipulated regions. The baseline model sometimes focuses on irrelevant non-forgery areas, and these `non-interesting' regions can lead to wrong decisions. Conversely, our method effectively identifies regions with significant forgery artifacts, contributing to superior face forgery detection performance. Specifically, our approach demonstrates an outstanding gradient-based localization across diverse environmental variables, such as extreme head poses in WDF, abnormal color hues in DFDC, facial occlusion (glasses) in DFDC-P, and extreme facial area ratios in FFIW. These observations further underscore the outstanding generalizability of our method in real-world applications.    

\begin{table}
  \centering
  \caption{\textcolor{black}{Comparison with other local-frequency extractor.}} \vspace{-1.0mm}
  \scalebox{0.85}{\begin{tabular}{l|cc|cc|cc|cc}
    \hline
    \multirow{2}{*}{\textbf{Setting}} & \multicolumn{2}{c|}{\textbf{CDF}} & \multicolumn{2}{c|}{\textbf{DFR}} & \multicolumn{2}{c|}{\textbf{WDF}} & \multicolumn{2}{c}{\textbf{AVG}} \\ \cline{2-9}
     & AUC$\uparrow$ & EER$\downarrow$ & AUC$\uparrow$ & EER$\downarrow$ & AUC$\uparrow$ & EER$\downarrow$ & AUC$\uparrow$ & EER$\downarrow$ \\	
    \hline			
    \hline
    Conv$_{3\times3}$ & 82.26 & 26.66 & 94.57 & 11.59 & 78.00 & 29.20  & 84.94 & 22.48 \\
    \hline
    Conv$_{5\times5}$ &80.39  &27.32  &91.35  &15.75  &76.98  &30.18  & 82.91 & 24.42 \\
    \hline
    Conv$_{7\times7}$ &80.44  &27.71  &92.92  &14.25  &78.10  &29.09  & 83.82 & 23.68 \\
    \hline
    SDC \cite{fei2022learning} & 79.14 & 27.89 & 92.87 & 11.21 & 77.72 & \textbf{28.32} & 83.24 & 22.47 \\
    \hline
    CDC & \textbf{83.35} & \textbf{24.47} & \textbf{95.54} & \textbf{9.90} & \textbf{78.53} & 28.46 & \textbf{85.81} & \textbf{20.94} \\
    \hline
  \end{tabular}}
  \label{local}
\end{table}

\begin{table}
  \caption{\textcolor{black}{Ablation experiment on $\lambda$ (cross-dataset evaluation).}}
  \label{lambda1}
  \vspace{-1.0mm}
  \centering
  \renewcommand\arraystretch{1.15}
  \textcolor{black}{\scalebox{1.05}{\begin{tabular}{ccccccc}
\hline
\multirow{2}{*}{$\lambda$}  & \multicolumn{2}{c}{CDF} & \multicolumn{2}{c}{DFR} & \multicolumn{2}{c}{WDF}  \\ \cline{2-7} 
 & AUC & EER & AUC & EER & AUC & EER \\ \hline
0  & 79.85 & 28.86 & 93.62 & 14.04 & 76.82 & 31.71 \\ \hline 
0.6  & 81.33 & 26.98 & 94.88 & 12.69 & 78.21 & 30.81 \\ \hline
0.8  & \underline{82.20} & \underline{25.43} & \underline{95.28} & \underline{11.66} & \textbf{79.03} & \textbf{28.31} \\ \hline
1.0 & \textbf{83.35} & \textbf{24.47} & \textbf{95.54} & \textbf{9.90} & \underline{78.53} & \underline{28.46} \\ \hline
1.2  & 81.93 & 25.52 & 95.02 & 12.51 & 78.44 & 29.02 \\ \hline	
1.4  & 80.86 & 24.41 & 94.66  & 11.96 & 77.97 & 31.05 \\ \hline 
\end{tabular}}}
\end{table}

\begin{table}
  \centering
  \caption{Impacts of different seeds.}
  \vspace{-2mm}
  \scalebox{0.9}{\begin{tabular}{l|ccccc}
   \hline
     & CDF & DFDC-P & DFR & WDF & FFIW\\
    \hline
    AUC & 79.59$\pm$1.83 & 77.24$\pm$1.06 & 95.22$\pm$0.63 & 76.74$\pm$0.95 & 84.82$\pm$0.56 \\
    \hline
    EER & 27.13$\pm$1.46 & 29.73$\pm$0.86 & 9.69$\pm$0.59 & 28.49$\pm$0.44 & 23.21$\pm$0.64 \\	
    \hline
  \end{tabular}}
  \label{seed}
\end{table}

\textcolor{black}{\subsubsection{Cross-dataset results using different seeds} To mitigate the influence of different seeds, we train our models for seven trials using seeds 0-6. The cross-dataset AUC and EER performances are shown in Table.~\ref{seed}. The results indicate that our approach achieves promising performance regardless of the different seeds, demonstrating its robustness across varied initializations. This robustness further underscores the superiority of our method from an additional perspective.}

\section{Conclusions and Future Work}
This paper introduces a parameter-efficient and generalized approach towards open-set face forgery detection. Our method, OSDFD, incorporates lightweight LoRA and Adapter layers into ViT backbones, updating only these modules during training to learn global and local forgery-aware features efficiently. On the other hand, we propose a novel forgery-style-mixture module that augments the forgery source domains. In this vein, our method achieves SOTA generalizability and robustness in face forgery detection. An ablation study further underscores that the designed network modules and objective functions contribute to the final detection performance. \textcolor{black}{Our proposed method, OSDFD, can be flexibly adapted to other transformer-based backbones in a plug-and-play manner.} Further ablation experiments demonstrate that the adopted parameter-efficient training strategy is effective in face forgery detection. Our proposed method will shed light on developing future face forgery detectors for real-world applications.

While our proposed method is effective, we limit our scope herein to non-sequential Deepfake detection. \textcolor{black}{Adapting our approach to leverage abundant sequential information could significantly improve video-level face forgery detection. The designed components can be seamlessly integrated into video transformers. Extending our designed method to video transformers for video-level Deepfake detection opens an important research path forward. We anticipate that the proposed modules can well adapt to video ViTs and enhance the generalizability.} 
\textcolor{black}{Furthermore, our method exhibits strong robustness against common image perturbations, as shown in Fig.~\ref{tmm_robustness}. However, its performance degrades under the highest perturbation severity (level 5). Additionally, although our approach contributes to safeguarding digital media integrity, Deepfake detection models may exhibit bias across different demographic groups due to imbalances in the training dataset. In future work, our aim is to improve robustness and ensure fairness in Deepfake detection.}


\section{Acknowledgement}
This research is supported in part by the National Nature Science Foundation of China (NSFC) under Grant 62502187; in part by the Natural Science Foundation of Jiangxi Province of China under Grant 20252BAC240015; in part by Sichuan Science and Technology Fund 2025ZNSFSC0511; in part by the Research Grant Council (RGC) of Hong Kong SAR, under a GRF Grant 12203124 and an ECS Grant 22201125; in part by the São Paulo Research Foundation (Fapesp) Horus project, Grant \#2023/12865-8 and the Brazilian National Council for Scientific and Technological Development (CNPq), Aletheia project, Grant \#442229/2024-0. This work was carried out at the Rapid-Rich Object Search (ROSE) Lab, School of Electrical \& Electronic Engineering, Nanyang Technological University (NTU), Singapore. This research is supported in part by A*STAR under its OTS Research Programme (Award S24T2TS006). Any opinions, findings and conclusions or recommendations expressed in this material are those of the author(s) and do not reflect the views of the A*STAR.

\ifCLASSOPTIONcaptionsoff
  \newpage
\fi

\bibliographystyle{IEEEtran}
\bibliography{main}

\begin{thebibliography}{100}
\providecommand{\url}[1]{#1}
\csname url@samestyle\endcsname
\providecommand{\newblock}{\relax}
\providecommand{\bibinfo}[2]{#2}
\providecommand{\BIBentrySTDinterwordspacing}{\spaceskip=0pt\relax}
\providecommand{\BIBentryALTinterwordstretchfactor}{4}
\providecommand{\BIBentryALTinterwordspacing}{\spaceskip=\fontdimen2\font plus
\BIBentryALTinterwordstretchfactor\fontdimen3\font minus \fontdimen4\font\relax}
\providecommand{\BIBforeignlanguage}[2]{{%
\expandafter\ifx\csname l@#1\endcsname\relax
\typeout{** WARNING: IEEEtran.bst: No hyphenation pattern has been}%
\typeout{** loaded for the language `#1'. Using the pattern for}%
\typeout{** the default language instead.}%
\else
\language=\csname l@#1\endcsname
\fi
#2}}
\providecommand{\BIBdecl}{\relax}
\BIBdecl

\bibitem{kong2022digital}
C.~Kong, S.~Wang, and H.~Li, ``Digital and physical face attacks: Reviewing and one step further,'' \emph{arXiv preprint arXiv:2209.14692}, 2022.

\bibitem{Wang2024DeepfakeSurvey}
T.~Wang, X.~Liao, K.~P. Chow, X.~Lin, and Y.~Wang, ``Deepfake detection: A comprehensive survey from the reliability perspective,'' \emph{ACM Computing Surveys}, 2024.

\bibitem{cardenuto2023age}
J.~P. Cardenuto, J.~Yang, R.~Padilha, R.~Wan, D.~Moreira, H.~Li, S.~Wang, F.~Andal{\'o}, S.~Marcel, A.~Rocha \emph{et~al.}, ``The age of synthetic realities: Challenges and opportunities,'' \emph{APSIPA Transactions on Signal and Information Processing}, vol.~12, no.~1, 2023.

\bibitem{intelligence2024synthetic}
S.~M. Daniel~Moreira and A.~Rocha, ``Synthetic realities and artificial intelligence-generated contents,'' \emph{IEEE Security \& Privacy}, vol.~22, no.~3, pp. 7--10, 2024.

\bibitem{wang2020face}
X.~Wang, T.~Yao, S.~Ding, and L.~Ma, ``Face manipulation detection via auxiliary supervision,'' in \emph{International Conference on Neural Information Processing}, 2020, pp. 313--324.

\bibitem{li2020identification}
H.~Li, B.~Li, S.~Tan, and J.~Huang, ``Identification of deep network generated images using disparities in color components,'' \emph{Signal Processing}, vol. 174, p. 107616, 2020.

\bibitem{qian2020thinking}
Y.~Qian, G.~Yin, L.~Sheng, Z.~Chen, and J.~Shao, ``Thinking in frequency: Face forgery detection by mining frequency-aware clues,'' in \emph{European Conference on Computer Vision}, 2020, pp. 86--103.

\bibitem{li2018ictu}
Y.~Li, M.-C. Chang, and S.~Lyu, ``In ictu oculi: Exposing ai created fake videos by detecting eye blinking,'' in \emph{IEEE International Workshop on Information Forensics and Security}, 2018, pp. 1--7.

\bibitem{yang2019exposing}
X.~Yang, Y.~Li, and S.~Lyu, ``Exposing deep fakes using inconsistent head poses,'' in \emph{IEEE International Conference on Acoustics, Speech and Signal Processing}, 2019, pp. 8261--8265.

\bibitem{li2018exposing}
Y.~Li and S.~Lyu, ``Exposing deepfake videos by detecting face warping artifacts,'' \emph{arXiv preprint arXiv:1811.00656}, 2018.

\bibitem{chollet2017xception}
F.~Chollet, ``Xception: Deep learning with depthwise separable convolutions,'' in \emph{Proceedings of the IEEE/CVF Conference on Computer Vision and Pattern Recognition}, 2017, pp. 1251--1258.

\bibitem{tan2019efficientnet}
M.~Tan and Q.~Le, ``Efficientnet: Rethinking model scaling for convolutional neural networks,'' in \emph{International Conference on Machine Learning}, 2019, pp. 6105--6114.

\bibitem{nguyen2019capsule}
H.~H. Nguyen, J.~Yamagishi, and I.~Echizen, ``Capsule-forensics: Using capsule networks to detect forged images and videos,'' in \emph{IEEE International Conference on Acoustics, Speech and Signal Processing}, 2019, pp. 2307--2311.

\bibitem{li2020face}
L.~Li, J.~Bao, T.~Zhang, H.~Yang, D.~Chen, F.~Wen, and B.~Guo, ``Face x-ray for more general face forgery detection,'' in \emph{Proceedings of the IEEE/CVF Conference on Computer Vision and Pattern Recognition}, 2020, pp. 5001--5010.

\bibitem{shiohara2022detecting}
K.~Shiohara and T.~Yamasaki, ``Detecting deepfakes with self-blended images,'' in \emph{Proceedings of the IEEE/CVF Conference on Computer Vision and Pattern Recognition}, 2022, pp. 18\,720--18\,729.

\bibitem{sun2022dual}
K.~Sun, T.~Yao, S.~Chen, S.~Ding, J.~Li, and R.~Ji, ``Dual contrastive learning for general face forgery detection,'' in \emph{Proceedings of the AAAI Conference on Artificial Intelligence}, vol.~36, no.~2, 2022, pp. 2316--2324.

\bibitem{dosovitskiy2020image}
A.~Dosovitskiy, L.~Beyer, A.~Kolesnikov, D.~Weissenborn, X.~Zhai, T.~Unterthiner, M.~Dehghani, M.~Minderer, G.~Heigold, S.~Gelly \emph{et~al.}, ``An image is worth 16x16 words: Transformers for image recognition at scale,'' \emph{arXiv preprint arXiv:2010.11929}, 2020.

\bibitem{zhuang2022uia}
W.~Zhuang, Q.~Chu, Z.~Tan, Q.~Liu, H.~Yuan, C.~Miao, Z.~Luo, and N.~Yu, ``Uia-vit: Unsupervised inconsistency-aware method based on vision transformer for face forgery detection,'' in \emph{European Conference on Computer Vision}, 2022, pp. 391--407.

\bibitem{10234594}
Y.~Yu, R.~Ni, S.~Yang, Y.~Zhao, and A.~C. Kot, ``Narrowing domain gaps with bridging samples for generalized face forgery detection,'' \emph{IEEE Transactions on Multimedia}, vol.~26, pp. 3405--3417, 2024.

\bibitem{cheng2024stacking}
J.~Cheng, Z.~Yan, Y.~Zhang, L.~Hao, J.~Ai, Q.~Zou, C.~Li, and Z.~Wang, ``Stacking brick by brick: Aligned feature isolation for incremental face forgery detection,'' \emph{arXiv preprint arXiv:2411.11396}, 2024.

\bibitem{dong2022protecting}
X.~Dong, J.~Bao, D.~Chen, T.~Zhang, W.~Zhang, N.~Yu, D.~Chen, F.~Wen, and B.~Guo, ``Protecting celebrities from deepfake with identity consistency transformer,'' in \emph{Proceedings of the IEEE/CVF Conference on Computer Vision and Pattern Recognition}, 2022, pp. 9468--9478.

\bibitem{xu2024learning}
Y.~Xu, J.~Liang, L.~Sheng, and X.-Y. Zhang, ``Learning spatiotemporal inconsistency via thumbnail layout for face deepfake detection,'' \emph{International Journal of Computer Vision}, vol. 132, no.~12, pp. 5663--5680, 2024.

\bibitem{rossler2019faceforensics++}
A.~Rossler, D.~Cozzolino, L.~Verdoliva, C.~Riess, J.~Thies, and M.~Nie{\ss}ner, ``Faceforensics++: Learning to detect manipulated facial images,'' in \emph{Proceedings of the IEEE/CVF International Conference on Computer Vision}, 2019, pp. 1--11.

\bibitem{dolhansky2019deepfake}
B.~Dolhansky, R.~Howes, B.~Pflaum, N.~Baram, and C.~C. Ferrer, ``The deepfake detection challenge (dfdc) preview dataset,'' \emph{arXiv preprint arXiv:1910.08854}, 2019.

\bibitem{jiang2020deeperforensics}
L.~Jiang, R.~Li, W.~Wu, C.~Qian, and C.~C. Loy, ``Deeperforensics-1.0: A large-scale dataset for real-world face forgery detection,'' in \emph{Proceedings of the IEEE/CVF Conference on Computer Vision and Pattern Recognition}, 2020, pp. 2889--2898.

\bibitem{zi2020wilddeepfake}
B.~Zi, M.~Chang, J.~Chen, X.~Ma, and Y.-G. Jiang, ``Wilddeepfake: A challenging real-world dataset for deepfake detection,'' in \emph{Proceedings of the ACM International Conference on Multimedia}, 2020, pp. 2382--2390.

\bibitem{Zhou_2021_CVPR}
T.~Zhou, W.~Wang, Z.~Liang, and J.~Shen, ``Face forensics in the wild,'' in \emph{Proceedings of the IEEE/CVF Conference on Computer Vision and Pattern Recognition}, June 2021, pp. 5778--5788.

\bibitem{kong2023enhancing}
C.~Kong, H.~Li, and S.~Wang, ``Enhancing general face forgery detection via vision transformer with low-rank adaptation,'' \emph{IEEE International Conference on Multimedia Information Processing and Retrieval}, 2023.

\bibitem{ciftci2020fakecatcher}
U.~A. Ciftci, I.~Demir, and L.~Yin, ``Fakecatcher: Detection of synthetic portrait videos using biological signals,'' \emph{IEEE Transactions on Pattern Analysis and Machine Intelligence}, 2020.

\bibitem{kong2022detect}
C.~Kong, B.~Chen, H.~Li, S.~Wang, A.~Rocha, and S.~Kwong, ``Detect and locate: Exposing face manipulation by semantic-and noise-level telltales,'' \emph{IEEE Transactions on Information Forensics and Security}, vol.~17, pp. 1741--1756, 2022.

\bibitem{luo2021generalizing}
Y.~Luo, Y.~Zhang, J.~Yan, and W.~Liu, ``Generalizing face forgery detection with high-frequency features,'' in \emph{Proceedings of the IEEE/CVF Conference on Computer Vision and Pattern Recognition}, 2021, pp. 16\,317--16\,326.

\bibitem{chen2021robust}
B.~Chen, X.~Liu, Y.~Zheng, G.~Zhao, and Y.-Q. Shi, ``A robust gan-generated face detection method based on dual-color spaces and an improved xception,'' \emph{IEEE Transactions on Circuits and Systems for Video Technology}, vol.~32, no.~6, pp. 3527--3538, 2021.

\bibitem{luo2023beyond}
A.~Luo, C.~Kong, J.~Huang, Y.~Hu, X.~Kang, and A.~C. Kot, ``Beyond the prior forgery knowledge: Mining critical clues for general face forgery detection,'' \emph{IEEE Transactions on Information Forensics and Security}, vol.~19, pp. 1168--1182, 2023.

\bibitem{zhang2023bi}
D.~Zhang, C.~Fu, D.~Lu, J.~Li, and Y.~Zhang, ``Bi-source reconstruction-based classification network for face forgery video detection,'' \emph{IEEE Transactions on Circuits and Systems for Video Technology}, vol.~34, no.~6, pp. 4257--4269, 2023.

\bibitem{zhang2024face}
D.~Zhang, J.~Chen, X.~Liao, F.~Li, J.~Chen, and G.~Yang, ``Face forgery detection via multi-feature fusion and local enhancement,'' \emph{IEEE Transactions on Circuits and Systems for Video Technology}, 2024.

\bibitem{nguyen2024laa}
D.~Nguyen, N.~Mejri, I.~P. Singh, P.~Kuleshova, M.~Astrid, A.~Kacem, E.~Ghorbel, and D.~Aouada, ``Laa-net: Localized artifact attention network for quality-agnostic and generalizable deepfake detection,'' in \emph{Proceedings of the IEEE/CVF Conference on Computer Vision and Pattern Recognition}, 2024, pp. 17\,395--17\,405.

\bibitem{wang2023noise}
T.~Wang and K.~P. Chow, ``Noise based deepfake detection via multi-head relative-interaction,'' in \emph{Proceedings of the AAAI Conference on Artificial Intelligence}, vol.~37, no.~12, 2023, pp. 14\,548--14\,556.

\bibitem{guan2022delving}
J.~Guan, H.~Zhou, Z.~Hong, E.~Ding, J.~Wang, C.~Quan, and Y.~Zhao, ``Delving into sequential patches for deepfake detection,'' \emph{Proceedings of the Advances in Neural Information Processing Systems}, vol.~35, pp. 4517--4530, 2022.

\bibitem{cheng2023voice}
H.~Cheng, Y.~Guo, T.~Wang, Q.~Li, X.~Chang, and L.~Nie, ``Voice-face homogeneity tells deepfake,'' \emph{ACM Transactions on Multimedia Computing, Communications and Applications}, vol.~20, no.~3, pp. 1--22, 2023.

\bibitem{wang2023deep}
T.~Wang, H.~Cheng, K.~P. Chow, and L.~Nie, ``Deep convolutional pooling transformer for deepfake detection,'' \emph{ACM Transactions on Multimedia Computing, Communications and Applications}, vol.~19, no.~6, pp. 1--20, 2023.

\bibitem{yu2023narrowing}
Y.~Yu, R.~Ni, S.~Yang, Y.~Zhao, and A.~C. Kot, ``Narrowing domain gaps with bridging samples for generalized face forgery detection,'' \emph{IEEE Transactions on Multimedia}, vol.~26, pp. 3405--3417, 2023.

\bibitem{cheng2024diffusion}
H.~Cheng, Y.~Guo, T.~Wang, L.~Nie, and M.~Kankanhalli, ``Diffusion facial forgery detection,'' in \emph{Proceedings of the ACM International Conference on Multimedia}, 2024, pp. 5939--5948.

\bibitem{luo2024forgery}
A.~Luo, R.~Cai, C.~Kong, Y.~Ju, X.~Kang, J.~Huang, and A.~C. Kot, ``Forgery-aware adaptive learning with vision transformer for generalized face forgery detection,'' \emph{IEEE Transactions on Circuits and Systems for Video Technology}, 2024.

\bibitem{yu2024distilling}
L.~Yu, T.~Xie, C.~Liu, G.~Jin, Z.~Ding, and H.~Xie, ``Distilling multi-level semantic cues across multi-modalities for face forgery detection,'' \emph{IEEE Transactions on Circuits and Systems for Video Technology}, 2024.

\bibitem{wang2023spatial}
Y.~Wang, C.~Peng, D.~Liu, N.~Wang, and X.~Gao, ``Spatial-temporal frequency forgery clue for video forgery detection in vis and nir scenario,'' \emph{IEEE Transactions on Circuits and Systems for Video Technology}, vol.~33, no.~12, pp. 7943--7956, 2023.

\bibitem{guo2023ldfnet}
Z.~Guo, L.~Wang, W.~Yang, G.~Yang, and K.~Li, ``Ldfnet: Lightweight dynamic fusion network for face forgery detection by integrating local artifacts and global texture information,'' \emph{IEEE Transactions on Circuits and Systems for Video Technology}, vol.~34, no.~2, pp. 1255--1265, 2023.

\bibitem{liao2023famm}
X.~Liao, Y.~Wang, T.~Wang, J.~Hu, and X.~Wu, ``Famm: Facial muscle motions for detecting compressed deepfake videos over social networks,'' \emph{IEEE Transactions on Circuits and Systems for Video Technology}, vol.~33, no.~12, pp. 7236--7251, 2023.

\bibitem{miao2023f}
C.~Miao, Z.~Tan, Q.~Chu, H.~Liu, H.~Hu, and N.~Yu, ``F2trans: High-frequency fine-grained transformer for face forgery detection,'' \emph{IEEE Transactions on Information Forensics and Security}, vol.~18, pp. 1039--1051, 2023.

\bibitem{cui2025forensics}
X.~Cui, Y.~Li, A.~Luo, J.~Zhou, and J.~Dong, ``Forensics adapter: Adapting clip for generalizable face forgery detection,'' in \emph{Proceedings of the IEEE/CVF International Conference on Computer Vision}, 2025, pp. 19\,207--19\,217.

\bibitem{guo2025face}
Z.~Guo, Y.~Liu, J.~Zhang, H.~Zheng, and S.~Shan, ``Face forgery video detection via temporal forgery cue unraveling,'' in \emph{Proceedings of the IEEE/CVF International Conference on Computer Vision}, 2025, pp. 7396--7405.

\bibitem{guo2025towards}
M.~Guo, Q.~Yin, W.~Lu, and X.~Luo, ``Towards open-world generalized deepfake detection: General feature extraction via unsupervised domain adaptation,'' \emph{Proceedings of the ACM International Conference on Multimedia}, 2025.

\bibitem{lai2024gm}
Y.~Lai, Z.~Yu, J.~Yang, B.~Li, X.~Kang, and L.~Shen, ``Gm-df: Generalized multi-scenario deepfake detection,'' \emph{Proceedings of the ACM International Conference on Multimedia}, 2025.

\bibitem{cheng2025fair}
H.~Cheng, M.-H. Liu, Y.~Guo, T.~Wang, L.~Nie, and M.~Kankanhalli, ``Fair deepfake detectors can generalize,'' \emph{Proceedings of the Advances in Neural Information Processing Systems}, 2025.

\bibitem{huang2024ffaa}
Z.~Huang, B.~Xia, Z.~Lin, Z.~Mou, W.~Yang, and J.~Jia, ``Ffaa: Multimodal large language model based explainable open-world face forgery analysis assistant,'' \emph{arXiv preprint arXiv:2408.10072}, 2024.

\bibitem{jia2024can}
S.~Jia, R.~Lyu, K.~Zhao, Y.~Chen, Z.~Yan, Y.~Ju, C.~Hu, X.~Li, B.~Wu, and S.~Lyu, ``Can chatgpt detect deepfakes? a study of using multimodal large language models for media forensics,'' in \emph{Proceedings of the IEEE/CVF Conference on Computer Vision and Pattern Recognition}, 2024, pp. 4324--4333.

\bibitem{yu2025unlocking}
P.~Yu, J.~Fei, H.~Gao, X.~Feng, Z.~Xia, and C.~H. Chang, ``Unlocking the capabilities of vision-language models for generalizable and explainable deepfake detection,'' \emph{arXiv preprint arXiv:2503.14853}, 2025.

\bibitem{zhang2025mfclip}
Y.~Zhang, T.~Wang, Z.~Yu, Z.~Gao, L.~Shen, and S.~Chen, ``Mfclip: Multi-modal fine-grained clip for generalizable diffusion face forgery detection,'' \emph{IEEE Transactions on Information Forensics and Security}, 2025.

\bibitem{guo2025rethinking}
X.~Guo, X.~Song, Y.~Zhang, X.~Liu, and X.~Liu, ``Rethinking vision-language model in face forensics: Multi-modal interpretable forged face detector,'' in \emph{Proceedings of the IEEE/CVF International Conference on Computer Vision}, 2025, pp. 105--116.

\bibitem{hu2025seeing}
J.~Hu, S.~Fan, and T.~Sim, ``Seeing through deepfakes: A human-inspired framework for multi-face detection,'' \emph{Proceedings of the IEEE/CVF International Conference on Computer Vision}, 2025.

\bibitem{chen2024x2}
Y.~Chen, Z.~Yan, G.~Cheng, K.~Zhao, S.~Lyu, and B.~Wu, ``X2-dfd: A framework for explainable and extendable deepfake detection,'' \emph{Proceedings of the Advances in Neural Information Processing Systems}, 2025.

\bibitem{houlsby2019parameter}
N.~Houlsby, A.~Giurgiu, S.~Jastrzebski, B.~Morrone, Q.~De~Laroussilhe, A.~Gesmundo, M.~Attariyan, and S.~Gelly, ``Parameter-efficient transfer learning for nlp,'' in \emph{International Conference on Machine Learning}, 2019, pp. 2790--2799.

\bibitem{sung2022lst}
Y.-L. Sung, J.~Cho, and M.~Bansal, ``Lst: Ladder side-tuning for parameter and memory efficient transfer learning,'' \emph{Proceedings of the Advances in Neural Information Processing Systems}, vol.~35, pp. 12\,991--13\,005, 2022.

\bibitem{hu2021lora}
E.~J. Hu, Y.~Shen, P.~Wallis, Z.~Allen-Zhu, Y.~Li, S.~Wang, L.~Wang, and W.~Chen, ``Lora: Low-rank adaptation of large language models,'' \emph{arXiv preprint arXiv:2106.09685}, 2021.

\bibitem{jia2022visual}
M.~Jia, L.~Tang, B.-C. Chen, C.~Cardie, S.~Belongie, B.~Hariharan, and S.-N. Lim, ``Visual prompt tuning,'' in \emph{European Conference on Computer Vision}, 2022, pp. 709--727.

\bibitem{jie2022convolutional}
S.~Jie and Z.-H. Deng, ``Convolutional bypasses are better vision transformer adapters,'' \emph{arXiv preprint arXiv:2207.07039}, 2022.

\bibitem{zhang2022neural}
Y.~Zhang, K.~Zhou, and Z.~Liu, ``Neural prompt search,'' \emph{arXiv preprint arXiv:2206.04673}, 2022.

\bibitem{ramasesh2021effect}
V.~V. Ramasesh, A.~Lewkowycz, and E.~Dyer, ``Effect of scale on catastrophic forgetting in neural networks,'' in \emph{International Conference on Learning Representations}, 2021.

\bibitem{su2021pixel}
Z.~Su, W.~Liu, Z.~Yu, D.~Hu, Q.~Liao, Q.~Tian, M.~Pietik{\"a}inen, and L.~Liu, ``Pixel difference networks for efficient edge detection,'' in \emph{Proceedings of the IEEE/CVF International Conference on Computer Vision}, 2021, pp. 5117--5127.

\bibitem{miao2022hierarchical}
C.~Miao, Z.~Tan, Q.~Chu, N.~Yu, and G.~Guo, ``Hierarchical frequency-assisted interactive networks for face manipulation detection,'' \emph{IEEE Transactions on Information Forensics and Security}, vol.~17, pp. 3008--3021, 2022.

\bibitem{fei2022learning}
J.~Fei, Y.~Dai, P.~Yu, T.~Shen, Z.~Xia, and J.~Weng, ``Learning second order local anomaly for general face forgery detection,'' in \emph{Proceedings of the IEEE/CVF Conference on Computer Vision and Pattern Recognition}, 2022, pp. 20\,270--20\,280.

\bibitem{yang2021mtd}
J.~Yang, A.~Li, S.~Xiao, W.~Lu, and X.~Gao, ``Mtd-net: learning to detect deepfakes images by multi-scale texture difference,'' \emph{IEEE Transactions on Information Forensics and Security}, vol.~16, pp. 4234--4245, 2021.

\bibitem{huang2017arbitrary}
X.~Huang and S.~Belongie, ``Arbitrary style transfer in real-time with adaptive instance normalization,'' in \emph{Proceedings of the IEEE International Conference on Computer Vision}, 2017, pp. 1501--1510.

\bibitem{zhou2021domain}
K.~Zhou, Y.~Yang, Y.~Qiao, and T.~Xiang, ``Domain generalization with mixstyle,'' \emph{arXiv preprint arXiv:2104.02008}, 2021.

\bibitem{he2016deep}
K.~He, X.~Zhang, S.~Ren, and J.~Sun, ``Deep residual learning for image recognition,'' in \emph{Proceedings of the IEEE/CVF Conference on Computer Vision and Pattern Recognition}, 2016, pp. 770--778.

\bibitem{cozzolino2018forensictransfer}
D.~Cozzolino, J.~Thies, A.~R{\"o}ssler, C.~Riess, M.~Nie{\ss}ner, and L.~Verdoliva, ``Forensictransfer: Weakly-supervised domain adaptation for forgery detection,'' \emph{arXiv preprint arXiv:1812.02510}, 2018.

\bibitem{nguyen2019multi}
H.~H. Nguyen, F.~Fang, J.~Yamagishi, and I.~Echizen, ``Multi-task learning for detecting and segmenting manipulated facial images and videos,'' in \emph{IEEE International Conference on Biometrics Theory, Applications and Systems}, 2019, pp. 1--8.

\bibitem{li2018learning}
D.~Li, Y.~Yang, Y.-Z. Song, and T.~Hospedales, ``Learning to generalize: Meta-learning for domain generalization,'' in \emph{Proceedings of the AAAI Conference on Artificial Intelligence}, vol.~32, no.~1, 2018.

\bibitem{sun2021domain}
K.~Sun, H.~Liu, Q.~Ye, Y.~Gao, J.~Liu, L.~Shao, and R.~Ji, ``Domain general face forgery detection by learning to weight,'' in \emph{Proceedings of the AAAI conference on Artificial Intelligence}, vol.~35, no.~3, 2021, pp. 2638--2646.

\bibitem{10315169}
A.~Luo, C.~Kong, J.~Huang, Y.~Hu, X.~Kang, and A.~C. Kot, ``Beyond the prior forgery knowledge: Mining critical clues for general face forgery detection,'' \emph{IEEE Transactions on Information Forensics and Security}, vol.~19, pp. 1168--1182, 2024.

\bibitem{huang2023implicit}
B.~Huang, Z.~Wang, J.~Yang, J.~Ai, Q.~Zou, Q.~Wang, and D.~Ye, ``Implicit identity driven deepfake face swapping detection,'' in \emph{Proceedings of the IEEE/CVF Conference on Computer Vision and Pattern Recognition}, 2023, pp. 4490--4499.

\bibitem{yan2023ucf}
Z.~Yan, Y.~Zhang, Y.~Fan, and B.~Wu, ``Ucf: Uncovering common features for generalizable deepfake detection,'' in \emph{Proceedings of the IEEE/CVF International Conference on Computer Vision}, 2023, pp. 22\,412--22\,423.

\bibitem{sun2024diffusionfake}
K.~Sun, S.~Chen, T.~Yao, H.~Liu, X.~Sun, S.~Ding, and R.~Ji, ``Diffusionfake: Enhancing generalization in deepfake detection via guided stable diffusion,'' \emph{arXiv preprint arXiv:2410.04372}, 2024.

\bibitem{radford2021learning}
A.~Radford, J.~W. Kim, C.~Hallacy, A.~Ramesh, G.~Goh, S.~Agarwal, G.~Sastry, A.~Askell, P.~Mishkin, J.~Clark \emph{et~al.}, ``Learning transferable visual models from natural language supervision,'' in \emph{International Conference on Machine Learning}, 2021, pp. 8748--8763.

\bibitem{king2009dlib}
D.~E. King, ``Dlib-ml: A machine learning toolkit,'' \emph{The Journal of Machine Learning Research}, vol.~10, pp. 1755--1758, 2009.

\bibitem{paszke2019pytorch}
A.~Paszke, S.~Gross, F.~Massa, A.~Lerer, J.~Bradbury, G.~Chanan, T.~Killeen, Z.~Lin, N.~Gimelshein, L.~Antiga \emph{et~al.}, ``Pytorch: An imperative style, high-performance deep learning library,'' \emph{Proceedings of the Advances in Neural Information Processing Systems}, vol.~32, 2019.

\bibitem{kingma2014adam}
D.~P. Kingma and J.~Ba, ``Adam: A method for stochastic optimization,'' \emph{arXiv preprint arXiv:1412.6980}, 2014.

\bibitem{deng2009imagenet}
J.~Deng, W.~Dong, R.~Socher, L.-J. Li, K.~Li, and L.~Fei-Fei, ``Imagenet: A large-scale hierarchical image database,'' in \emph{Proceedings of the IEEE/CVF Conference on Computer Vision and Pattern Recognition}, 2009, pp. 248--255.

\bibitem{hm16_20}
Deepfakes, [Online]. Available: \url{https://github.com/deepfakes/faceswap}, 2018.

\bibitem{thies2016face2face}
J.~Thies, M.~Zollhofer, M.~Stamminger, C.~Theobalt, and M.~Nie{\ss}ner, ``Face2face: Real-time face capture and reenactment of rgb videos,'' in \emph{Proceedings of the IEEE/CVF Conference on Computer Vision and Pattern Recognition}, 2016, pp. 2387--2395.

\bibitem{dfcode}
FaceSwap, [Online]. Available: \url{https://github.com/MarekKowalski/FaceSwap}, 2016.

\bibitem{thies2019deferred}
J.~Thies, M.~Zollh{\"o}fer, and M.~Nie{\ss}ner, ``Deferred neural rendering: Image synthesis using neural textures,'' \emph{Acm Transactions on Graphics}, vol.~38, no.~4, pp. 1--12, 2019.

\bibitem{cao2022end}
J.~Cao, C.~Ma, T.~Yao, S.~Chen, S.~Ding, and X.~Yang, ``End-to-end reconstruction-classification learning for face forgery detection,'' in \emph{Proceedings of the IEEE/CVF Conference on Computer Vision and Pattern Recognition}, 2022, pp. 4113--4122.

\bibitem{wang2023dynamic}
Y.~Wang, K.~Yu, C.~Chen, X.~Hu, and S.~Peng, ``Dynamic graph learning with content-guided spatial-frequency relation reasoning for deepfake detection,'' in \emph{Proceedings of the IEEE/CVF Conference on Computer Vision and Pattern Recognition}, 2023, pp. 7278--7287.

\bibitem{yan2024transcending}
Z.~Yan, Y.~Luo, S.~Lyu, Q.~Liu, and B.~Wu, ``Transcending forgery specificity with latent space augmentation for generalizable deepfake detection,'' in \emph{Proceedings of the IEEE/CVF Conference on Computer Vision and Pattern Recognition}, 2024, pp. 8984--8994.

\bibitem{cao2024towards}
J.~Cao, K.-Y. Zhang, T.~Yao, S.~Ding, X.~Yang, and C.~Ma, ``Towards unified defense for face forgery and spoofing attacks via dual space reconstruction learning,'' \emph{International Journal of Computer Vision}, pp. 1--26, 2024.

\bibitem{tan2024frequency}
C.~Tan, Y.~Zhao, S.~Wei, G.~Gu, P.~Liu, and Y.~Wei, ``Frequency-aware deepfake detection: Improving generalizability through frequency space domain learning,'' in \emph{Proceedings of the AAAI Conference on Artificial Intelligence}, vol.~38, no.~5, 2024, pp. 5052--5060.

\bibitem{fu2025exploring}
X.~Fu, Z.~Yan, T.~Yao, S.~Chen, and X.~Li, ``Exploring unbiased deepfake detection via token-level shuffling and mixing,'' in \emph{Proceedings of the AAAI Conference on Artificial Intelligence}, vol.~39, no.~3, 2025, pp. 3040--3048.

\bibitem{wang2022lisiam}
J.~Wang, Y.~Sun, and J.~Tang, ``Lisiam: Localization invariance siamese network for deepfake detection,'' \emph{IEEE Transactions on Information Forensics and Security}, vol.~17, pp. 2425--2436, 2022.

\bibitem{zheng2021exploring}
Y.~Zheng, J.~Bao, D.~Chen, M.~Zeng, and F.~Wen, ``Exploring temporal coherence for more general video face forgery detection,'' in \emph{Proceedings of the IEEE/CVF Conference on Computer Vision and Pattern Recognition}, 2021, pp. 15\,044--15\,054.

\bibitem{10138555}
Y.~Yu, R.~Ni, Y.~Zhao, S.~Yang, F.~Xia, N.~Jiang, and G.~Zhao, ``Msvt: Multiple spatiotemporal views transformer for deepfake video detection,'' \emph{IEEE Transactions on Circuits and Systems for Video Technology}, vol.~33, no.~9, pp. 4462--4471, 2023.

\bibitem{cai2023glitch}
Z.~Cai, S.~Ghosh, A.~Dhall, T.~Gedeon, K.~Stefanov, and M.~Hayat, ``Glitch in the matrix: A large scale benchmark for content driven audio--visual forgery detection and localization,'' \emph{Computer Vision and Image Understanding}, vol. 236, p. 103818, 2023.

\bibitem{li2020celeb}
Y.~Li, X.~Yang, P.~Sun, H.~Qi, and S.~Lyu, ``Celeb-df: A large-scale challenging dataset for deepfake forensics,'' in \emph{Proceedings of the IEEE/CVF Conference on Computer Vision and Pattern Recognition}, 2020, pp. 3207--3216.

\bibitem{yan2024df40}
Z.~Yan, T.~Yao, S.~Chen, Y.~Zhao, X.~Fu, J.~Zhu, D.~Luo, C.~Wang, S.~Ding, Y.~Wu \emph{et~al.}, ``Df40: Toward next-generation deepfake detection,'' \emph{Proceedings of the Advances in Neural Information Processing Systems}, vol.~37, pp. 29\,387--29\,434, 2024.

\bibitem{khalid2021fakeavceleb}
H.~Khalid, S.~Tariq, M.~Kim, and S.~S. Woo, ``Fakeavceleb: A novel audio-video multimodal deepfake dataset,'' \emph{arXiv preprint arXiv:2108.05080}, 2021.

\bibitem{liu2021swin}
Z.~Liu, Y.~Lin, Y.~Cao, H.~Hu, Y.~Wei, Z.~Zhang, S.~Lin, and B.~Guo, ``Swin transformer: Hierarchical vision transformer using shifted windows,'' in \emph{Proceedings of the IEEE/CVF International Conference on Computer Vision}, 2021, pp. 10\,012--10\,022.

\bibitem{selvaraju2017grad}
R.~R. Selvaraju, M.~Cogswell, A.~Das, R.~Vedantam, D.~Parikh, and D.~Batra, ``Grad-cam: Visual explanations from deep networks via gradient-based localization,'' in \emph{Proceedings of the IEEE International Conference on Computer Vision}, 2017, pp. 618--626.

\bibitem{van2008visualizing}
L.~Van~der Maaten and G.~Hinton, ``Visualizing data using t-sne,'' \emph{Journal of Machine Learning Research}, vol.~9, no.~11, 2008.

\end{thebibliography}

\begin{IEEEbiography}[{\includegraphics[width=1in,height=1.25in,clip,keepaspectratio]{ 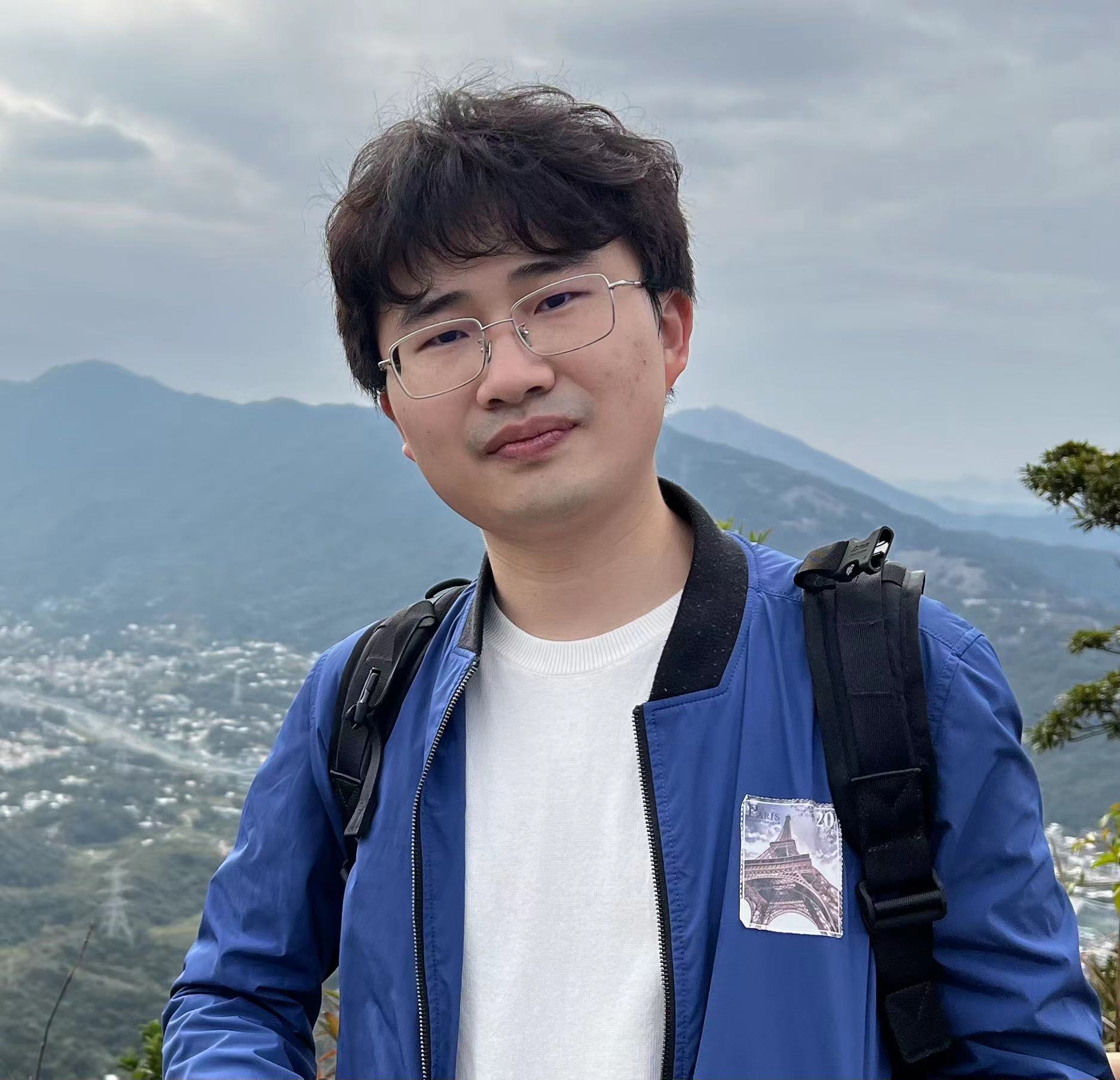}}]{Chenqi Kong} received the B.S. and M.S. degrees from Harbin Institute of Technology, Harbin, China, in 2017 and 2019, respectively. He received the Ph.D. degree in the Department of Computer Science, City University of Hong Kong, Hong Kong SAR, China in 2023. He is currently a research fellow at ROSE Lab, School of Electrical and Electronic Engineering, Nanyang Technological University, Singapore. His research interests include AI security, information forensics, and trustworthy AI.
\end{IEEEbiography}
\vspace{-5mm}
\begin{IEEEbiography}[{\includegraphics[width=1in,height=1.25in,clip,keepaspectratio]{ 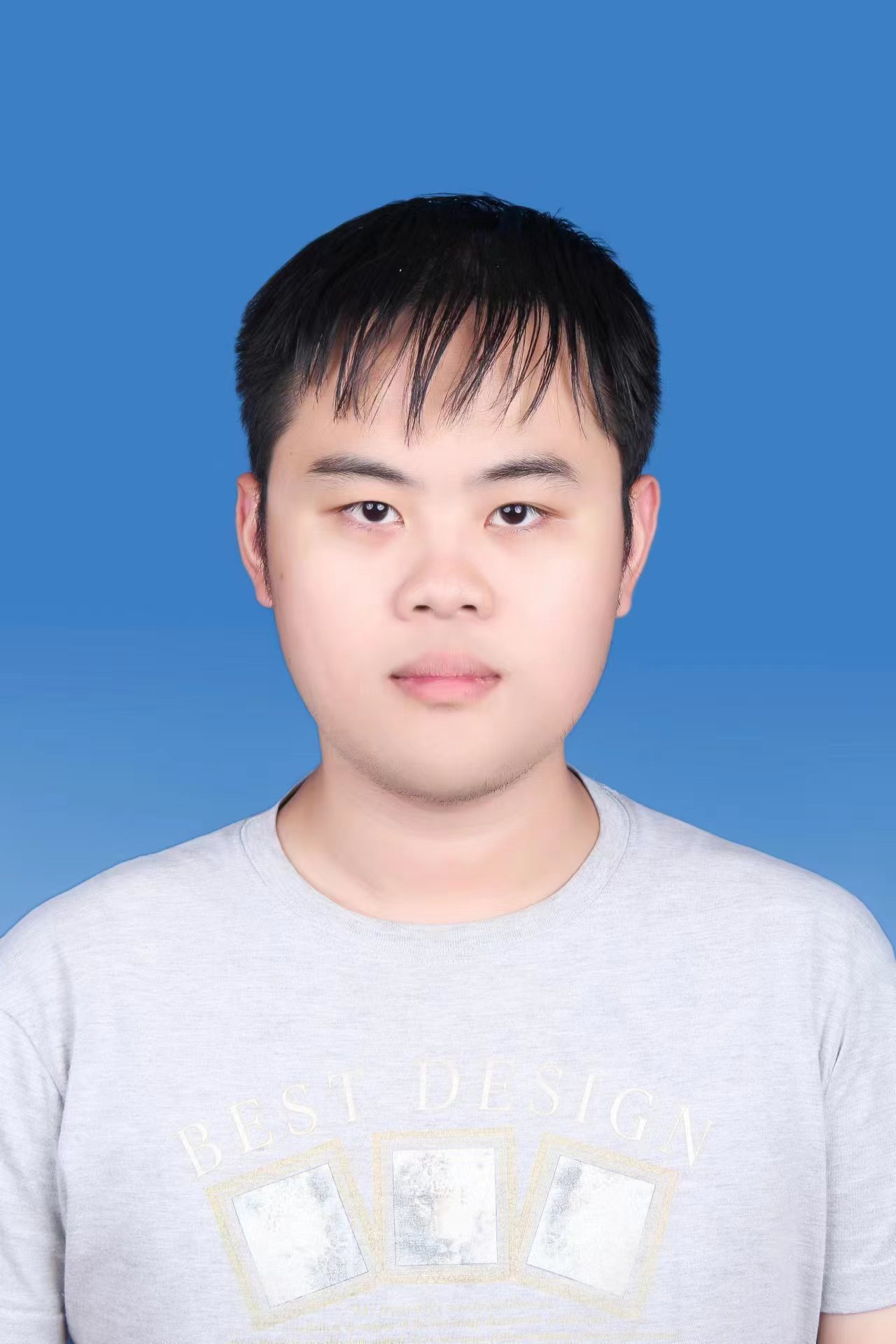}}]{Anwei Luo} received the B.S. degree from Jilin University, Changchun, China, in 2013 and ph. D. degree from Sun Yat-sen University, Guangzhou,
China, in 2024. Currently, he is a lecture with the School of Computing and Artificial Intelligence,Jiangxi University of Finance and Economics, Nanchang. His research interests include digital multimedia forensics, watermarking and AI security.
\end{IEEEbiography}

\vspace{-5mm}
\begin{IEEEbiography}
[{\includegraphics[width=1in,height=1.25in]{ 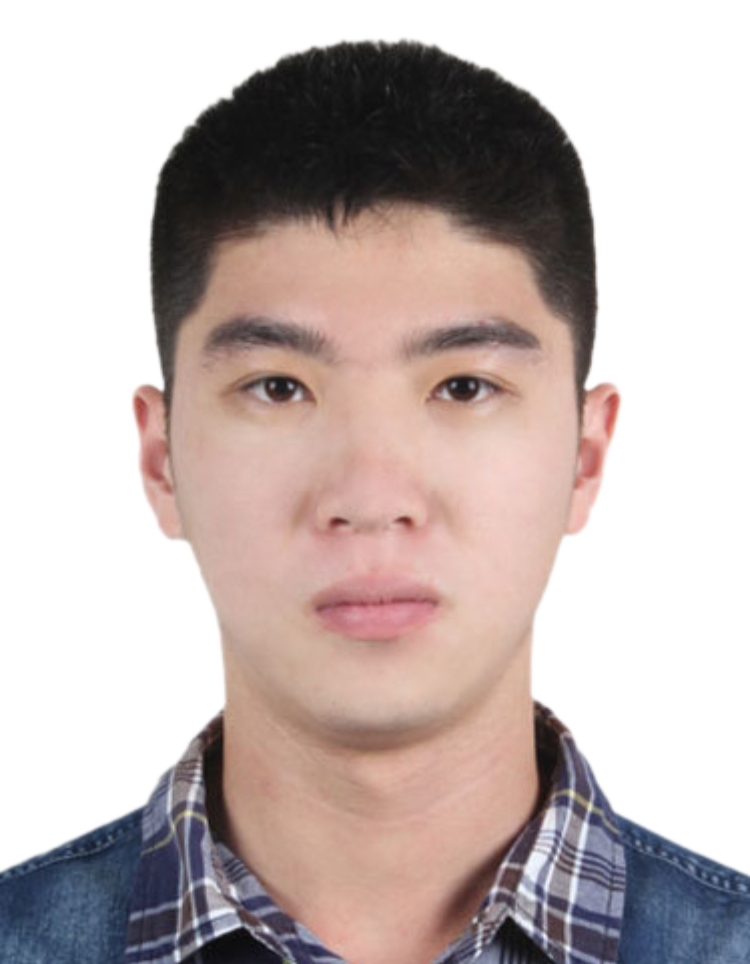}}]
{Peijun Bao} is currently a postdoctoral research fellow at Nanyang Technological University, Singapore. He received the B.S. (Hons.) degree from Northwestern Polytechnical University, China, the M.S. degree from Peking University, China, and the Ph.D. degree from Nanyang Technological University, Singapore. He has published multiple first-author papers in top-tier conferences, including CVPR, ICCV, ECCV, and AAAI, and has served as a reviewer for major venues such as CVPR, ICCV, ECCV, and TIP. His research interests include video semantic analysis and multimodal learning.
\end{IEEEbiography}

\vspace{-5mm}
\begin{IEEEbiography}[{\includegraphics[width=1in,height=1.25in,clip,keepaspectratio]{ 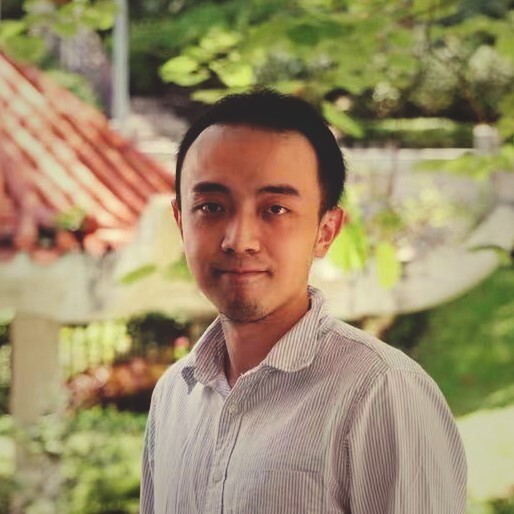}}]{Haoliang Li} received the B.S. degree in communication engineering from University of Electronic Science and Technology of China (UESTC) in 2013, and his Ph.D. degree from Nanyang Technological University (NTU), Singapore in 2018. He is currently an assistant professor in Department of Electrical Engineering, City University of Hong Kong. His research mainly focuses on AI security, multimedia forensics and transfer learning. His research works appear in international journals/conferences such as TPAMI, IJCV, TIFS, NeurIPS, CVPR and AAAI. He received the Wallenberg-NTU presidential postdoc fellowship in 2019, doctoral innovation award in 2019, and VCIP best paper award in 2020.  
\end{IEEEbiography}
\vspace{-2mm}
\begin{IEEEbiography}[{\includegraphics[width=1in,height=1.25in,clip,keepaspectratio]{ 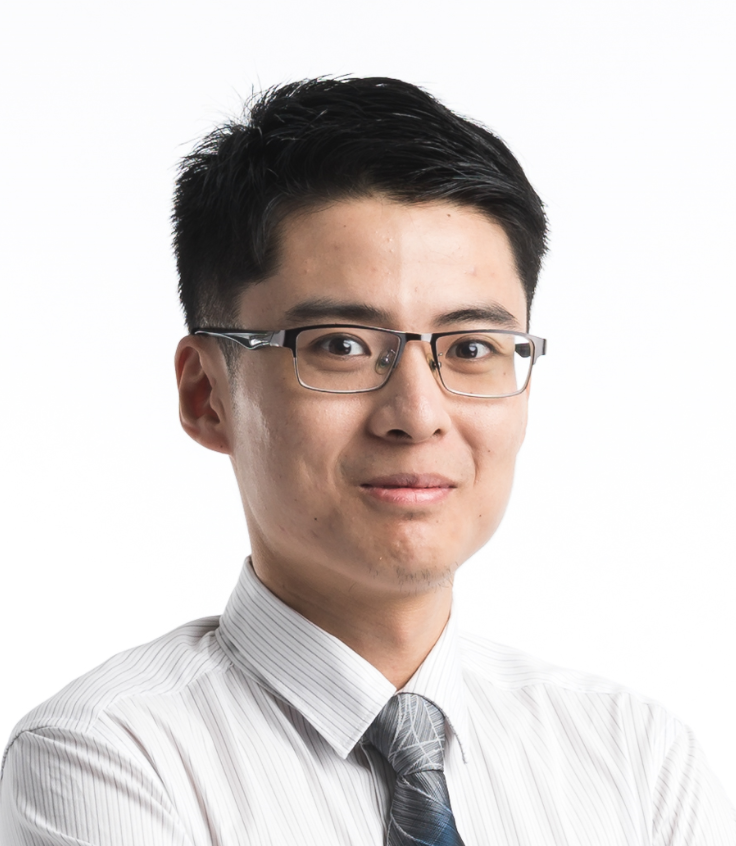}}]{Renjie Wan} (Member, IEEE) received the B.Eng. degree from the University of Electronic Science and Technology of China in 2012 and the Ph.D. degree from Nanyang Technological University, Singapore, in 2019. He is currently an Assistant Professor with the Department of Computer Science, Hong Kong Baptist University, Hong Kong. He was a recipient of the Microsoft CRSF Award, the 2020 VCIP Best Paper Award, and the Wallenberg-NTU Presidential Postdoctoral Fellowship. He is the outstanding reviewer of the 2019 ICCV.
\end{IEEEbiography}

\begin{IEEEbiography}[{\includegraphics[width=1in,height=1.25in,clip,keepaspectratio]{ 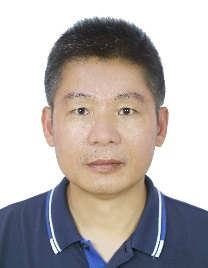}}]{Zengwei Zheng} He received the B.S and M. Ec degrees in Computer Science and Western Economics from Hangzhou University, China in 1991 and 1998, respectively. He received the Ph. D degree in Computer Science and Technology from Zhejiang University, China in 2005. He is currently a Qiantang Distinguished Professor of the School of Computer and Computing Science, the director of Zhejiang Provincial Engineering Research Center for Intelligent Plant Factory, the director of Zhejiang International Cooperation Base for Science and Technology on Artificial Intelligence of Things (AIoT) Technology \& Systems, the director of Hangzhou Key Laboratory for IoT Technology \& Application at Hangzhou City University. His research interests include Artificial Intelligence of Things (AIoT), Artificial Intelligence, Pervasive Computing and Digital Agriculture. He is a senior member of the CCF and a member of the IEEE.
\end{IEEEbiography}

\begin{IEEEbiography}[{\includegraphics[width=1in,height=1.25in,clip,keepaspectratio]{ 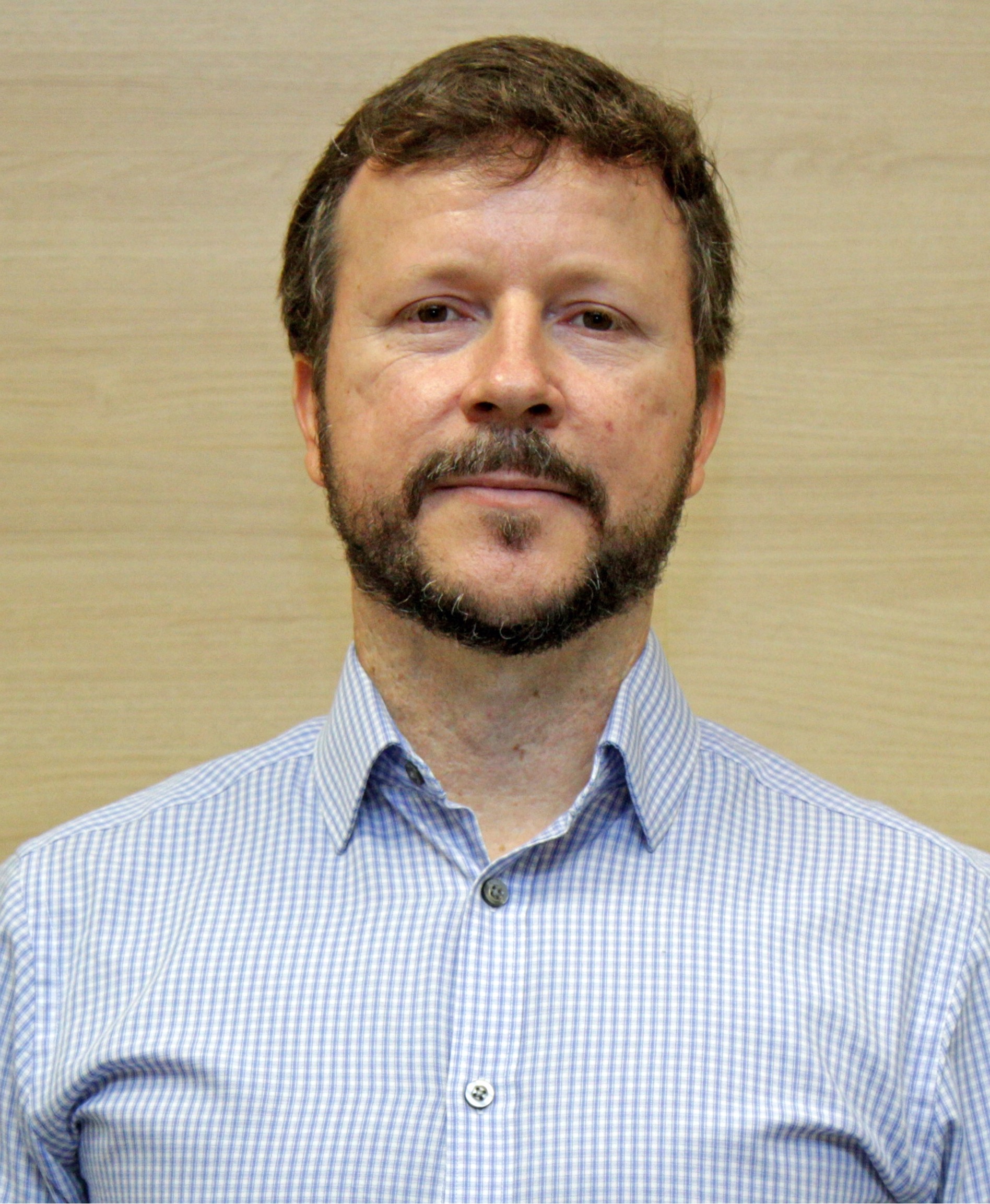}}]{Anderson Rocha} received his Ph.D. degree in computer science. He is a full professor of artificial intelligence and digital forensics at the Institute of Computing, University of Campinas, Campinas 13083-852, Brazil, where he is the coordinator of the Artificial Intelligence Lab. A Microsoft and Google Faculty Fellow, he is a former chair of the IEEE Information Forensics and Security Technical Committee (2019–2020) and an affiliated member of the Brazilian Academy of Sciences and the Brazilian Academy of Forensics Sciences. His research interests include artificial intelligence, digital forensics, and reasoning for complex data. He is a Fellow of IEEE.
\end{IEEEbiography}

\begin{IEEEbiography}[{\includegraphics[width=1in,height=1.25in,clip,keepaspectratio]{ 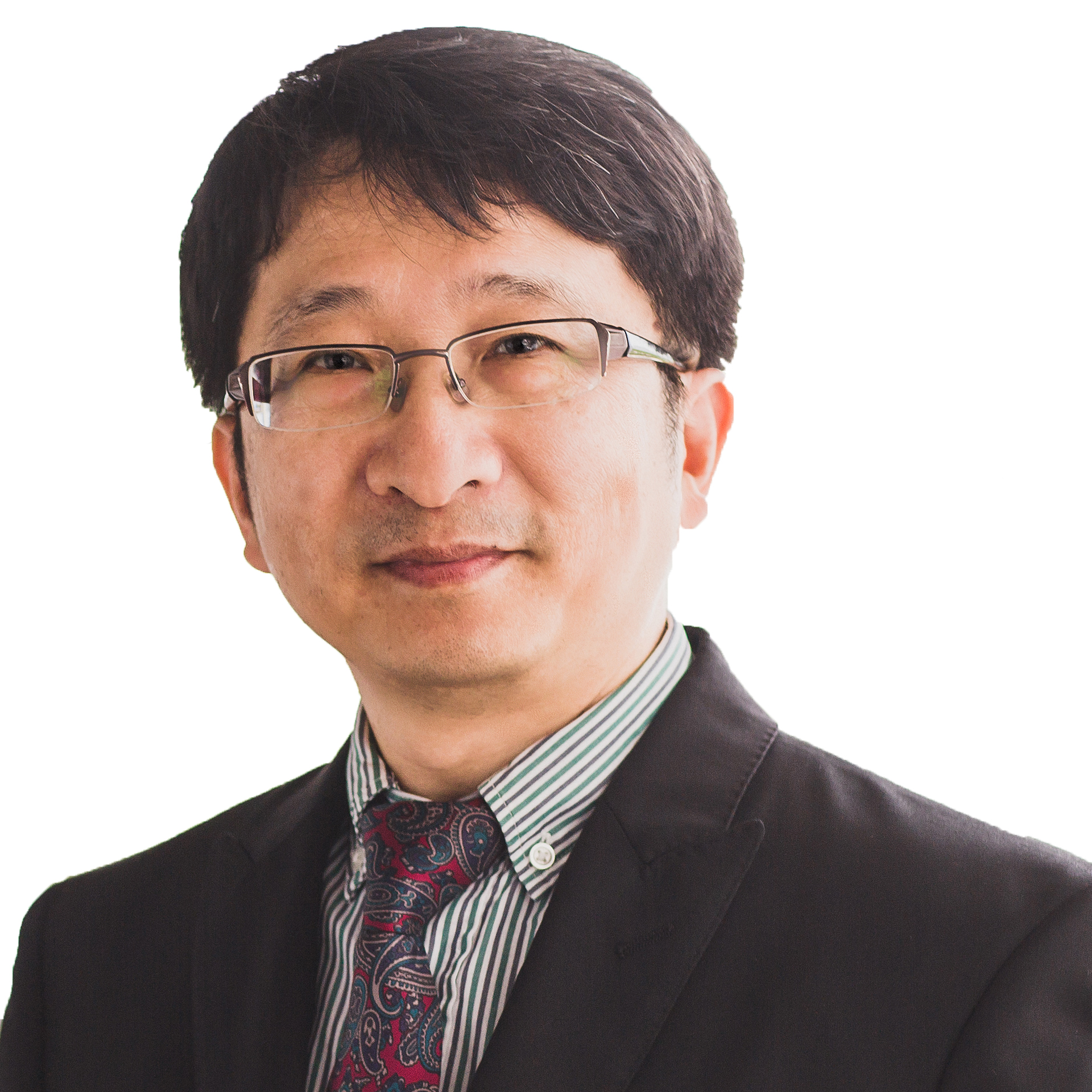}}]
{Alex C. Kot} is Fellow of the Academy of Engineering Singapore and IEEE Life Fellow. He is currently Emeritus Professor at the School of Electrical and Electronic Engineering, Nanyang Technological University (NTU), Singapore, and serves as Chief Scientist at Shenzhen MSU-BIT University (SMBU). He has long been engaged in research on signal processing, artificial intelligence, and multimedia security. He founded the Rapid-Rich Object Search (ROSE) Lab at NTU, secured more than SGD 25 million in research funding, and has published extensively in leading international journals and conferences. His research has made internationally recognized contributions in areas including signal processing for communication, digital image forensics, face anti-spoofing detection, and person re-identification, with key technologies successfully deployed in various applications.
\end{IEEEbiography}

\end{document}